\documentclass[5p,twocolumn,preprint]{elsarticle}

\usepackage{multicol}
\usepackage[bookmarks=true]{hyperref}
\usepackage{enumerate}
\usepackage{graphics} 
\usepackage{epsfig} 
\usepackage{wrapfig}
\usepackage{amsmath} 
\usepackage{amssymb}  
\usepackage{amsthm} 
\usepackage{bm}  
\usepackage{subcaption}
\usepackage[font=footnotesize,labelfont=footnotesize]{caption}
\usepackage{hyperref}
\usepackage{color}
\usepackage{todonotes}
\usepackage[capitalise]{cleveref}
\usepackage{siunitx}
\usepackage{adjustbox} 
\usepackage{booktabs}
\usepackage{tikz}
\usetikzlibrary{arrows.meta}
\usepackage{tabularx, colortbl}

\graphicspath{{Figures/}}

\DeclareSIUnit{\million}{\text{MM}}
\newcommand{\etal}{\MakeLowercase{\textit{et al.\ }}}

\newcommand{\textquote}[1]{\textquotedblleft #1\textquotedblright}
\newcommand{\tphsmm}{\mathbf{\Theta}}
\definecolor{lightestGray}{gray}{0.9}
\def\normal{\mathcal{N}}
\def\R{\mathbb{R}}
\renewcommand{\vec}[1]{\boldsymbol{#1}}
\newcommand{\trsp}{\mathsf{T}}
\newcommand{\transp}[2]{\mathcal{A}_{\parallel_{#1}^{#2}}}

\DeclareMathOperator{\st}{subject~to}
\DeclareMathOperator*{\argmax}{argmax}

\DeclareRobustCommand{\graydemo}{\raisebox{2pt}{\tikz{\draw[gray,solid,line width = 1.1pt](0,0) -- (4mm,0);}}}
\DeclareRobustCommand{\blackrepro}{\raisebox{2pt}{\tikz{\draw[black,solid,line width = 1.1pt](0,0) -- (4mm,0);}}}
\DeclareRobustCommand{\greenellipse}{\tikz{ \filldraw[color=black!50!green, fill=black!10!green, thick] (1.5,0) ellipse (.35 and 0.1);}}

\DeclareRobustCommand{\pinkellipse}{\tikz{ \filldraw[color=magenta, fill=black!10!pink, thick] (1.5,0) ellipse (.35 and 0.1);}}
\DeclareRobustCommand{\greencircle}{\tikz{ \filldraw[color=black!50!green, fill=black!10!green, thick](0,0) circle (.1);}}
\DeclareRobustCommand{\orangecircle}{\tikz{ \filldraw[color=black!30!orange, fill=orange, thick](0,0) circle (.1);}}
\DeclareRobustCommand{\pinkcircle}{\tikz{ \filldraw[color=magenta, fill=black!10!pink, thick](0,0) circle (.1);}}
\DeclareRobustCommand{\blackdashed}{\raisebox{2pt}{\tikz{\draw[black,dashed,line width = 1.1pt](0,0) -- (5mm,0);}}}
\DeclareRobustCommand{\blackarrow}{{\tikz{\draw[-{Latex[length=2mm]}] (0,0)--(0.6,0);}}}

\journal{}

\begin{document}
	\begin{frontmatter}
		
	\title{The e-Bike Motor Assembly:\\ Towards Advanced Robotic Manipulation for Flexible Manufacturing}
	
	\author[bcai]{Leonel Rozo\corref{cor1}}
	\ead{leonel.rozo@de.bosch.com}
	\author[bcai]{Andras G. Kupcsik}
	\author[bcai]{Philipp Schillinger}
	\author[UnPe]{Meng Guo}		
	\author[bcai]{Robert Krug}
	\author[bcai]{Niels van Duijkeren} 
	\author[bcai]{Markus Spies} 
	\author[bcai]{Patrick Kesper} 
	\author[bcai]{Sabrina Hoppe}
	\author[bcai]{Hanna Ziesche}
	\author[bcai]{Mathias B\"{u}rger} 
	\author[bcai]{Kai O. Arras}
	
	\cortext[cor1]{Corresponding author}
	
	\affiliation[bcai]{organization={Bosch Center for Artificial Intelligence},
		city={Renningen},
		country={Germany}}

	\affiliation[UnPe]{organization={College of Engineering (CoE) at Peking University},
	city={Beijing},
	country={China}}	
	
	\begin{abstract}
		Robotic manipulation is currently undergoing a profound paradigm shift due to the increasing needs for flexible manufacturing systems, and at the same time, because of the advances in enabling technologies such as sensing, learning, optimization, and hardware. 
		This demands for robots that can observe and reason about their workspace, and that are skillfull enough to complete various assembly processes in weakly-structured settings. 
		Moreover, it remains a great challenge to enable operators for teaching robots \emph{on-site}, while managing the inherent complexity of perception, control, motion planning and reaction to unexpected situations.   
		Motivated by real-world industrial applications, this paper demonstrates the potential of such a paradigm shift in robotics on the industrial case of an e-Bike motor assembly.  
		The paper presents a concept for teaching and programming adaptive robots \emph{on-site} and demonstrates their potential for the named applications.
		The framework includes: \emph{(i)} a method to teach perception systems onsite in a self-supervised manner, \emph{(ii)} a general representation of object-centric motion skills and force-sensitive assembly skills, both learned from demonstration, \emph{(iii)} a sequencing approach that exploits a human-designed plan to perform complex tasks, and \emph{(iv)} a system solution for adapting and optimizing skills online. 
		The aforementioned components are interfaced through a four-layer software architecture that makes our framework a tangible industrial technology.
		To demonstrate the generality of the proposed framework, we provide, in addition to the motivating e-Bike motor assembly, a further case study on dense box packing for logistics automation.
	\end{abstract}


\end{frontmatter}

	\section{Motivation}
	\label{sec:motivation}
	Flexible manufacturing is a core concept of Industry 4.0 that involves the use of advanced automation and robotics technologies to build efficient and adaptable manufacturing processes. 
This approach allows manufacturers to quickly adjust production processes to meet changing customer demands, reduce waste, and improve overall efficiency. 
To implement flexible manufacturing, companies must work with a range of advanced sensors, control systems, and robots, which demands a high degree of integration and communication between different systems and components. 
Robots play a critical role in flexible manufacturing systems (FMS) due to their ability to perform a wide range of different tasks such as pick and place, quality control, visual inspection, among others~\citep{ZhihaoLiu22:RobotLearningSmartManuf}. 

Nevertheless, many advanced industrial tasks common in FMS (e.g. assembly) are still hard to automate by robotic manipulators.
Particular challenges include the perception of unstructured workspaces, operation in dynamic environments, and physical interaction with tools~\citep{JingzhouSong21:AssemblyLfDGMM} or manufactured elements~\cite{stoica2017berkeley,yang2018grand}.
More specifically, some parts often need to be picked up from arbitrary poses and then put together or processed for various subsequent steps.
However, the particular choice of how and where to pick such a part often depends on how it needs to be handled in the remaining process.
Explicitly programming such a task requires a lot of manually-defined conditions and carefully-tuned parameters~\cite{johannsmeier2019framework}, which is a challenge even for experienced human operators.

\begin{figure}[t]
	\centering
	\begin{subfigure}[b]{0.485\textwidth}
		\centering
		\includegraphics[width=0.99\linewidth]{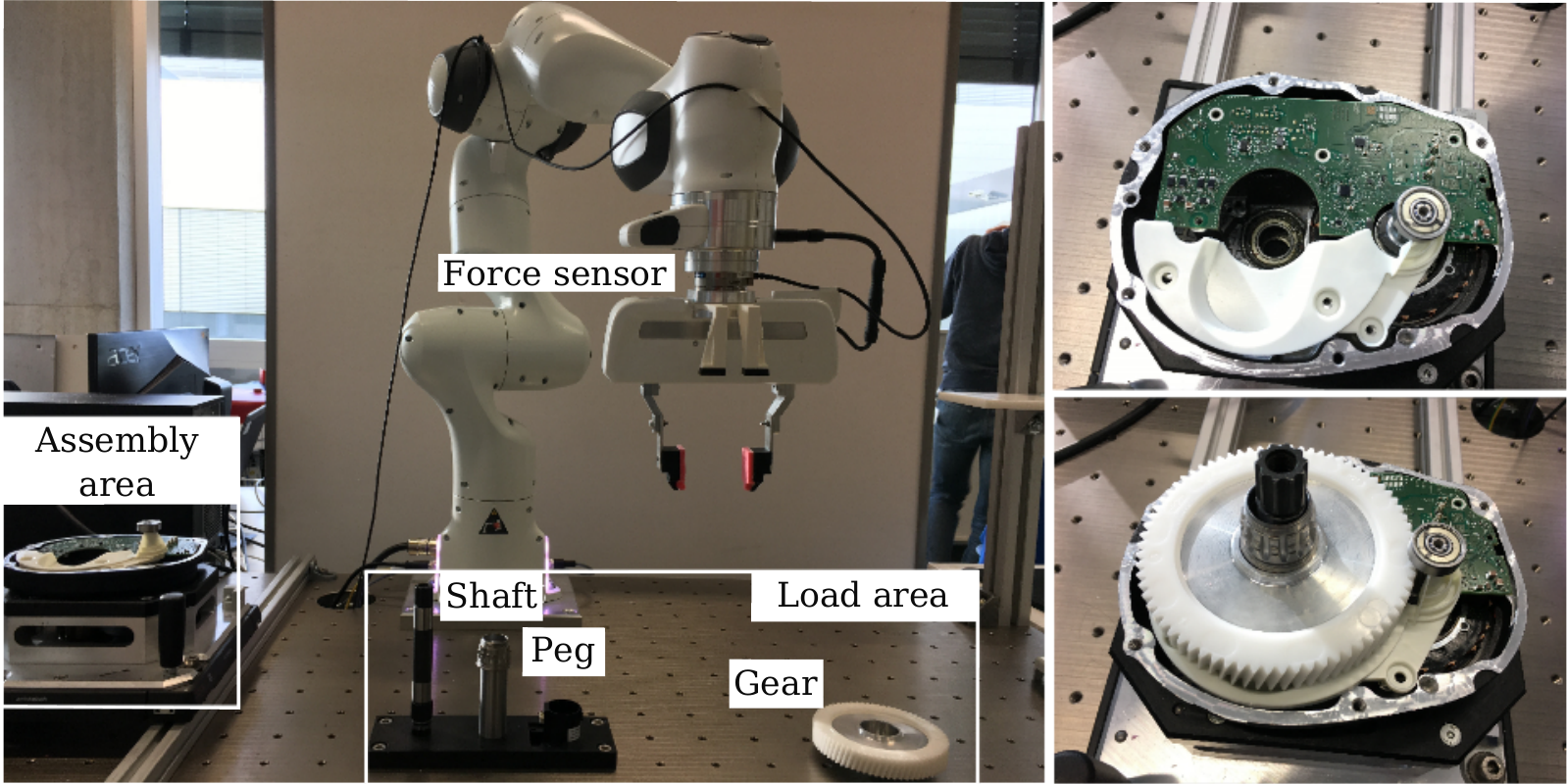}
		\caption{Experimental setup of the e-Bike motor assembly case.}
		\label{subfig:AssemblySetup}
	\end{subfigure}
	\\
	\begin{subfigure}[b]{0.48\textwidth}
		\centering
		\includegraphics[width=0.323\linewidth]{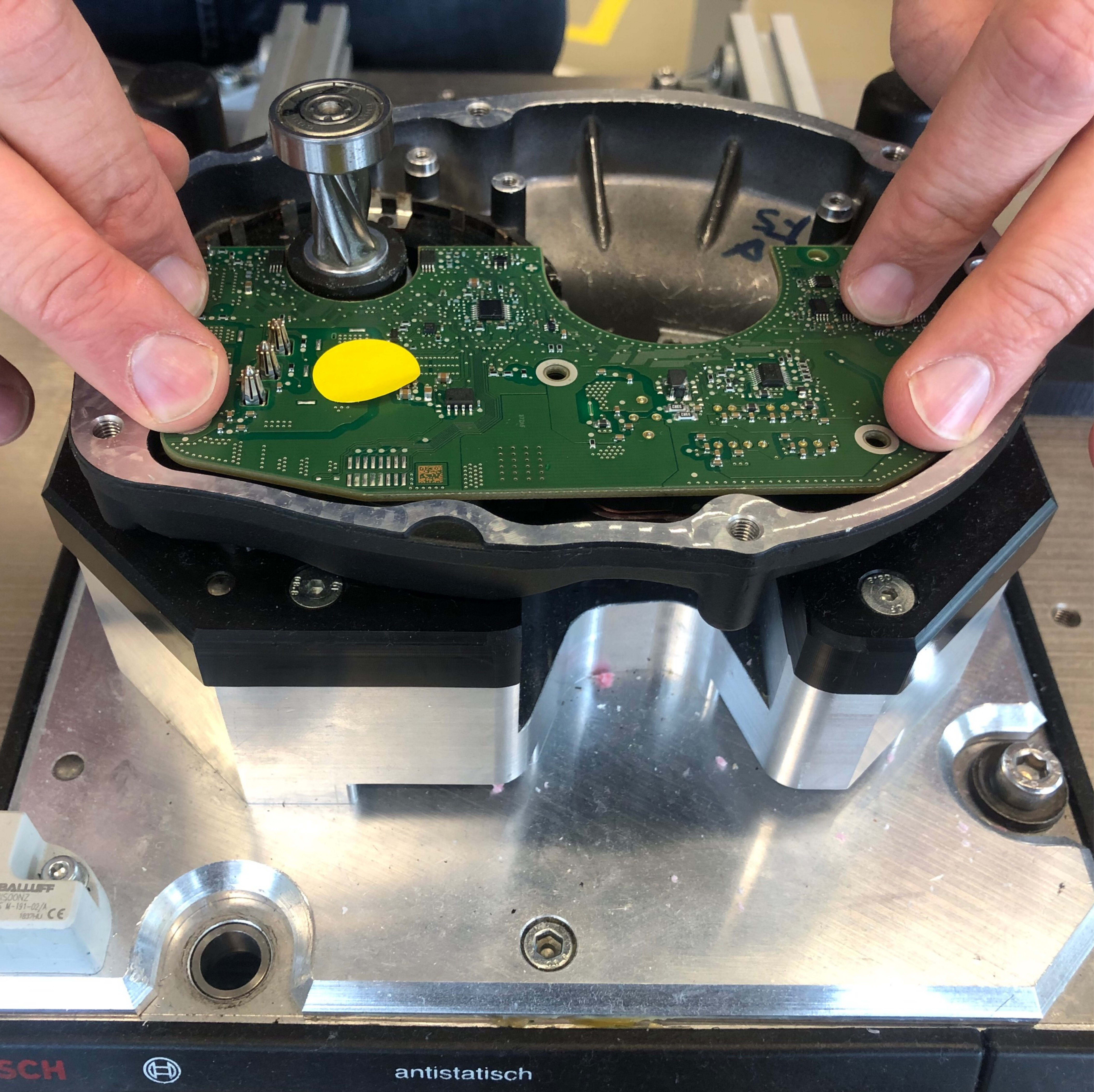}
		\includegraphics[width=0.323\linewidth]{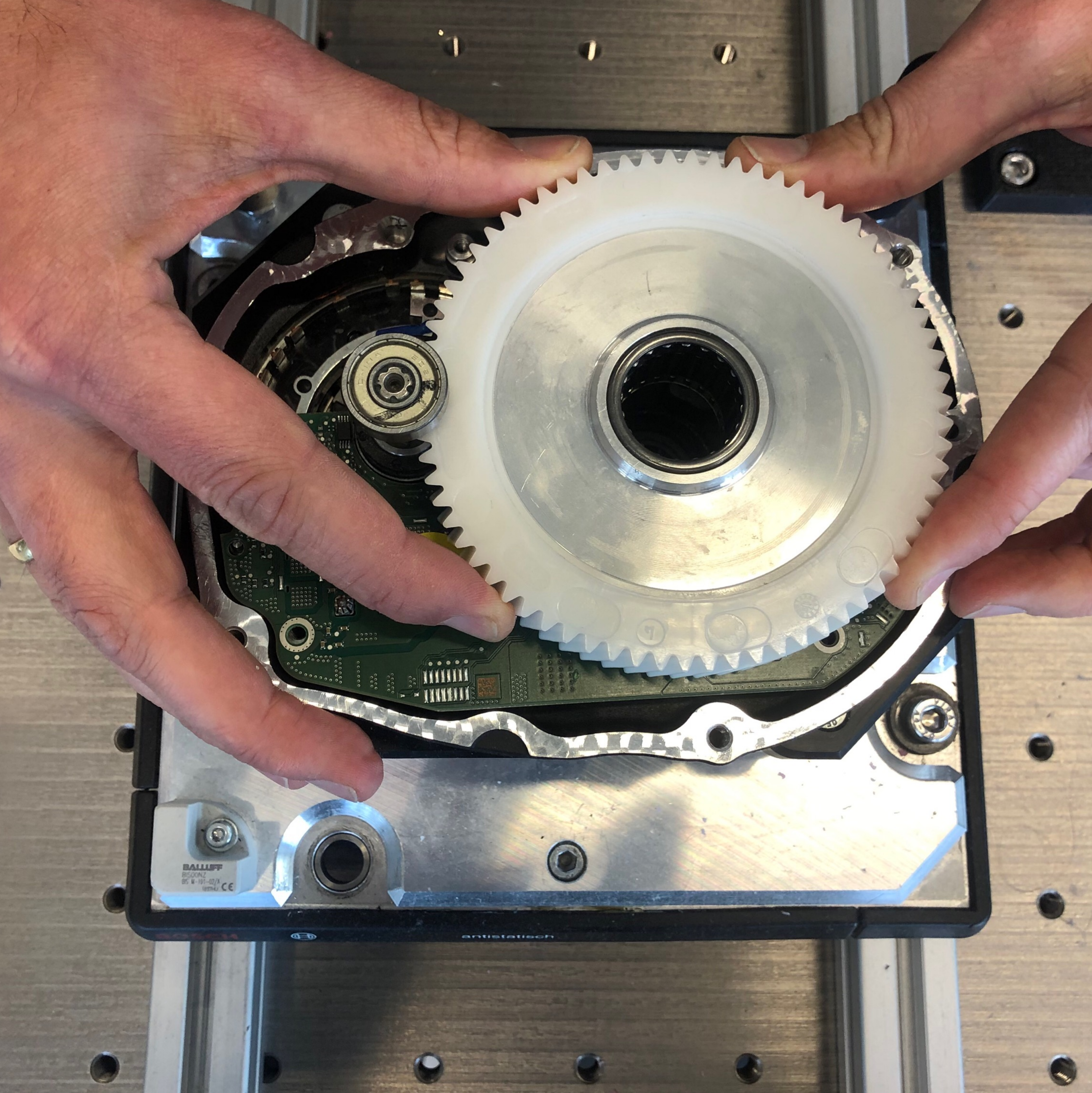}
		\includegraphics[width=0.323\linewidth]{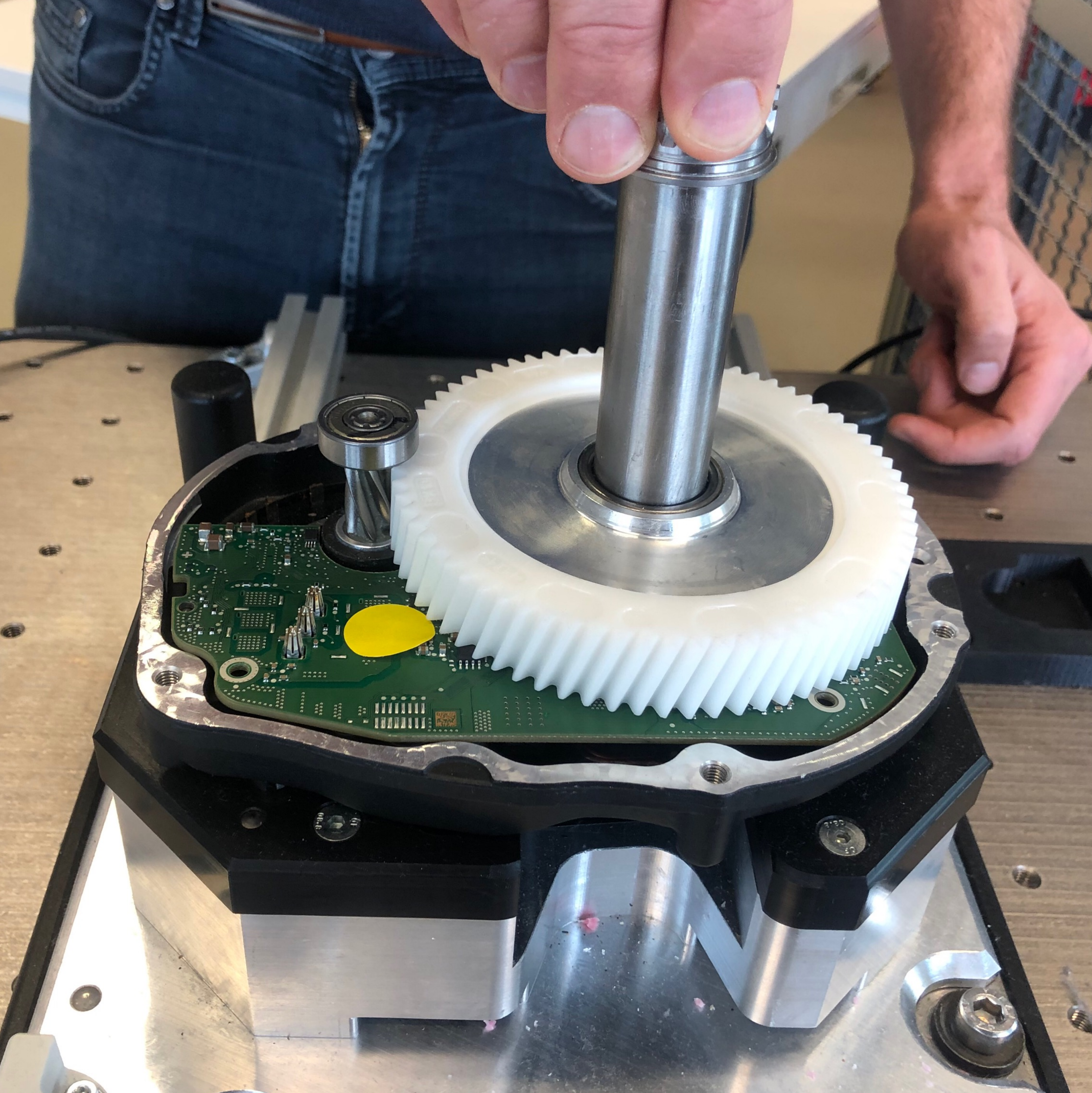}
		\caption{Snapshots of the manual e-Bike assembly process.}
		\label{subfig:ManualAssembly}
	\end{subfigure}
	\caption{\textbf{(a)} \emph{Left:} Robot workspace, assembly area, and main e-Bike assembly components; \emph{Right}: Top view of the initial (top) and final (bottom) stages of the assembled e-Bike motor. Snapshots of the assembly process can be seen in Fig.~\ref{fig:assembly-snapshots}. \textbf{(b)} \emph{Left:} Manual PCB placing; \emph{Middle:} Manual gear placing; \emph{Right:} Manual peg insertion.} 
	\label{fig:teaser}
\end{figure}

In recent years, end-to-end learning methods were investigated as a complementary approach to explicit programming for significantly reducing the amount of manual programming work.
However, by today, these methods still need expert developers to work safely and reliably~\cite{amodei2016concrete,henderson2018deep}.
Also, the additional time and effort to collect training data as well as the lack of explainability when handling unforeseen conditions further limit ease of use in practical FMS.

A compromise between explicit programming and end-to-end learning methods is the paradigm of learning from expert demonstrations (LfD)~\cite{Billard16}.
This paradigm relies on rich and intuitive human directions for defining motion skills while still using learning models to abstract the motion patterns and generalize them from a small set of demonstrations.
In the context of this paper, a \emph{motion skill} is understood as a model that outputs a task-space reference trajectory -- that encapsulates the main trend of the demonstrations -- for an underlying tracking controller, similarly to the definition in~\cite{Billard16}. 

Regardless of the motion generation strategy, another challenge is to design controllers that able to accurately track the desired trajectories associated to a motion skill. 
The tracking precision of these controllers is often compromised by accumulated modeling errors, demanding to leverage model identification or learning methods~\cite{Hollerbach16:ModelID,Reuss22:End2EndInvDyn}. 
While these approaches enhance control precision, inaccuracy may still arise from other parts of the integrated system, for example from inaccurate pose estimation of the objects of interest in the robot workspace. 

This creates the need for endowing the robot with adaptation capabilities aimed at refining the skills or controllers to perform the task satisfactorily. 
In this context, sample efficiency is critical as skill or control adaptation may entail real-world interactions, and therefore the fewer samples are required for adaptation, the safer and less costly the process is. 
These requirements make black-box optimizers a promising approach due to their simplicity and data efficiency~\cite{Chatzilygeroudis20:SurveyPS}. 

Combining learned skills with dynamic perception of objects to determine their poses, as well as optimization techniques to automatically improve configuration choices, provides a powerful framework for robotic manipulation in FMS, as considered in this paper.
We show that the learned skills can be sequenced according to a human-designed task plan to allow robots to perform industrial tasks of varying complexity.
Furthermore, we show how the different system components support an overall adaptability of task execution and explainability of the actions performed by the robot.
In the following, we explain how we approach the foregoing challenges around a real-world industrial task: the assembly of an e-Bike motor.

\subsection{The e-Bike Motor Assembly Challenge}
\label{subsec:IntroEbike}
Our work is guided by a real-world challenge that concerns the free assembly of an e-Bike motor, which is shown in Figs.~\ref{fig:teaser},~\ref{fig:assembly-snapshots} and~\ref{fig:shaft_insertion}. 
Bikes that are supported by an electric motor (e-Bikes) are modern means of transportation and are rapidly adopted across the world. 
Solely in Europe more than \SI{5}{\million} e-Bikes were sold in 2021, and the demand is continuously growing. Some people expect that in few years one out of two bikes sold in Europe will be equipped with an electric motor.
The e-Bike motor considered here is a Bosch Performance Line motor, a mid-motor for pedal-assisted bicycles. 
It consists of multiple components, including: \emph{(i)} a PCB board with the electronic control unit, \emph{(ii)} gears for the transmission, \emph{(iii)} a shaft that connects the motor to the pedals, and \emph{(iv)} a peg that is placed around the shaft as part of the transmission system (see Fig.~\ref{fig:teaser}).

Industrial production for e-Bike motors is faced with many challenges. 
Due to the high demand, high volumes of e-Bike motors must be produced. 
The quality requirements on the motors are extremely high, as many people rely on them in their daily usage. 
In addition, e-Bike motors are demanded with a high variance -- e.g., Bosch is currently offering seven variants of e-Bike motors -- and the products are continuously improved with new variants regularly coming to market. 
Therefore, robot-based flexible manufacturing systems promise to enable flexible, high-quality, and high-volume production lines for this and other products with similar manufacturing requirements.  

Due to the social and economical impact that this product may have in the near future and its challenging manufacturing requirements, we chose this use case to analyze and propose a potential solution to the technical challenges that a robotic manipulator may face when required to assemble a subset of the e-Bike motor parts.
To do so, we designed an e-Bike assembly workstation consisting of a loading platform where some e-Bike motor components are picked, and the assembly station where the components are assembled together (see Fig.~\ref{subfig:AssemblySetup}). 
The whole e-Bike motor is roughly $165\si{\mm} \times 146\si{\mm} \times 150\si{\mm}$ in size. 
A pre-defined sequence of sub-tasks should be followed during the whole assembly process. 
Particularly, we focus on the following four sub-tasks:

\subsubsection{[Press-PCB]} 
A PCB of size $156\si{\mm} \times 75\si{\mm} \times 2\si{\mm}$ is pressed into three pins on the motor base to secure the board (see Fig.~\ref{subfig:ManualAssembly}).
We assume here that in a previous (possibly manual) assembly step the PCB board has been placed appropriately and only the final insertion is carried out by the robot. 
Then, this sub-task consists of four stages: approaching, pressing, transition (between pins location), pressing and retreating as shown in Fig.~\ref{fig:press-pcb-execute}. 
The pressing should be done with an appropriate force, if too small then the pins are not securely inserted; if too large then the PCB may be damaged; 
while the transition should be accurate to not miss the pins (with less than $\SI{2}{\mm}$ tolerance).  

\subsubsection{[Mount-Gear]}
A spur gear of size $115\si{\mm} \times 115\si{\mm} \times 37\si{\mm}$ is mounted above the rotor casing (see Fig.~\ref{subfig:ManualAssembly}).
The gear must be gently slid into the correct position on the motor. 
Therefore, the robot must place the gear on the motor, gently slide it to the right position and finally lightly push it to secure it
in-place. 
Thus, this sub-task consists of three stages: approaching, sliding and pushing. 
The sliding stage is quite delicate as the metal bottom of the gear needs to follow a tunnel on the casing into the desired location. 

\subsubsection{[Insert-Shaft]} 
A drive shaft of size $16\si{\mm} \times 16\si{\mm} \times 150\si{\mm}$ is inserted through the opening of the gear into a hole in the metal casing. 
This step requires to place the shaft properly on the motor, to gently wiggle it for finding the right positioning, and finally to push it into the casing. 
This sub-task consists of three stages: approaching, wiggling and pushing. 
The ``wiggling'' stage is to insert the bottom of the shaft into the hole beneath the gear, which has a tolerance of around $1 \si{\milli\meter}$. 
Furthermore, the shaft is pushed with a \emph{large} force in a \emph{stiff} manner to ``click'' in position during the ``pushing'' stage. 
This proper placement is vital for the functionality. 

\subsubsection{[Slide-Peg]}
The peg of size $24\si{\mm} \times 24\si{\mm} \times 102\si{\mm}$ is slid \emph{between} the inserted shaft and the mounted gear (see Fig.~\ref{subfig:ManualAssembly}).
Inserting this peg requires to slide it along the previously inserted shaft. 
The peg bottom must reach the top of the shaft with a tolerance of about $1 \si{\mm}$.  
Once properly placed, the peg has to be slid down the shaft and must be twisted such that the inline tooth of the shaft and the peg match. 
While executing this twisting, the peg should be pushed with a \emph{small} force. 
Therefore, this last sub-task consists of four stages: attaching, sliding, twisting and pushing. 

It is worth mentioning that the shaft and peg are placed freely on the loading platform, which means that we do not assume previously-grasped objects, as commonly done in several works researching insertion tasks. 
Moreover, our robot is not equipped with dedicated grippers for the aforementioned e-Bike components. 
Furthermore, due to the subsequent insertion and pushing tasks, they should be re-oriented from lying-flat to standing into a fixture, 
such that they can be grasped from the top, as shown in Fig.~\ref{fig:assembly-snapshots}.

\subsection{Approach}
In this paper, we propose a \emph{system approach} to program a robot manipulator \emph{on-site by demonstration} to perform flexible automation tasks. 
The presented approach is designed to serve the requirements identified in case studies of the aforementioned e-Bike motor assembly as well as the dense box-packing in logistics presented in~\cref{sec:experiments}. 
Both case studies require a robot to grasp items using vision sensors and to place them skillfully according to the task plan. 
Despite this common theme, these case studies offer a broad variety of more subtle challenges on grasping objects and skillful placing. 
Our intention is to provide here a comprehensive overview on the methods and the system integration that proved successful in these industrial case studies. 
Parts of the methods have been published previously in other works.
The contribution of the present paper is to describe how the individual methods are integrated into a system and how this system creates a powerful and flexible automation solution.
We present two real use-cases from manufacturing and logistics and report on the performance and limitations of the system in these settings. 
Finally, we discuss if the state of the art in robot manipulation, exemplified by our methods, is ready for flexible manufacturing on the shop floor and provide directions for future research.
To the best of our knowledge, our integrated learning-based system for advanced manipulation as presented in this paper is unique in its completeness, technical timeliness and practical deployment on two real use-cases both with a significant economic background.

In high-level terms, our integration framework consists of \emph{(i)} a vision perception system that estimates poses of objects of interest for the task (see~\cref{subsec:DON});
\emph{(ii)} a learning component that builds an object-centric skill model using the vision information along with the demonstration data -- generated by a human operator who kinesthetically teaches the robot the desired movements for a particular skill (as explained in~\cref{subsec:LfD});
\emph{(iii)} a skills sequencing module which builds a complete task model, out of a set of motion skills, according to a human-designed task plan (see~\cref{subsec:sequencing});
\emph{(iv)} a task execution component that reproduces the full task, where the motion skills adapt according to the perceived objects state (see~\cref{subsec:reproduction}); 
and lastly, \emph{(v)} a task optimization component which refines motion skills as needed (as detailed in~\cref{subsec:skills_refinement}). 
A high-level introduction to each of our framework components is given in~\cref{sec:system}, while the corresponding technical details are provided in~\ref{sec:appendix}. 

\begin{figure*}[t]
	\centering 
	\includegraphics[width=0.98\linewidth]{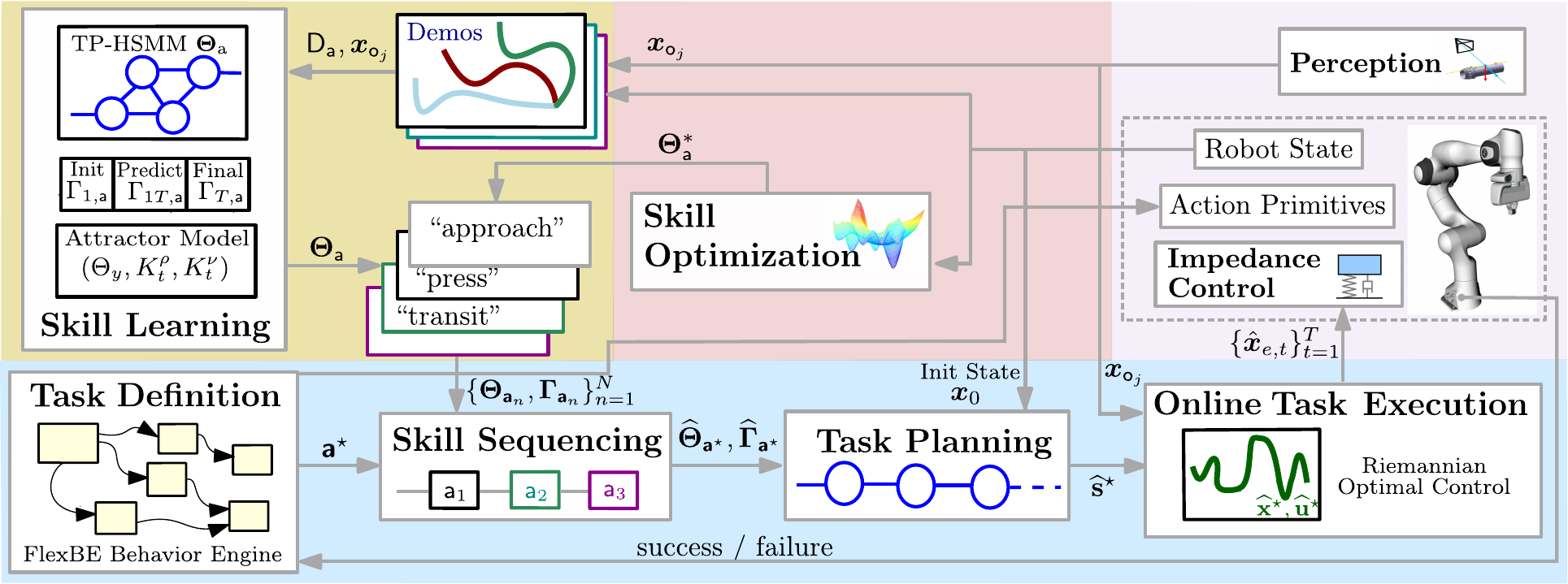}
	\caption{
		Overview of our system structure with its main modules: Perception, low-level impedance control, skill learning and sequencing, task planning, task online execution, and skill optimization (see Table~\ref{tab:blocks_notation} for a definition of the module inputs and outputs).
		The yellow, blue, red, and purple blocks respectively enclose the modules involved into the learning, task execution, skill optimization, and hardware. 
		Using kinesthetic demonstrations we encode object-centric skills into task-parametrized hidden semi-Markov models (TP-HSMMs), 
		learn the condition models, and estimate the attractor model if demonstrated force profiles should be tracked. 
		Given a high-level task plan, a desired sequence of states is determined and later combined with online perception data to compute the optimal state sequence for the task model, which serves to estimate the reference trajectory tracked by our Riemannian optimal controller. 
		The skill optimization is triggered whenever a learned skill requires small refinements for a successful execution. 
		Our behavior engine encapsulates the task plan, including action primitives, learned skills, skill refinement and failure recovery behaviors.}
	\label{fig:overview}
\end{figure*}

\subsection{Related Work}
The goal of this section is to provide a broad overview on recent developments of robotic manipulation frameworks and to briefly highlight relevant works on each of the subfields that our interdisciplinary framework builds on.
 
\paragraph{Robotic manipulation frameworks}
While most of the research agenda on robotic manipulation has focused on solving domain-specific challenges, there exist recent efforts on integrating full manipulation systems in real-world settings~\cite{Pozzi21:Editorial}.
The optimization-based control framework introduced in~\cite{Gold20:MPCframework} decomposes manipulation tasks into skills and manipulation primitives, which resembles our task abstraction. 
Their experimental work is also close to ours as their framework allows for free-space motions and contact-rich skills, which are also part of the case studies considered in this paper. 
However, Gold \etal~\cite{Gold20:MPCframework} did not consider the integration of vision perception and skill optimization in their approach. 
In a similar vein, Johannsmeier~\etal~\citep{johannsmeier2019framework} proposed a manipulation-skills framework, where each skill was represented by a simple directed graph, and subsequently a full manipulation task corresponded to a larger graph composed of sequenced skills.
Unlike our framework, their work builds on hand-coded parametrized skills. 
However, the parametric nature of these skills is leveraged to optimize their performance for different manipulation tasks using black-box optimizers, similarly to our skill optimization approach.   

In contrast to our approach and that of Gold \etal~\cite{Gold20:MPCframework}, deep learning-based approaches have recently been employed to build manipulation frameworks. 
On one side of the solutions spectrum, we may consider imitation learning as the mechanism to train end-to-end manipulation frameworks as in~\cite{Jung21:ILframework}. 
On the other end, we can find solutions leveraging reinforcement learning to overcome the need of labeled training data at the cost of longer training times~\cite{zhan2021a:ILRLframework} or to improve generalization of imitation learning-based frameworks~\cite{Singh19:RLframework}.
Most of these deep learning-based frameworks do not impose a specific task abstraction but instead formulate the manipulation problem as learning the correct control commands for a large set of task states.
Also, these framework overcome the need of object pose estimation as the task state is often described at pixel levels. 
However, these frameworks are often not designed to solve long-horizon tasks as the ones considered in this paper. 
Moreover, their lack of explainability limits its use in real-world industrial settings.    

\paragraph{Learning from demonstration (LfD)}
Learning motions by kinesthetic teaching is an intuitive and natural way to transfer human skills to robots~\cite{Billard16}.
Specifically, task-parameterized Gaussian models, reviewed by Calinon~\cite{calinon2016tutorial}, consider an object-centric formulation of motion skills by incorporating observations from the perspective of different coordinate systems.
This enables the adaptation of learned skills to new object configurations in weakly-structured and dynamic environments. 
This object-centric formulation is indeed exploited in our framework to build adaptable skill models, which favors their easy reusability. 

\paragraph{Skills sequencing}
In complex settings where the robot needs to execute multiple actions, a key limitation of LfD arises: the need to collect demonstrations of the whole task that is to be learned.
To address this, recent work focuses on learning not only complete robot motions for a task but also sequences of re-usable motion skills.
For instance, Lioutikov \etal\citep{lioutikov2016learning} allowed sequencing of simple motion skills encoded by dynamic movement primitives for complex tasks over bimanual manipulations.
The approach proposed by Manschitz \etal\cite{manschitz2014learning} learned a sequence graph of skills within a movement library from kinesthetic demonstrations, where the classification of different skills was carried out via Gaussian mixture models.
In Konidaris \etal\cite{konidaris2012robot}, the skill sequencing problem was investigated from a hierarchical reinforcement learning perspective, where each skill is represented by a general control policy.
Most of the above approaches still required human demonstrations at the complete-task level.  
In contrast, we introduce an approach that requires only demonstrations at the level of individual motion skills, which are automatically sequenced given the task plan, goal and initial system conditions. 
This approach is beneficial when different tasks share a subset of these skills, which can be reused to avoid teaching them every time a new task is defined. 

\paragraph{System integration}
Finally, a key challenge for integrating learned motions is their interplay with the robot hardware and other software components~\cite{rajan2017towards}, in particular for adaptation and recovery from failure cases.
Abstraction layers between the learning system and the hardware, as for example used by De Coninck \etal\cite{de2020learning}, are a common approach in this regard.
In our system, we use ROS~\cite{ros} and FlexBE~\cite{flexbe} to implement a software structure of exchangeable components following common software standards~\cite{malavolta2020you}, where the learned motion skills are one option for computing controller reference trajectories.
This focus on components rather than strict layers allows for improved adaptation, e.g., by closing the perception-action loop also at lower layers.

	\section{Proposed System}
	\label{sec:system}
	Despite the heterogeneity of the challenges arising in the e-Bike assembly setting, described in Section~\ref{subsec:IntroEbike}, there are several similarities across the required robot skills. 
Most of them require, in some form, to pick an object, to move it to a target, and to execute a force-sensitive insertion. However, despite this common theme, the required realization of the skills is quite diverse and specific for each assembly step.
Furthermore, the initial object and target configurations vary along the course of the assembly process, and therefore the robot must adapt online
accordingly. 
Taken together, these requirements amount to a significant complexity that must be managed during the robot programming process.

\renewcommand{\arraystretch}{1.4}
\begin{table*}[]
	\footnotesize
	\centering
	\begin{tabularx}{\textwidth}{|l|X|l|}
		\rowcolor{lightgray}
		\hline
		\textbf{Module} & \textbf{Inputs} & \textbf{Outputs} \\
		\hline
		Perception & RGB image $\bm{I} \in \mathbb{R}^{W \times H \times 3}_{\tiny{+}}$ & Object pose ${\vec{x}_{\mathsf{o}_j}\in \mathbb{R}^3 \times \mathcal{S}^3}$\\
		\rowcolor{lightestGray}
		Skill learning & Demonstrations set $\mathsf{D}_{\mathsf{a}}$ for skill $\mathsf{a}$ and object pose ${\vec{x}_{\mathsf{o}_j}}$ & Learned model parameters $\bm{\Theta}_\mathsf{a}$ \\
		Skills sequencing & Skills sequence $\boldsymbol{\mathsf{a}}^\star=(\mathsf{a}_1, \cdots, \mathsf{a}_N)$, associated models parameters $\mathbf{\Theta}_{\mathsf{a}_n}$ and condition models $\boldsymbol{\Gamma}_{\mathsf{a}_n} \forall \; 1 \leq n \leq N$ & Full task model parameters $\widehat{\mathbf{\Theta}}_{\boldsymbol{\mathsf{a}}^\star}$ and condition model $\widehat{\boldsymbol{\Gamma}}_{\boldsymbol{\mathsf{a}}^\star}$ \\
		\rowcolor{lightestGray}
		Task planning & Task model parameters $\widehat{\mathbf{\Theta}}_{\boldsymbol{\mathsf{a}}^\star}$, initial end-effector pose $\bm{x}_{\mathsf{e},0}$, initial objects pose ${\vec{x}_{\mathsf{o}_j,0}}$, and task desired goal $\bm{x}_{\mathsf{G}}$ & Most-likely state sequence $\widehat{\bm{s}}^\star$ \\  
		Task online execution & State sequence $\widehat{\bm{s}}^\star$, objects pose ${\vec{x}_{\mathsf{o}_j,t}}$ and end-effector pose $\bm{x}_{\mathsf{e},t}$ & Reference end-effector trajectory $\{\hat{\bm{x}}_{\mathsf{e},t}\}_{t=1}^T \in \mathbb{R}^3 \times \mathcal{S}^3$ \\
		\rowcolor{lightestGray}
		Impedance controller & Reference end-effector trajectory $\{\hat{\bm{x}}_{\mathsf{e},t}\}_{t=1}^T$ & Control torque inputs $\bm{\tau}_u$ \\
		Skill optimization & Learned model parameters $\bm{\Theta}_\mathsf{a}$ for skill $\mathsf{a}$ & Optimized model parameters $\bm{\Theta}^{*}_\mathsf{a}$\\
		\hline
	\end{tabularx}
	\caption{Notation of the high-level information flow of our manipulation framework illustrated in Fig.~\ref{fig:overview}.}
	\label{tab:blocks_notation}
\end{table*}
\renewcommand{\arraystretch}{1}

Different machine learning techniques contribute substantially to managing the complexity of this process. 
While major research efforts have been dedicated to using machine learning without much structure for robotics, we advocate here a highly structured and compositional approach, similar to recent works~\cite{johannsmeier2019framework,Gold20:MPCframework}. 
Guided by the e-Bike challenges described previously, we argue that it remains useful to keep a robotics control system architecture, but extensively make use of machine learning approaches within it.
Robot control, learning, planning, adaptation, and software integration are essential components in our robotic manipulation architecture.
Here we provide a high-level technical description of our control framework, of the models used for skills learning, sequencing, and synthesis, as well as of the task online execution.
We also describe how to enhance the robot performance via task-driven skills optimization.
Finally, we show how the control, learning and planning models are interfaced from a software-engineering perspective, which is imperative to make robotic manipulation a tangible technology in industry.
Figure~\ref{fig:overview} provides an overview diagram showing the main modules of our framework.
We provide low-level technical details for each of the framework components in~\ref{sec:appendix} for the interested reader.

\subsection{Problem description}
Firstly, let us provide a more formal description of the robotic manipulation problems we tackle here.
Let us consider a multi-DoF robotic arm, whose end-effector has state $\vec{x}_{\mathsf{e}} \in \mathbb{R}^3 \times \mathcal{S}^3$ (describing the Cartesian position in Euclidean space and orientation quaternion in the $3$-sphere), that operates within a weakly-structured and known workspace.
Also, we assume that in the robot workspace there are objects of interest denoted by ${\mathsf{O}=\{\mathsf{o}_1, \mathsf{o}_2,\cdots,\mathsf{o}_J\}}$, each of which has state ${\vec{x}_{\mathsf{o}_j}\in \mathbb{R}^3 \times \mathcal{S}^3}$, provided by a vision system.
For simplicity, the overall system state is denoted by ${\vec{x}_{\mathsf{s}}=\{{\vec{x}_{\mathsf{e}}}, \{\vec{x}_{\mathsf{o}_j}, \forall \mathsf{o}_j\in \mathsf{O}\}\}}$.
Within this setup, an operator performs several kinesthetic demonstrations on the arm to manipulate one or several objects for certain manipulation skills.
Denote by ${\mathsf{A}=\{\mathsf{a}_1,\mathsf{a}_2,\cdots,\mathsf{a}_H\}}$ the set of demonstrated skills, $\mathsf{O}_{\mathsf{a}}$ the set of objects involved in $\mathsf{a}$, and $\mathsf{D}_{\mathsf{a}}$ the set of available demonstrations.
All demonstrations follow an object-centric structure, i.e., they are recorded from multiple coordinate systems associated to the objects in $\mathsf{O}_{\mathsf{a}}$, which often represent the objects pose in the workspace (see Fig.~\ref{fig:tphsmm} for an illustration).
For example, the skill \textquote{insert the peg in the cylinder} involves the objects \textquote{peg} and \textquote{cylinder}, and the associated demonstration data are recorded from the robot, the \textquote{peg} and the \textquote{cylinder} coordinate systems.

As described in the e-Bike assembly challenge, the manipulation tasks we consider consist of a \emph{given} sequence of skills, chosen from the demonstrated skills set $\mathsf{A}$, and pre-defined \emph{action primitives}\footnote{Action primitives refer to simple commands that can be easily integrated into the workflow and that are necessary for the execution of the task.} (e.g. gripper commands).
For example, consider an insertion task involving \textquote{locate the cap, pick the cap, close gripper, re-orient the cap, open gripper, pick the cap again, close gripper, insert the cap, and open gripper}.
In this case, \textquote{locate the cap}, \textquote{close gripper} and \textquote{open gripper} are pre-defined \emph{action primitives} and involve, e.g., $6$D pose estimation, while the rest corresponds to learned \emph{motion skills}.
Given a sequence of skills~$\boldsymbol{\mathsf{a}}^\star$, our framework computes the corresponding task plan and subsequently retrieves the desired end-effector pose trajectory $\{\hat{\bm{x}}_{\mathsf{e},t}\}_{t=1}^T$ that is then tracked by a Cartesian impedance controller to fulfill the task at hand.
Finally, the task execution may be improved to overcome possible failures or address new task requirements via data-efficient task optimizers.
Next, we provide high-level technical descriptions for each of the components involved to solve the aforementioned robotic manipulation challenges.

\subsection{Robot Perception for Objects Pose Estimation}
\label{subsec:DON}
In all case studies, perceiving the environment is essential. 
All our tasks require to detect objects and to precisely obtain a grasp pose.
The set of objects we aim to detect and grasp in our setting is fairly broad. 
Some objects are characterized by intense colors and textures (e.g. yellow boxes in the packing setting described in Section~\ref{sec:experiments}), while other objects are metallic, reflective and textureless (e.g. the peg in the e-Bike motor assembly). 
We emphasize here that many established methods for $6$D pose estimation are notoriously labor intensive in terms of data labeling. 
Given the heterogeneity of our objects portfolio, such a data labeling effort is prohibitive and will not be accepted.
Rather, we explore methods that do not require intensive human labeling, but rather generate the labeled data autonomously.

\begin{figure*}[t]
	\centering
	\includegraphics[width=.24\linewidth]{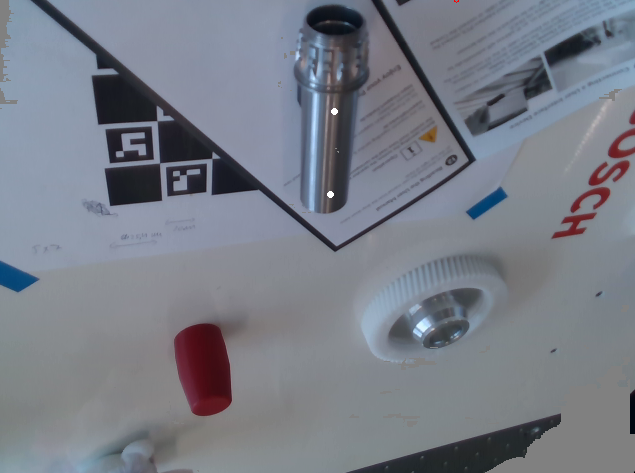}
	\includegraphics[width=.24\linewidth]{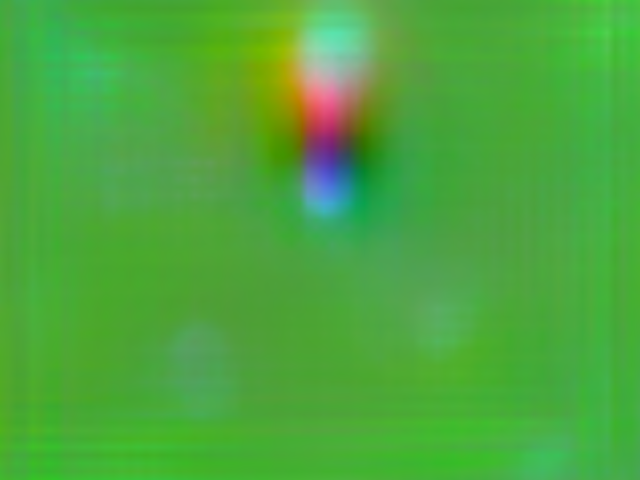}		
	\includegraphics[width=.24\linewidth]{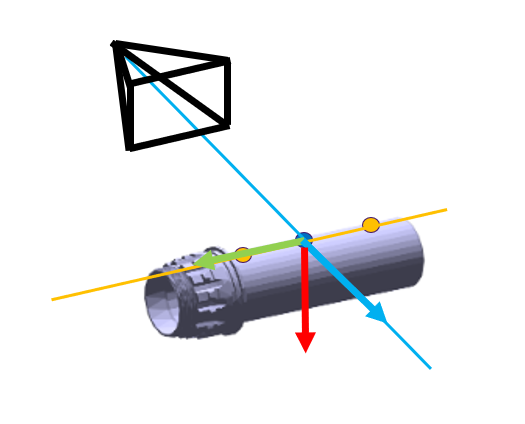}
	\includegraphics[width=.24\linewidth]{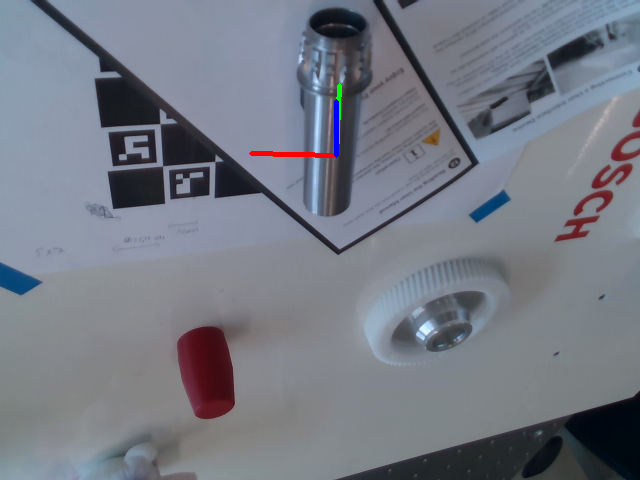}
	\caption{\textbf{Left:} the RGB image depicting the metallic shaft with two manually annotated keypoints, and the evaluated descriptor space image. Note that for simplicity we use a $3$-dimensional descriptor space, such that the descriptor values of pixels can be associated with RGB colors.
		\textbf{Right:} the simplified pose estimation approach and the pose estimation result using the annotated keypoints and the current camera view.}
	\label{fig:don_pose}
	\vspace{-0.4cm}
\end{figure*}

Specifically, we leverage \emph{Dense Object Nets} (DONs)~\cite{Florence2018,kupcsik2021supervised}, which provide dense pixel-wise descriptors $\vec I_{\mathsf{DON}}~\in~\mathbb{R}^{W\times H\times D}_{\tiny{+}}$ of objects from RGB images $\vec I \in\mathbb{R}^{W \times H \times 3}_{\tiny{+}}$ (see Fig.~\ref{fig:don_pose}--left). 
The dimension $D\in \mathbb{N}^+$ represents a user-defined number that controls the resolution of the descriptor space.
The descriptor image $\vec I_{\mathsf{DON}}$ serves as an object-specific keypoint detector that can be used in a variety of downstream tasks, such as controller learning~\cite{Manuelli2020}, rope manipulation~\cite{Sundaresan2020}, and grasp pose prediction~\cite{Florence2018, kupcsik2021supervised}, among others.

While computing first an intermediate, descriptor-space object representation might seem unintuitive, DONs have several beneficial properties that makes them particularly appealing for industrial settings.
They can be trained fully self-supervised without human annotation from registered RGB-D video streams recorded, e.g, by a robot with a wrist-mounted camera. 
DONs can be trained and deployed within hours for a set of new objects.
Finally, DONs work well under changing lighting conditions, partial occlusions, and provide dense features even for non-textured and metallic objects.

In our work we use the dense descriptor representation to estimate $6$D object poses.
We collect a dataset with objects that we wish to learn using a robot and a wrist-mounted camera.
After training the DON, we define a set of object-specific keypoints for every object.
Note that in case we train the network with multiple object classes, the keypoint sets will be clearly separated in descriptor space without overlap.
Thus, the keypoint sets will be unambiguous and unique to each object class.
During inference, for every frame we detect these sets of keypoints in image plane, unproject them to world coordinates and extract $6$D poses.
For the detailed description of our inference pipeline we refer to~\ref{app:6d_don_pose} and our previous work~\cite{kupcsik2021supervised}.

\subsection{Robot Impedance Control}
\label{subsec:control}
All our industrial tasks require the robot to work in contact with the environment, while being precise (e.g. apply
right forces for insertion) and safe (e.g. not breaking anything).  
From a control perspective, this means that the kinematic task-space reference quantities $\hat{\bm{x}}_{\mathsf{e},t}$, which are produced by our skills, need to be tracked fast, accurately and in a compliant manner to avoid large interaction forces. 
As we have full access to the robot system itself, we advocate here to rely on a model-based control architecture which has proven to be efficient for a long time.
 
Specifically, we employ a Cartesian impedance controller that is implemented via a classical computed-torque scheme~\cite{Sici09}. 
We formulate the controller in $\operatorname{SE}(3)$, the Euclidean group of rigid body motions. 
This allows a proper treatment of the tracking errors based in the Lie algebra, which avoids approximations introduced by representing orientations via, e.g. Euler angles~\cite{Sici09}. 
Note that the controller performance heavily depends on the quality of the underlying robot dynamics model, which is often unavailable.   
Hence, we leverage learning techniques to come up with a more precise robot model to improve control tracking accuracy.

\subsubsection{Cartesian Impedance Control in $\operatorname{SE}(3)$}
\label{subsubsec:CIC}
As pointed out previously, several of the robot skills considered in our use cases involve contact with the environment. 
To control a variety of required contact forces across skills, we resort to indirect force control~\cite{Villani2016}, which has recently gained interest due to the rise of torque-controlled robots, for which impedance control is a standard strategy. 
Impedance controllers provide a compliant behavior in all phases of a contact task (e.g, non-contact, transition and contact) but are limited in their force tracking ability, mainly from partial knowledge of the environment. 
To cope with this limitation, we adopt two distinct methodologies: impedance and set-point adaptation. 
Impedance adaptation adjusts the controller parameters (e.g., inertia, damping, and stiffness) to improve tracking in response to force, position, or velocity measurements. 
Set-point adaptation improves force tracking by adjusting the controller set-point (e.g., the reference position) based on force tracking errors or on estimations of the environment stiffness variations.
Both adaptation strategies are tackled from a learning perspective as described in Section~\ref{subsec:LfD}. 

Our control framework employs a classical acceleration-resolved Cartesian impedance scheme~\cite{Ott2008}, which includes an optional feedforward term that can be designed, for example, to maintain downward pressure in insertion tasks. 
As we work with a $7$-DoF robot, we also add a nullspace joint-space acceleration to resolve redundancy by tracking a reference joint configuration.
Moreover, we selectively use an integral action on the error term to enforce high-accuracy along specific directions of free-space motion skills (see~\ref{app:CIC} for technical details). 
Note that for the task-space directions which are absent of interaction forces, the postulated impedance control law is equivalent to an inverse dynamics position control scheme~\cite{Sici09}. 
Therefore, the more compliant the controller is, the more the closed-loop behavior degrades due to disturbances.
Consequently, we place high importance on accurate modeling of the manipulator dynamics as described next. 

\begin{figure*}[t]
	\centering
	\begin{subfigure}[b]{0.24\textwidth}
		\includegraphics[width=0.99\textwidth]{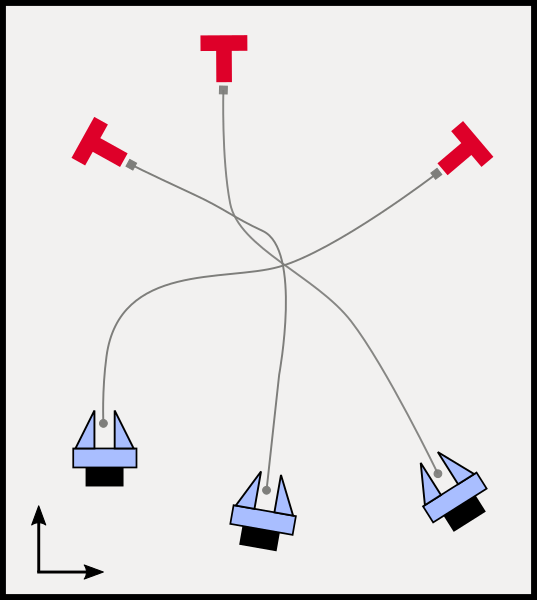}
		\caption{Demonstrations}
	\end{subfigure}
	\begin{subfigure}[b]{0.24\textwidth}
		\includegraphics[width=0.99\textwidth]{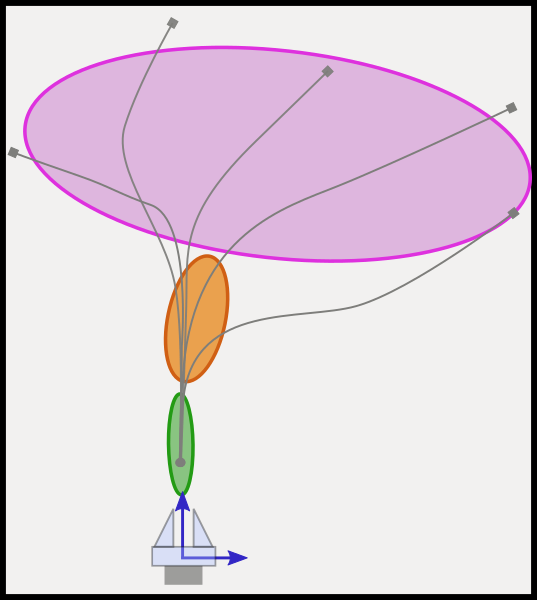}
		\caption{End-effector local model}
	\end{subfigure}
	\begin{subfigure}[b]{0.24\textwidth}
		\includegraphics[width=0.99\textwidth]{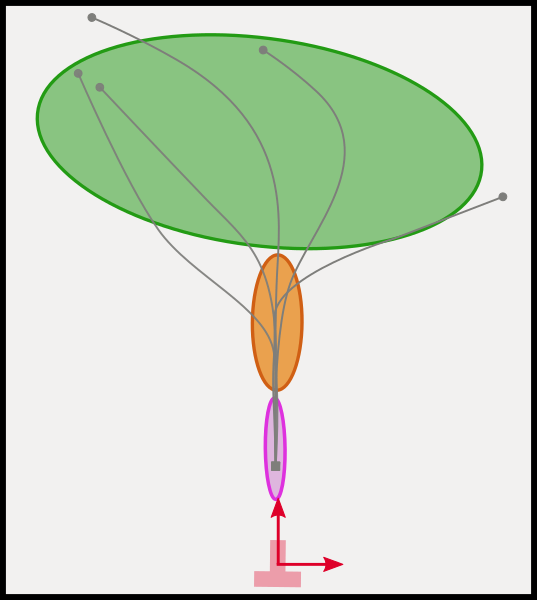}
		\caption{Peg local model}
	\end{subfigure}
	\begin{subfigure}[b]{0.24\textwidth}
		\includegraphics[width=0.99\textwidth]{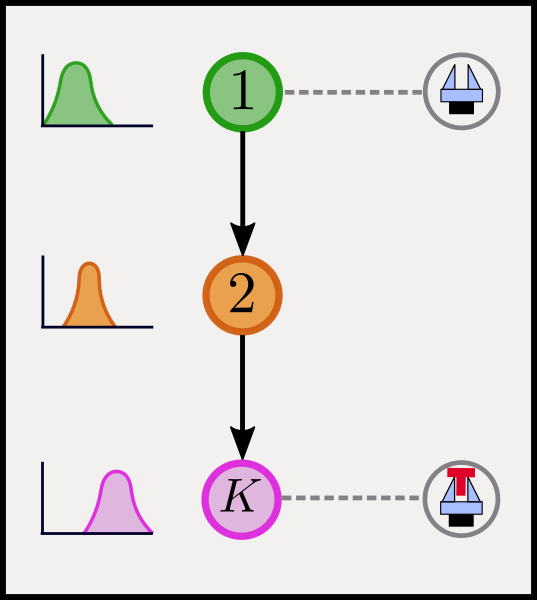}
		\caption{HSMM transition graph}
		\label{subfig:hsmm}
	\end{subfigure}
	\caption{Illustration of object-centric demonstrations of a reaching task encoded by a TP-HSMM. \textbf{(a)} End-effector trajectories (\graydemo) are recorded from the perspective of the initial end-effector and red peg poses (\textbf{(b)}-\textbf{(c)}, respectively). High-precision patterns are modeled by narrow Gaussians, in different coordinate systems, at the beginning (green ellipse \greenellipse in \textbf{(b)}) and the end (light purple ellipse \pinkellipse in \textbf{(c)}) of the demonstrations. \textbf{(d)} The temporal evolution of the skill is represented by the transition graph and duration distributions associated to each Gaussian component. Note that in this example, each TP-HSMM state is represented by a single Gaussian. The first state (\greencircle) encodes the start phase of the reaching task, the second state (\orangecircle) models the intermediate robot motion, and the third state (\pinkcircle) encapsulates the time instants where the end-effector reaches the peg.}
	\label{fig:tphsmm}
\end{figure*}

\subsubsection{Model Identification}
\label{subsubsec:ModelId}
As discussed earlier, the controller performance may degrade if the robot dynamics model is inaccurate.
To address this problem, we resort to classical model identification methods to access more precise dynamic parameter estimates to improve the robot control accuracy.  
Specifically,  an inverse model of the robot is identified to map the desired joint accelerations to actuation torques in the computed-torque control law.
Given that there are no external forces acting on the robot in the identification experiments, the robot dynamics model can be written as $\bm{\tau}_u = \bm{\Phi}(\bm{\theta}, \dot{\bm{\theta}}, \ddot{\bm{\theta}}) \bm{\Psi}$,
such that it is linear in the so-called barycentric parameters~\cite{lewis2003robot}, $\bm{\Psi} = \left[ \bm{\Psi}_1^\trsp, \ldots, \bm{\Psi}_7^\trsp \right]^\trsp$.
These represent the barycentric parameters of every link and friction coefficients of the joints connecting all consecutive links, starting with the joint connecting the base with link $1$.
The barycentric parameters consist of the mass, the first- and second-order moments of inertia.
We parameterize asymmetric Coulomb friction and viscous friction by three parameters per joint.

Since $\bm{\Phi}$ is not unique in general, let us assume an equivalent relation $\bm{\tau}_u = \hat{\bm{\Phi}}(\bm{\theta}, \dot{\bm{\theta}}, \ddot{\bm{\theta}}) \hat{\bm{\Psi}}$ where the columns of $(\bm{\theta}, \dot{\bm{\theta}}, \ddot{\bm{\theta}})~\mapsto~\hat{\bm{\Phi}}(\bm{\theta}, \dot{\bm{\theta}}, \ddot{\bm{\theta}})$ are linearly independent.
The goal is to identify suitable parameters $\hat{\bm{\Psi}}$ based on experimental data.
Due to the above linearity, this can in principle be achieved by solving a convex regression problem (see~\ref{app:CIC} for details).
To this end, we designed an experiment to ``maximize'' the information contained in the data by using a parametric harmonic trajectory as in~\cite{Swevers1997}.
The base frequency is chosen sufficiently low such that a large part of the configuration space is covered without violating joint velocity, acceleration, and jerk limits.
The number of harmonics is limited to not excite undesired high-frequency phenomena.
Under the foregoing conditions, we formulate a non-convex optimization problem that can be solved to achieve local optimality through nonlinear programming solvers (see~\ref{app:CIC} for details).
With this aim, we use CasADi~\cite{Andersson2019} to formulate the optimization, perform automatic differentiation, and solve the problem using the interior-point solver IPOPT~\cite{Waechter2006}.

Once measurements from the optimal trajectories are collected, the convex regression problem to estimate the barycentric parameters is solved while enforcing physical feasibility in a semi-definite program~\cite{Sousa2014,Wensing17:InertialParamID,Sousa2019}.
Physical feasibility refers to whether bodies that correspond to the inertia parameters could exist in reality,
this includes the body masses to be positive, the inertia tensors to be positive definite, and furthermore that the so-called triangle inequality on the Eigenvalues of the inertia tensors hold~\cite{Traversaro2016}.
Without the additional constraints proposed in~\cite{Wensing17:InertialParamID,Sousa2019}, the parameters obtained from solving the convex regression are not guaranteed to be physically feasible because of noise and unparameterized phenomena captured by the data.
By ensuring physical feasibility to hold, the estimated parameters may not only used for controlling our real robot but also for forward simulation.

\subsection{Learning Motion Skills from Human Demonstrations}
\label{subsec:LfD}
As outlined above, we have multiple requirements on the representation of robot motion skills, namely: \emph{(i)} they should be easy to teach and train, \emph{(ii)} able to reproduce motions and forces with high precision, \emph{(iii)} adapt to various object pose configurations, and \emph{(iv)} be data efficient. 
We further assume that the manipulated objects are only characterized by their pose in the robot workspace, provided by the perception module (see Section~\ref{subsec:DON}).
Given the aforementioned conditions, the available learning models matching these criteria lie midway through the spectrum of LfD approaches (where one extreme corresponds to pure deep neural networks, while the opposite extreme represents plain replay methods)~\cite{Osa18:ImitationLearning}.

Specifically, we leverage a task-parametrized version of hidden semi-markov models (TP-HSMMs, more details in Section~\ref{subsubsec:hsmm}). 
This approach provides a probabilistic encoding of the spatial and temporal patterns of human demonstrations, it is easy to train, and usually requires less than ten demonstrations for most skills considered in our use cases.
As our manipulation skills display full-pose trajectories, composed of Cartesian and quaternion data, we take advantage of Riemannian-manifold theory to extend TP-HSMMs to handle data belonging to non-Euclidean spaces.
Moreover, we exploit the same TP-HSMM to learn force-based skills, where the model encodes trajectories of a virtual attractor system that encapsulates the demonstrated force patterns (see Section~\ref{subsubsec:force-lfd}), analogously to set-point adaptation methods in indirect force control~\cite{Villani2016}.

\subsubsection{TP-HSMM}
\label{subsubsec:hsmm}
Within our LfD setup, an operator provides several kinesthetic demonstrations of a motion or force-based skill. 
The demonstration data follow an object-centric structure, meaning that the data are recorded from multiple coordinate systems, often associated to the pose of the objects manipulated during the skill at hand.
The representation of these coordinate systems via translation and rotation components provides the task parameters for the skill model.
These task parameters are used to define an affine transformation of the recorded data $\mathsf{D}_{\mathsf{a}}$, resulting in a set of demonstrations that are locally projected with respect to each object pose.
The training procedure consists of running a modified version of the Expectation-Maximization (EM) algorithm to learn a task-parametrized hidden semi-Markov model.
 
More specifically, the TP-HSMM, with parameters $\bm{\Theta}_\mathsf{a}$, fits  Gaussian mixture models (GMMs) to the projected data in order to encode the spatial patterns of the demonstrations. 
This model also encapsulates the sequential and temporal patterns of the skill by learning a transition graph and a set of duration probabilities.
The former allows us to learn different skill instances (e.g., grasping an object with different relative orientations), while the latter permits to encode the skill timing, which may be relevant when approaching or dropping an object. 
Figure~\ref{fig:tphsmm} provides an illustration of the object-centric demonstration data, the local data encoding, and the TP-HSMM graph.
In summary, a TP-HSMM provides different abstraction levels of the demonstrations, namely, raw spatial trajectories encoding (via local GMMs), demonstrations timing (via duration probabilities), and skill structure (via the transition graph).
The first two levels are relevant for the reproduction of a single skill (as described in Section~\ref{subsubsec:task-execution}), while the latter is exploited for sequencing several skills (see Section~\ref{subsec:sequencing}).  
A technical explanation of this model is provided in~\ref{app:skill_learn}.
Further details on the use of TP-HSMM in industrial settings can be found in~\cite{RozoEtal:IROS20}.

\subsubsection{Force-based Skills Learning}\label{subsubsec:force-lfd}
\begin{figure}[t]
	\centering
	\includegraphics[width=\linewidth]{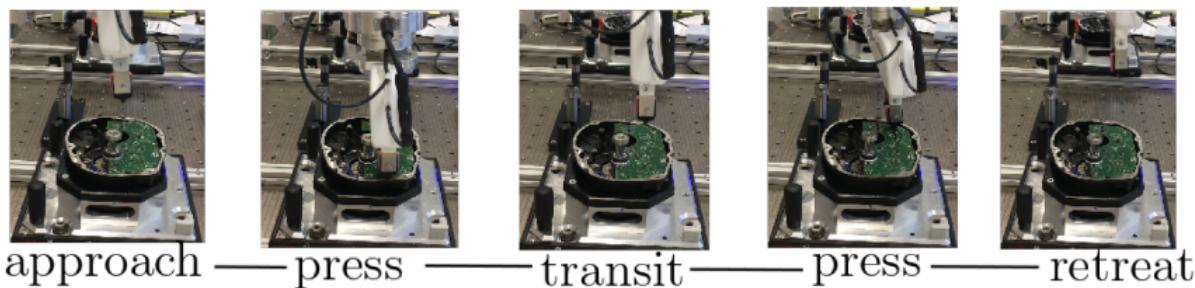}
	\caption{Stages of the force-based skill Press-PCB described in Section~\ref{subsec:IntroEbike}. Note that different forces need to be applied at different locations as shown in Fig.~\ref{fig:press-pcb-traj}.}
	\label{fig:press-pcb-execute}
\end{figure}

\begin{figure}[t]
\centering
\includegraphics[width=.95\linewidth]{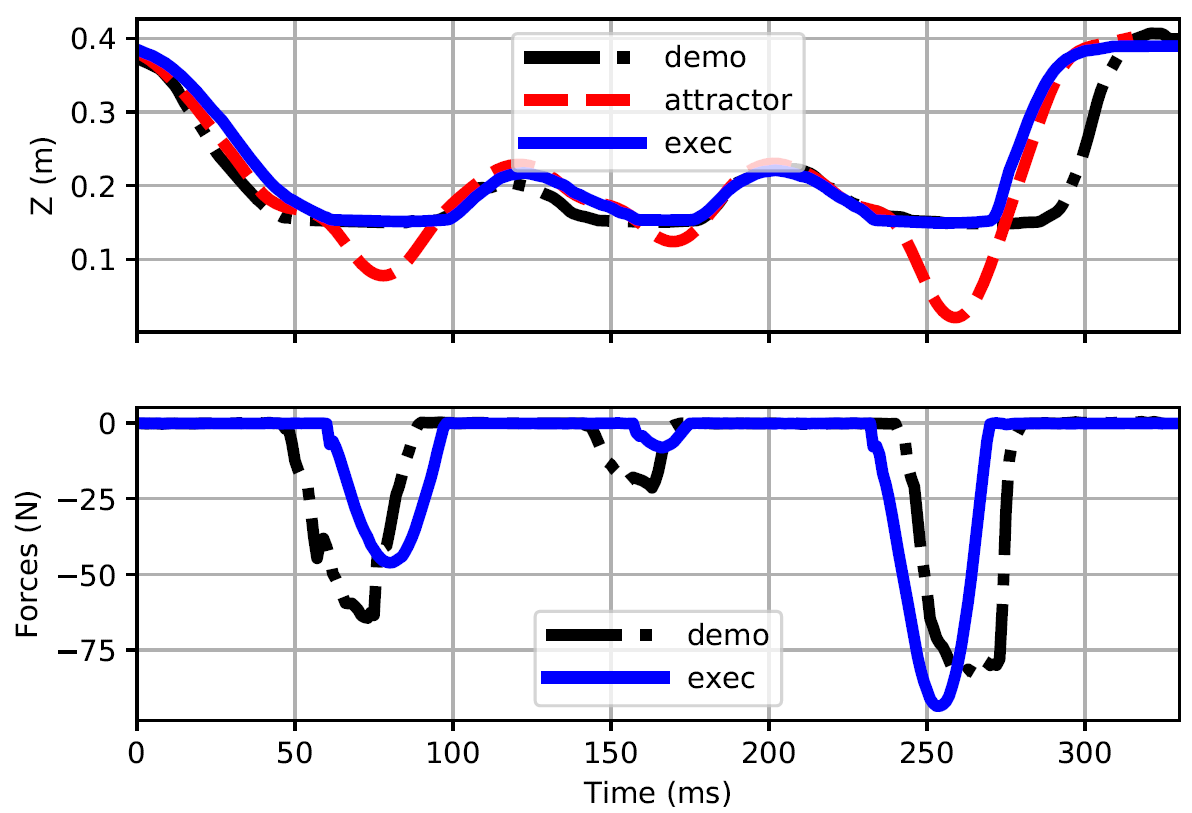}
\caption{Time-series plots of demonstrated position and force along the $z$-axis, and corresponding attractor trajectory for the Press-PCB skill shown in Fig.~\ref{fig:press-pcb-execute}. }
\label{fig:press-pcb-traj}
\end{figure}

For assembly tasks that involve force-sensitive interaction with the environment, stiff kinematic trajectory tracking as mentioned above is often inadequate.
As shown in Fig.~\ref{fig:press-pcb-execute} and Fig.~\ref{fig:press-pcb-traj}, the electrical PCB board should be pressed into various pins with different forces.
As a result, this pressing skill requires tracking not only for kinematic, but also force trajectories and associated stiffness levels.
To address these challenges we leverage a force-based LfD approach which explicitly accounts for the interaction forces between the robot end-effector and the environment to build the skill model.

Specifically, our force-based LfD framework builds on the concept of set-point adaptation for indirect force control. 
The main idea is to leverage a virtual mass-spring-damper (MSD) system subject to an external force, whose dynamics is used to define an attractor trajectory to drive the robot motion, as previously proposed in~\cite{Rozo16:pHRC_learning}.
The objective of using the virtual MSD model is twofold: \emph{(i)} it allows us to encapsulate the dynamics of the demonstrated interactions into a compact and simple model, and \emph{(ii)} it provides us a mechanism to vary the robot compliance by either adapting the MSD set point or its stiffness matrix (see the technical details in~\ref{app:force-lfd}). 

Thus, learning a force-based skill involves recording the position, velocity, acceleration and sensed forces of the robot end-effector from a set of kinesthetic demonstrations.
These data are then transformed to attractor trajectories using the virtual MSD model, whose stiffness matrix matches the default stiffness of the underlying robot impedance controller. 
Examples of a demonstrated pose trajectory and the corresponding attractor trajectory for the press-pcb skill are shown in Fig.~\ref{fig:press-pcb-traj}.
Note that the resulting attractor pose can differ greatly from the demonstrated pose when large velocities and sensed forces are present.  
The resulting set of attractor trajectories is then used to learn a TP-HSMM model as described in Section~\ref{subsubsec:hsmm}.
Moreover, we can associate a ``local'' stiffness matrix to each of the Gaussian components of the learned TP-HSMM by estimating the stiffness that best explains the local dynamics encoded by the component (see details in~\ref{app:force-lfd}).
These local stiffness matrices allow us to define a time-varying robot compliance during the reproduction of the force-based skill.
This model fulfilled learning and reproduction requirements of several force-based skills that are part of the e-Bike use case~\cite{le2021learning}.

\subsubsection{Skill Condition Models Learning} \label{subsubsec:skill-models}
As explained in Section~\ref{subsec:IntroEbike}, many of the sub-tasks of the e-Bike motor assembly process involve sequencing several types of skills. 
This means that every time the robot executes a skill, the state of the environment changes. 
For example, after executing an object reorientation skill, the pose of the manipulated object and the robot end-effector vary. 
In order to adapt the learned skill models to these changes in the environment, we need to capture the relative configuration among the workspace objects and the robot end-effector before and after executing a specific skill.
To achieve this, we propose to extend the skill model by learning condition and effect models.

Specifically, for each skill $\mathsf{a}\in \mathsf{A}$ and corresponding demonstrations set $\mathsf{D}_{\mathsf{a}}$, an extended model $\widetilde{\boldsymbol{\Theta}}_{\mathsf{a}} = (\boldsymbol{\Theta}_\mathsf{a},\, \boldsymbol{\Gamma}_{\mathsf{a}})$ is learned by the method proposed in~\citep{RozoEtal:IROS20,Schwenkel2019Optimizing}, see~\ref{app:condition-model} for details.
This extended model represents:
\emph{(i)} the \emph{skill} model~$\boldsymbol{\Theta}_{\mathsf{a}}$
as the TP-HSMM that encapsulates both temporal and spatial properties in the demonstrations set $\mathsf{D}_{\mathsf{a}}$, as described in~\cref{subsubsec:hsmm};
and \emph{(ii)} the \emph{condition} model $\boldsymbol{\Gamma}_{\mathsf{a}} = (\boldsymbol{\gamma}_{1, \mathsf{a}}, \, \boldsymbol{\gamma}_{1T,\mathsf{a}}, \boldsymbol{\gamma}_{T,\mathsf{a}})$, which is composed of:
\begin{enumerate}[(a)]
	\item The \emph{precondition} model~$\boldsymbol{\gamma}_{1, \mathsf{a}}$ encoding the configuration of the system state \emph{before} executing the skill;
	\item The \emph{effect} model~$\boldsymbol{\gamma}_{1T,\mathsf{a}}$	encapsulating how the system state \emph{changes} from the initial state to the final state after executing the skill;
	\item The \emph{final condition} model $\boldsymbol{\gamma}_{T, \mathsf{a}}$ encoding the system state configuration \emph{after} executing the skill.
\end{enumerate}
The aforementioned condition models have a GMM structure and are described in detail in~\ref{app:condition-model}.

\begin{figure}[t]
	\centering
	\includegraphics[width=.97\linewidth]{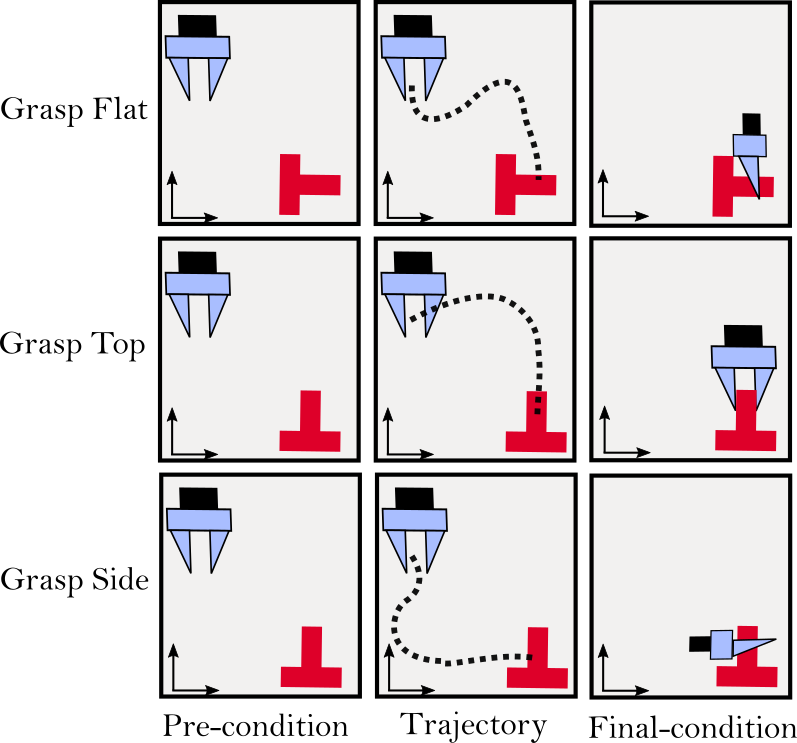}
	\caption{Illustration of the condition and effect models for the \texttt{grasp} skill as described in~\cref{sec:experiments}. Each row represents a different instance of the \texttt{grasp} skill.}
	\label{fig:condition-effect}
	\vspace{-0.4cm}
\end{figure}

Once the associated task parameters are given, the above models can be used to compute the full trajectory, check the initial satisfiability, and predict the resulting state for the skill $\mathsf{a}$.
An illustration of the condition and effect models for a grasping skill is shown in Fig.~\ref{fig:condition-effect}.
It displays the relative pose configurations between the end-effector and the object at the beginning and the end of the demonstrations, respectively.
Notice that these models encapsulate all different instances, i.e. branches in the TP-HSMM model, of the same skill.
These condition models are also leveraged to sequence several skills as explained next.
We omit the detailed notations and their derivations, and refer the readers to our previous works~\citep{RozoEtal:IROS20,Schwenkel2019Optimizing} and the technical~\ref{app:condition-model}.

%

\subsection{Skills Sequencing and Task Planning}
\label{subsec:sequencing}

\subsubsection{Modular Skills Sequencing for a Task} \label{subsubsec:modular-composition}
Consider a given task as the desired skill sequence $\boldsymbol{\mathsf{a}}^\star=(\mathsf{a}_1, \cdots, \mathsf{a}_N)$, which can be either manually specified or derived from a high-level planner.
Since each skill in $\boldsymbol{\mathsf{a}}^\star$ may have many branches (i.e. the skill instances), the entire task can be executed in many different ways, which is combinatorial to the number of branches of each skill $\mathsf{a}_n$ within the task.
A common approach is to manually specify these branching conditions to guide the desired task execution.
Our objective is to avoid this manual process via a modular composition of skills within the task.

More precisely, the task model associated with $\boldsymbol{\mathsf{a}}^\star$, denoted by $\widehat{\mathbf{\Theta}}_{\boldsymbol{\mathsf{a}}^\star}$, is computed recursively as follows.
Starting from the first two consecutive skills $(\mathsf{a}_1, \mathsf{a}_2)$, the corresponding skill models $\widetilde{\boldsymbol{\Theta}}_{\mathsf{a}_1}$ and $\widetilde{\boldsymbol{\Theta}}_{\mathsf{a}_2}$ are composed into one $\widehat{\boldsymbol{\Theta}}_{{\mathsf{a}}_1\boldsymbol{\mathsf{a}}_2}$ by applying the three following steps:
\begin{enumerate}[(1)]
	\item Concatenate copies of $\mathbf{\Theta}_{\mathsf{a}_2}$ to \emph{every} final state of $\mathbf{\Theta}_{\mathsf{a}_1}$; for example, if the skill model $\mathbf{\Theta}_{\mathsf{a}_1}$ contains two instances, then it has two final states, and thus two copies of $\mathbf{\Theta}_{\mathsf{a}_2}$ are concatenated to $\mathbf{\Theta}_{\mathsf{a}_1}$.
	\item Modify the components of each copy of $\mathbf{\Theta}_{\mathsf{a}_2}$ according to the effect model $\boldsymbol{\gamma}_{1T,\mathsf{a}_1}$; this means that the effect model indicates the possible variations of task parameters for each final state in $\mathbf{\Theta}_{\mathsf{a}_1}$, and then each concatenated copy of $\mathbf{\Theta}_{\mathsf{a}_2}$ is updated accordingly.
	\item Computing the transition probability from every final state of $\mathbf{\Theta}_{\mathsf{a}_1}$ to every initial state of the copies of $\mathbf{\Theta}_{\mathsf{a}_2}$, according to the final condition model $\boldsymbol{\gamma}_{T, \mathsf{a}_1}$ and the precondition model $\boldsymbol{\gamma}_{1, \mathsf{a}_2}$.
\end{enumerate}
Afterwards, $\widehat{\mathbf{\Theta}}_{\boldsymbol{\mathsf{a}}_1\boldsymbol{\mathsf{a}}_2}$ is composed with the skill model $\widetilde{\boldsymbol{\Theta}}_{\mathsf{a}_3}$ in the same way.
This process is repeated until $\mathsf{a}_N$ is reached, yielding $(\widehat{\mathbf{\Theta}}_{\boldsymbol{\mathsf{a}}^\star}, \widehat{\boldsymbol{\Gamma}}_{\boldsymbol{\mathsf{a}}^\star})$,
that is the composed model encapsulating the whole task in a TP-HSMM.
The technical details about the aforementioned steps are given in our previous work~\cite{RozoEtal:IROS20}.

\subsubsection{Task Planning} \label{subsubsec:task-execution}
The derived model $\widehat{\mathbf{\Theta}}_{\boldsymbol{\mathsf{a}}^\star}$ is used to reproduce the skill sequence~$\boldsymbol{\mathsf{a}}^\star$ as follows.
First, the initial system state $\vec{x}_{\mathsf{s},1}$, describing the initial robot and objects poses, is obtained from the robot perception system. 
The goal system state $\boldsymbol{x}_{\mathsf{s},\mathsf{G}}$ is defined, e.g., by a high-level task planner.
Then, the most-likely state sequence $\widehat{\boldsymbol{s}}^\star$ within model $\widehat{\mathbf{\Theta}}_{\boldsymbol{\mathsf{a}}^\star}$, given $\vec{x}_{\mathsf{s},1}$ and $\boldsymbol{x}_{\mathsf{s},\mathsf{G}}$, is generated by a Viterbi-like algorithm over the task model $\widehat{\mathbf{\Theta}}_{\boldsymbol{\mathsf{a}}^\star}$, as in~\cite{RozoEtal:IROS20}.
Note that $\widehat{\boldsymbol{s}}^\star$ contains skill-specific subsequence of states, denoted by $\widehat{\boldsymbol{s}}_n^\star$, to be followed for \emph{each} skill $\mathsf{a}_n\in \boldsymbol{\mathsf{a}}^\star$.

During the online execution of each skill $\mathsf{a}_n\in \boldsymbol{\mathsf{a}}^\star$, the current system state $\vec{x}_{\mathsf{s},t}$ is observed (e.g., the pose of objects of interest) and used to compute the task parameters and update the global GMMs associated with the states in~$\widehat{\boldsymbol{s}}_n^\star$.
Afterwards, the trajectory tracking control described in the next section is used to track~$\widehat{\boldsymbol{s}}_n^\star$.
This process repeats for all skills in $\boldsymbol{\mathsf{a}}^\star$.
Note that simply tracking $\widehat{\boldsymbol{s}}^\star$ without observing the actual intermediate system state $\vec{x}_{\mathsf{s},t}$ would often fail due to perception and motion noise.
More implementation details for this workflow are given in~\cref{subsec:software}.

\begin{figure}[t]
	\centering
	\includegraphics[width=.45\linewidth]{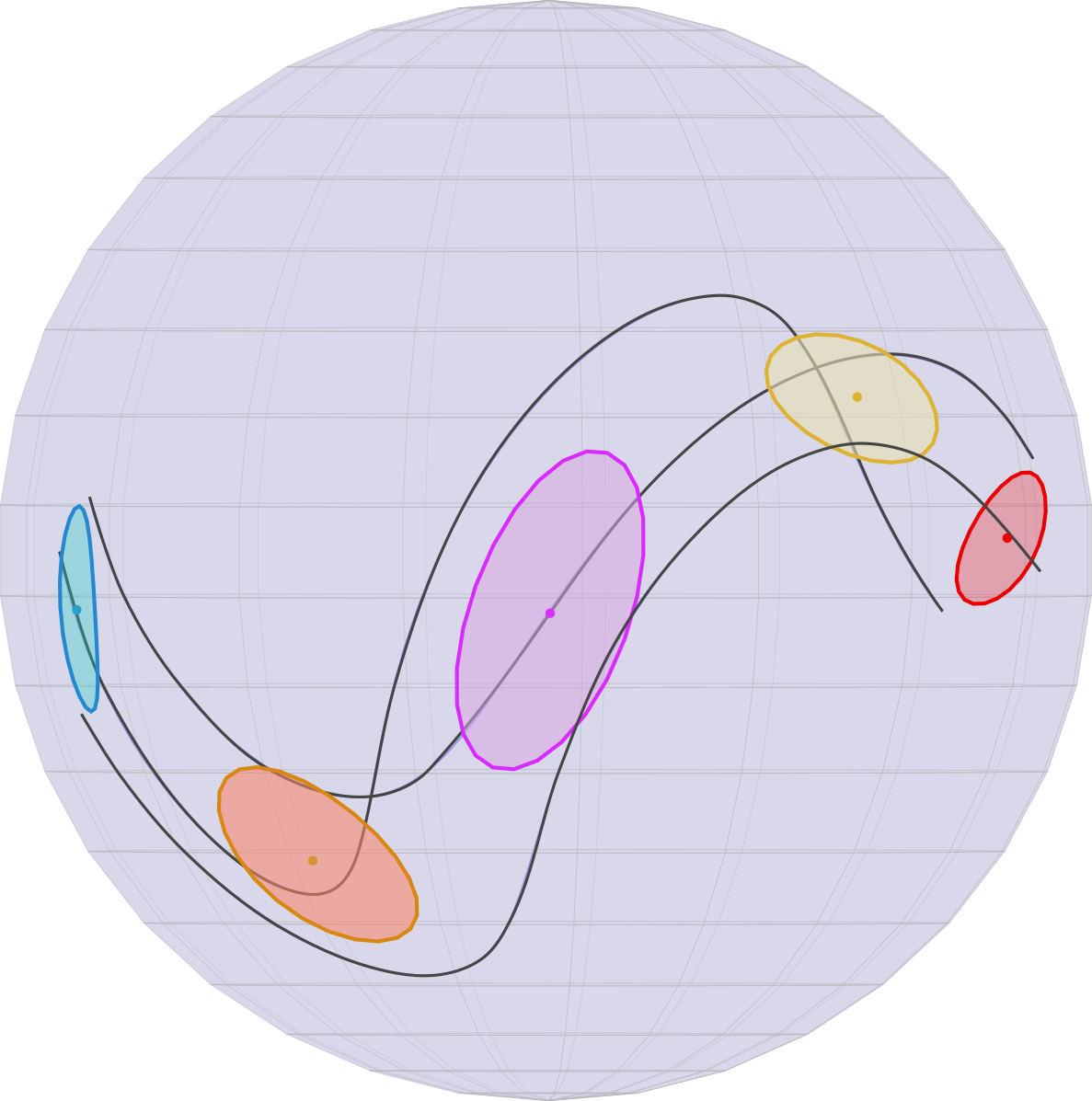}
	\hspace{.15cm}
	\includegraphics[width=.45\linewidth]{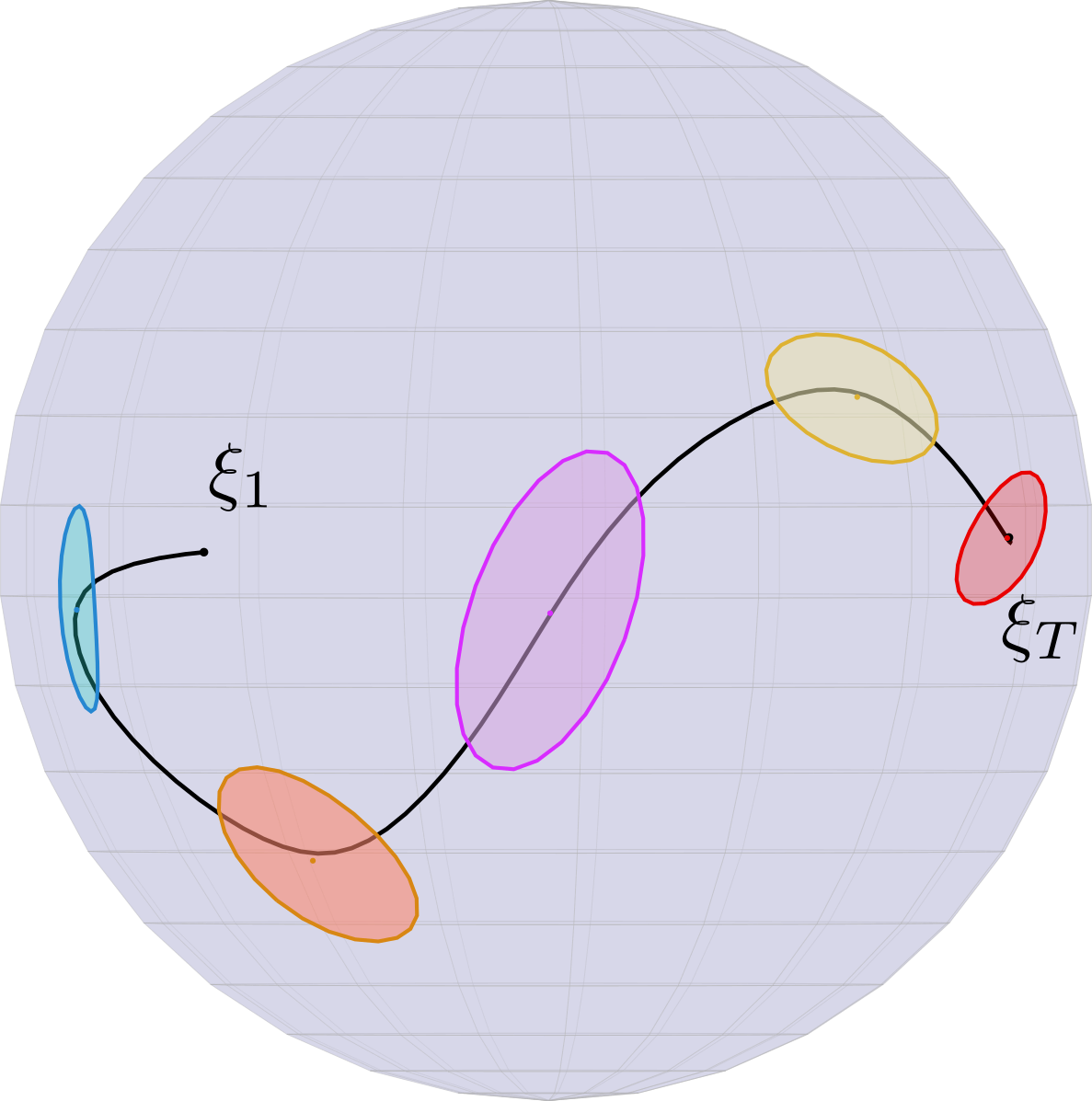}
	\caption{\textbf{Left:} Illustration of a learned TP-HSMM with $5$ components on the Riemannian manifold $\mathcal{S}^2$ from $3$ demonstrations (\graydemo).
		\textbf{Right:} The reproduced trajectory (\blackrepro) from an arbitrary initial state $\vec \xi_1$ using Riemannian LQT. }
	\label{fig:riemannian_gmm}
	\vspace{-0.4cm}
\end{figure}

\subsection{Motion Trajectory Generation}
\label{subsec:reproduction}
To generate a smooth reference trajectory following the transformed GMM components, we use linear quadratic optimal tracking (LQT) as a trajectory generation tool.
To do so we exploit a linear double integrator dynamics in the control formulation such that the reference end-effector follows a virtual spring-damper system.
This allows us to compute an optimal and smooth reference trajectory in closed form without knowledge of the actual robot dynamics.
The cost function components are composed of the time stamped means and covariances of the components provided by the HSMM model.

Figure~\ref{fig:riemannian_gmm} illustrates the learning and reproduction pipeline on Riemannian manifolds.
For simplicity, we consider the $\mathcal{S}^2$ manifold and $3$ demonstrations.
We learn a TP-HSMM with $5$ states and a single task parameter (we illustrate the resulting GMM on the left side of Fig.~\ref{fig:riemannian_gmm}).
Then, we arbitrarily choose an initial state $\vec \xi_1 \in \mathcal{S}^2$ and solve the control problem in~\eqref{eq:lqt}.
We retrieve the optimal trajectory via forward integration and illustrate it on the right side of Fig.~\ref{fig:riemannian_gmm}.
Note that our system architecture allows a completely independent development of the learning and the control framework due to the abstraction provided by the interface layer, see~\cref{subsec:software}.
More details about motion trajectory generation are given in~\ref{app:lqt}.


\subsection{Skill optimization}\label{subsec:skills_refinement}
Although learning robot skills from human demonstrations using TP-HSMM suffices to perform relatively-elaborated tasks, the skill reproduction might still be prone to errors due to sensor noise or inaccurate tracking control. 
For example, reaching and placing skills may not demand high accuracy, as opposed to insertion tasks which require highly-precise force-sensitive interactions due to low geometric tolerances.
This often demands to refine the skill to overcome sensory and control inaccuracies for successful executions.

Another way we exploit skill refinement in our robotic manipulation tasks concerns minimizing the interaction forces of the force-sensitive skills.
When learning from demonstrations, the human operator may not pay particular attention to the forces applied to the handled pieces.
Therefore, reducing the interaction forces is of particular interest to mitigate possible damage of the assembly parts due to unnecessarily high forces.
Hence, our objective is to make the robot more compliant while still being able to perform the insertion skills successfully. 
Consequently, endowing robots with fast and data-efficient refinement strategies is imperative to adapt previously-learned skills under uncertain environmental conditions or to optimize their performance. 
In this section, we present two approaches that we leverage to address this problem: self-supervised learning and Bayesian optimization.


\subsubsection{Self-supervised strategies}
\label{subsubsec:self_refinement}
One possibility to achieve the required skill precision considers training a neural network to predict the necessary refinements for the learned skill. 
This strategy is particularly useful in situations where fine-tuning of the learned skills is only necessary for a very limited number of parameters. 
In this work we use this approach to predict small corrections to the final end-effector pose of the reaching skill, prior to insertion. 
To avoid tedious manual data labeling, we adopt a self-supervised training strategy: We first collect self-supervised training data during an online phase, which  we then use in an offline phase to train the refinement network.
 
For data collection, we employ a simple random-search strategy that applies pose offsets w.r.t. the final end-effector pose of the reaching skill and subsequently tries to execute an insertion skill at the resulting explored poses. 
Since the final end-effector pose of the skill is already close to the target, this random strategy usually finds the correct pose after few trials.
Specifically, we record training data composed of the force/torque sensor readings $\bm{f}_t$ and the robot end-effector pose $\bm{x}_{\mathsf{e},t}$ for each trial pose, and we label each data point with the ground-truth pose.
Note that goal achievement can be easily determined by the $z$-position of the robot end-effector and the corresponding $x$ and $y$ coordinates can thus be used as ground-truth labels. 

To deal with sequential data, we use an LSTM as refinement model, which is trained on the collected insertion sequences. 
At each step, the LSTM model $g_{\bm{\theta}}(\bm{x}_{\mathsf{e},t}^{\mathrm{rel}}, \bm{f}_t; \bm{x}_{\mathsf{e},1:t-1}^{\mathrm{rel}}, \bm{f}_{1:t-1})$ receives as inputs the force/torque measurements $\bm{f}_t$ as well the robot end-effector pose $\bm{x}_{\mathsf{e},t}^{\mathrm{rel}}$ relative to the final pose of the reaching skill.
The LSTM model $g(\bm{\theta})$ predicts the offset $\Delta \bm{x}_t$ relative to the ground-truth goal pose, conditioned on all previous measurements $\bm{x}_{\mathsf{e},1:t-1}^{\mathrm{rel}}, \bm{f}_{1:t-1}$.
The LSTM parameters $\bm{\theta}$ are optimized using the mean squared error between predicted and ground-truth offsets.


\subsubsection{Geometry-aware Bayesian Optimization}
\label{subsubsec:optimization}
An alternative approach to refine robot skills is black-box optimization.  
This is advantageous due to its data efficiency, making it favorable over deep reinforcement learning when the refinement process may require real-world interactions. 
In this work, we leverage Bayesian optimization (BO)~\cite{Shahriari16:BOreview} to refine the parameters of the skill model.   
For example, we employ BO to adapt a skill reference pose trajectory by refining the associated TP-HSMM parameters $\bm{\Theta}_\mathsf{a}$, or to modify the applied forces in contact-rich tasks by optimizing the stiffness of the attractor model of a force-based skill.   

BO is a sequential search algorithm for finding a global maximizer $\bm{\psi}^* = \argmax_{\bm{\psi} \in \mathcal{X}} f(\bm{\psi})$ of an unknown objective function $f$, where $\mathcal{X}$ is some domain of interest. 
The black-box function $f$ has no closed form, but can be observed point-wise at any arbitrary query point $\bm{\psi} \in \mathcal{X}$. 
These observations are noise-corrupted outputs $y \in \mathbb{R}$, for which ${\mathbb{E}[y | f(\bm{\psi})] = f(\bm{\psi})}$ is assumed to be the true value.
Commonly, BO specifies a prior belief over the possible objective functions via a Gaussian Process (GP)~\citep{Rasmussen06}. 
This model is refined at each BO iteration according to the observed data $\mathcal{D}_n=\{(\bm{\psi}_i, y_i)\}_{i=1}^n$ via Bayesian posterior updates.
An acquisition function $\gamma_n \colon \mathcal{X} \to \mathbb{R}$ guides the search for the optimum by resolving the explore-exploit tradeoff. 
This function assesses the utility of candidate query points for the next evaluation of $f$ using the GP posterior mean and variance. 
After $N$ queries, BO makes a final recommendation $\bm{\psi}_N$, representing its best estimate of the optimizer $\bm{\psi}^*$.


\subsection{Software Structure}
\label{subsec:software}

\begin{figure}[t]
	\includegraphics[width=0.98\linewidth]{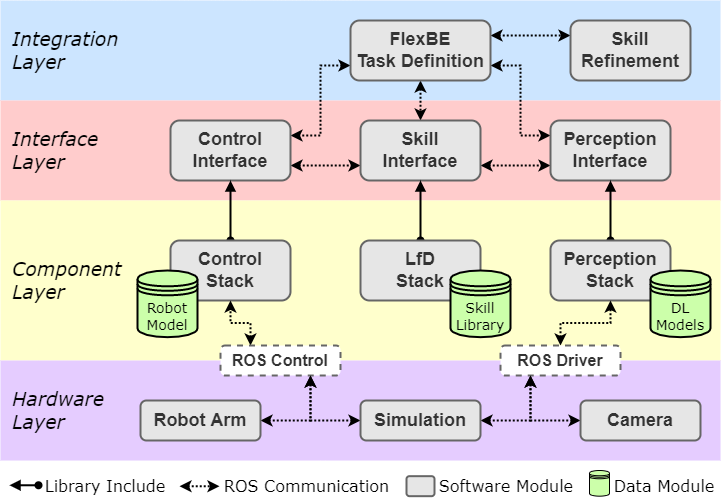}
	\centering
	\caption{System architecture of our robotic manipulation framework.}
	\label{fig:system-diagram}
	\vspace{-0.4cm}
\end{figure}

For the aforementioned refinement cases, the domain of interest $\mathcal{X}$ possesses a non-Euclidean geometric structure.
Therefore, we leverage our recent works on geometry-aware BO (GaBO)~\cite{jaquier2019bayesian,jaquier2021geometryaware} to work on domains represented by Riemannian manifolds.
This allows us to leverage the geometry of $\mathcal{X}$ as inductive bias in BO, which favors faster convergence and low-variance solutions.
For example, when optimizing the TP-HSMM parameters $\bm{\Theta}_\mathsf{a}$, non-Euclidean geometries arise in the mean vectors of the Gaussian distributions that encode demonstration data in $\mathbb{R}^3 \times \mathcal{S}^3$.
Also, the attractor model's stiffness of a force-based skill belongs to the manifold of symmetric positive-definite matrices $\mathcal{S}_{++}$ and it can therefore be more efficiently optimized using GaBO.   

Specifically, we employed GaBO to refine the peg insertion skill so that it produces lower interaction forces. 
To do so, we optimize the mean of the last Gaussian component of the TP-HSMM model (which dominates the approaching motion before insertion) and the corresponding attractor's stiffness matrix (which influences the applied forces). 
Thus, the optimization parameters are $\bm{\psi} = \{\hat{\bm{\mu}}_K, \bm{K}^{\rho}_K\}$, and consequently the problem domain $\mathcal{X} \equiv \mathbb{R}^3 \times \mathcal{S}^3 \times \mathcal{S}^3_{++}$. 
Our objective $f$ is a function of the mean norm of the sensed forces and the mean squared error between the end-effector and ground-truth goal poses. 
Note that a clear drawback of BO is that its exploration-exploitation tradeoff inherently requires the robot to fail.
To reduce the failed trials, we extended GaBO with a GP classifier to bias the acquisition function so that new query points are more likely to belong to high-reward -- success -- regions, similarly to~\cite{Englert16:BOandRL,Driess17:GPclassBO}.
After the optimization process, the best estimate $\bm{\psi}^*$ replaces the former model parameters set, thus resulting in new optimized TP-HSMM parameters $\bm{\Theta}_{\mathsf{a}}^*$.
We also recently leveraged signal temporal logic (STL) specifications to define objective functions capturing desired spatial and temporal requirements~\cite{Dhonthi22:STL_BO}.
This allowed us to represent high-level task objectives as logical semantics, this making their design more user-friendly.
For the insertion task, we could define a STL specification that encouraged the robot to reduce the approaching motion timing and the contact forces during insertion.
Details about the insertion skill refinement results are given in Section~\ref{subsubsec:assembly-results}.

\begin{figure}[t]
	\includegraphics[width=0.9\linewidth]{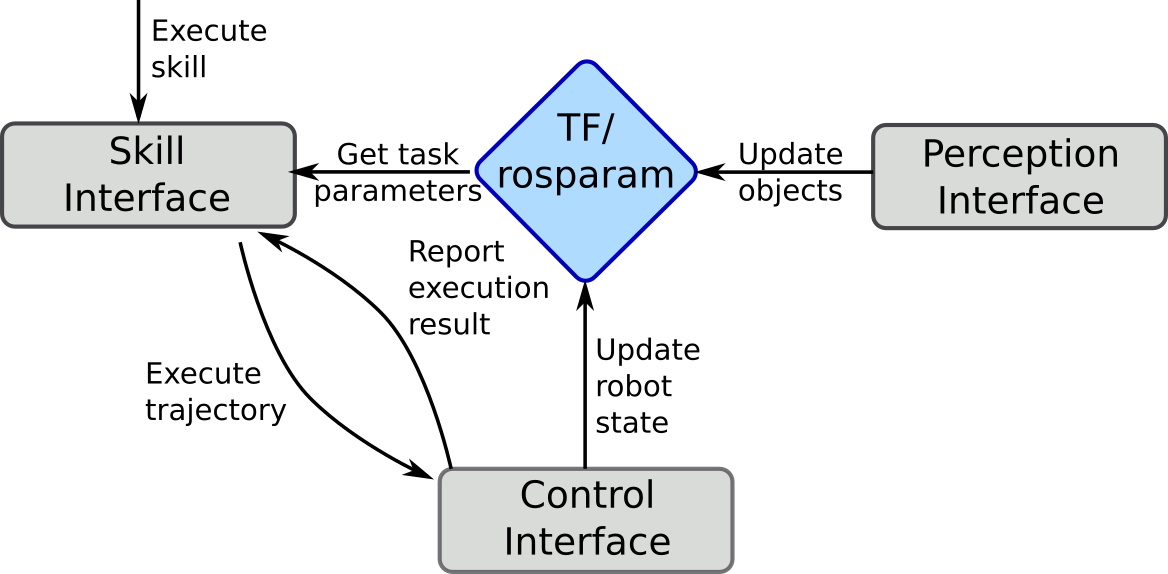}
	\centering
	\caption{Example of the information flow during skill execution.}
	\label{fig:system-example}
	\vspace{-0.4cm}
\end{figure}

The system architecture of our flexible robotic manipulation framework is illustrated in \cref{fig:system-diagram}.
In general, our system is divided into four layers as described below.

The highest layer is the \emph{Integration Layer}, which coordinates the different software modules and is used to define the task for a specific use case.
Here, we use FlexBE~\cite{flexbe} to implement a specific task like the E-Bike motor assembly and coordinate among function modules.
In addition, task refinement is considered at this layer to improve task execution over time and identify improved task parameters to be passed to the individual modules.

The integration layer is connected to the \emph{Interface Layer} using common ROS interfaces, e.g. action servers or services.
This layer provides ROS interfaces to our algorithmic components, and each interface is implemented by a set of ROS nodes.
This allows us to develop algorithms regardless of the used middleware or programming language.
Different interfaces may communicate with each other, e.g., (I) when executing a skill, the skill interface may use the control interface to move the robot; (II) the skill interface may query the perception interface for the current system state.

The \emph{Component Layer} contains our core algorithmic contributions described in~\cref{sec:system}.
Each component is connected to a ROS node in the interface layer by importing the components as libraries.
Maintaining this separation allows us to develop algorithms in different programming languages best suited to the concrete goal of the component while still being able to use them jointly in our system via the interface layer.
Most components are connected to underlying data bases, e.g., deep-learning networks or skill models.

The lowest layer in our system is the \emph{Hardware Layer}.
This layer provides interfaces to the actual hardware.
Our main connection between the hardware and the component layer is ROS Control~\cite{ros_control}, which allows for easy switching between hardware and simulation.
Due to the layered architecture in our system, all other components can transparently run both on hardware and in simulation.

Information about the current world state is mainly exchanged on the interface layer through \emph{tf}, the transform library~\cite{tf}, and the ROS parameter server, see~\cref{fig:system-example}.
All components of the interface layer can access this information and use it to react to changes.
This also allows asynchronous reads and writes on the world state, e.g., always executing a skill with the most recent task parameters obtained by the perception stack.

\begin{figure}[t]
	\centering
	\includegraphics[width=\linewidth,trim={10 10 10 10},clip]{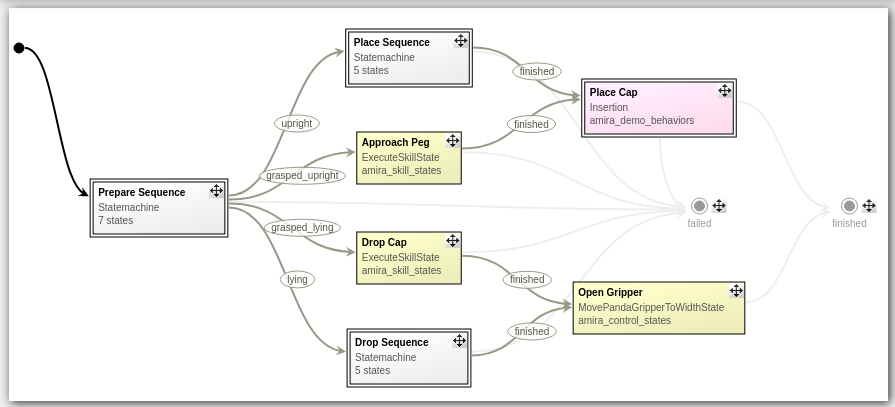}
	\includegraphics[width=\linewidth,trim={10 10 10 10},clip]{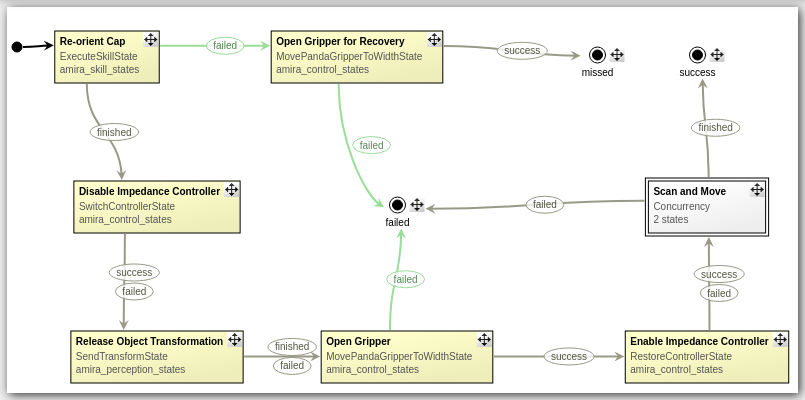}		
	\caption{Two examples for coordinating the different motions and primitives in FlexBE. \textbf{Top}: the main behavior to select one of the four described task variations. \textbf{Bottom}: re-orientation procedure that includes the re-orientation skill (initial state) and several primitives.}
	\label{fig:flexbe_examples}
\end{figure}

\subsubsection{Runtime View}\label{subsubsec:runtime-view}
A typical execution flow includes the following steps:
(I) a skill execution is requested by the integration layer, in our case FlexBE.
This request includes all necessary information for task execution as described in \cref{subsubsec:task-execution}, e.g., the desired goal state;
(II) the skill interface retrieves the relevant information from tf/rosparam.
These are mainly task parameters and the current world state, see \cref{subsubsec:task-execution};
(III) after computing a suitable solution in \cref{subsec:reproduction}, the resulting trajectory is sent to the control interface for execution;
(IV) the state of the execution is monitored by the skill interfacefor adapting to a changing world state;
(V) after finishing the execution, the control interface reports the outcome to the skill interface, which can then react, e.g., by recomputing (and executing) the skill on failure or reporting success to the integration layer.

\subsubsection{Failure Recovery}\label{subsubsec:failure-recovery}
One of the key issues in fully autonomous robotic systems is failure detection and recovery. 
Our current framework can monitor failures that may arise due to sensing or perception errors, inadequate grasping, unpredicted environment changes, among others.
On the one hand, a learned skill model may be used to supervise the execution process by comparing the sensory-motor patterns encoded by the model with those sensed during task reproduction.
Thus, possible failures may lead to sensory data that significantly differs from the patterns encoded in these learned models.
On the other hand, as shown in Fig.~\ref{fig:flexbe_examples}, the flexibility of FlexBE allows us to define recovery behaviors that re-scan the robot workspace to update and verify the current system state.
Both failure detection and recovery behaviors go hand in hand with an online adaptation strategy in order to sequence remaining skills on-the-fly.
Once the system state is updated, the sequencing algorithm can be used to re-compute the optimal execution branch for each skill.
In this way, human intervention is only needed when the current system state is not recognized.

	\section{Experiments}
	\label{sec:experiments}
\begin{figure*}[t]
\centering
\includegraphics[width=0.45\linewidth]{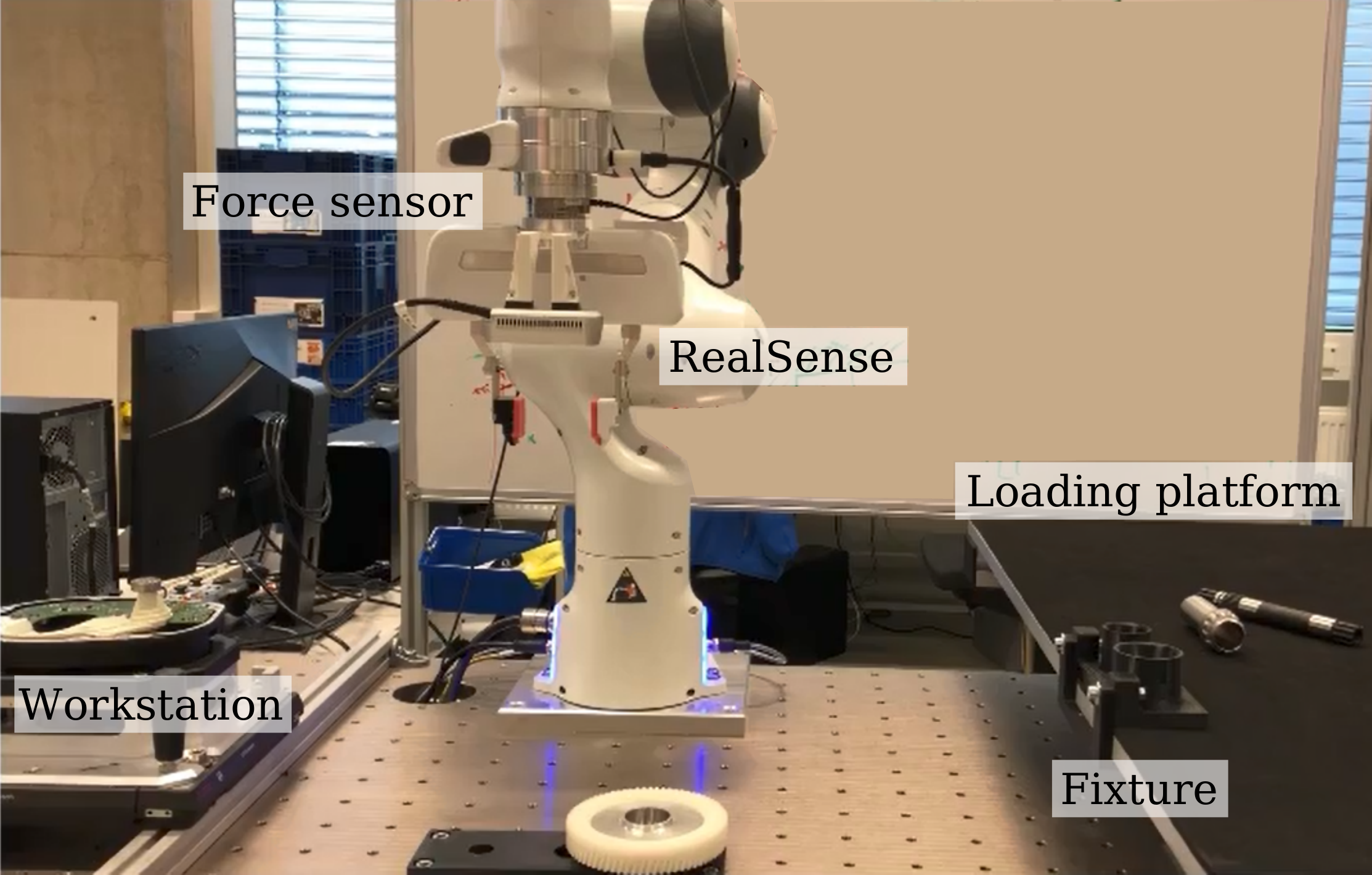}
\includegraphics[width=0.22\linewidth]{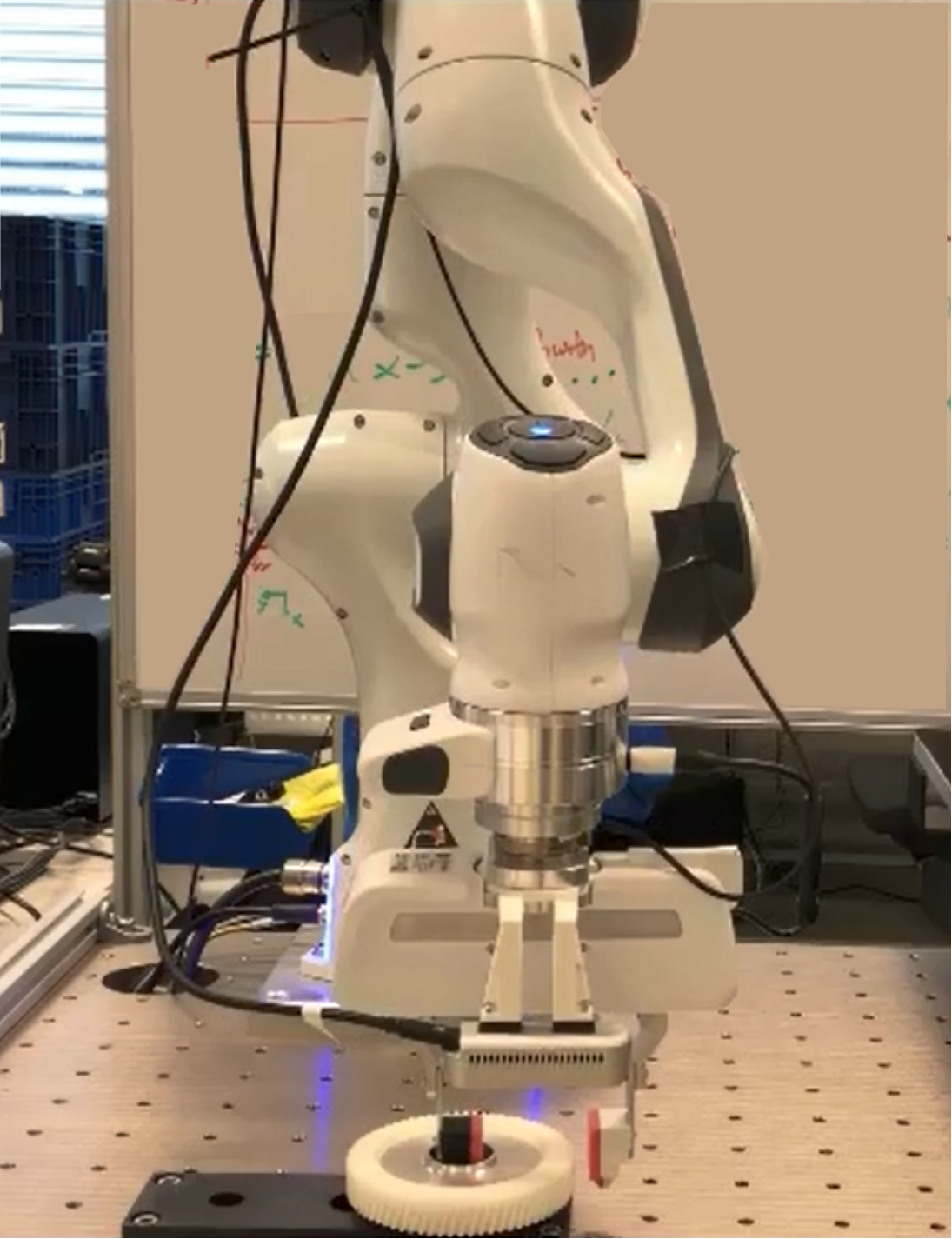}
\includegraphics[width=0.22\linewidth]{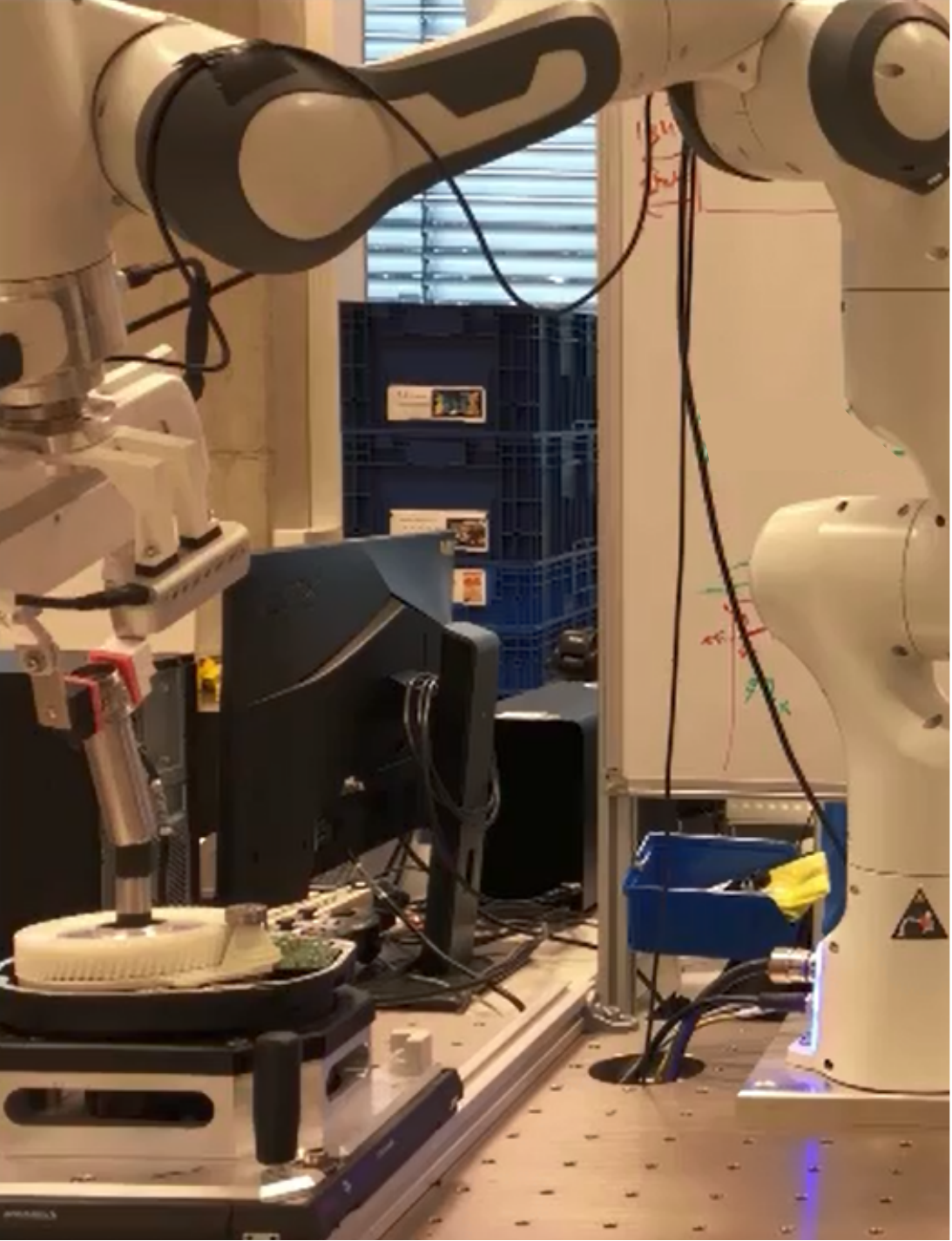}
\caption{
The experiment setup (\textbf{left}) with marked entities, and snapshots of the \emph{[Mount-Gear]} task showing the \texttt{grasp\_gear} (\textbf{middle}) and  \texttt{mount\_gear} (\textbf{right}) skills.
}
\label{fig:assembly-snapshots}
\end{figure*}

In this section, we describe the experimental setup on a 7-DoF robotic manipulator.
As initially motivated, we consider the e-bike motor assembly task as an application that can be solved under our framework.
Different steps of this task are shown in Fig.~\ref{fig:assembly-snapshots}.
In addition, we also consider a bin sorting task in~\cref{subsec:bin-sorting} to demonstrate the generality of our framework for addressing other manipulation tasks.
Videos of both experiments are attached as supplementary material.

\subsection{e-Bike Motor Assembly Task}\label{subsec:assembly}

As introduced in the task description provided in~\cref{subsec:IntroEbike}, the assembly task consists of four sub-tasks that are to be executed in a specific sequence.
We first give technical details on how this task is realized in hardware and prepared by providing the required skill demonstrations and refinements.

\begin{figure}[t]
	\centering
	\includegraphics[width=0.32\linewidth]{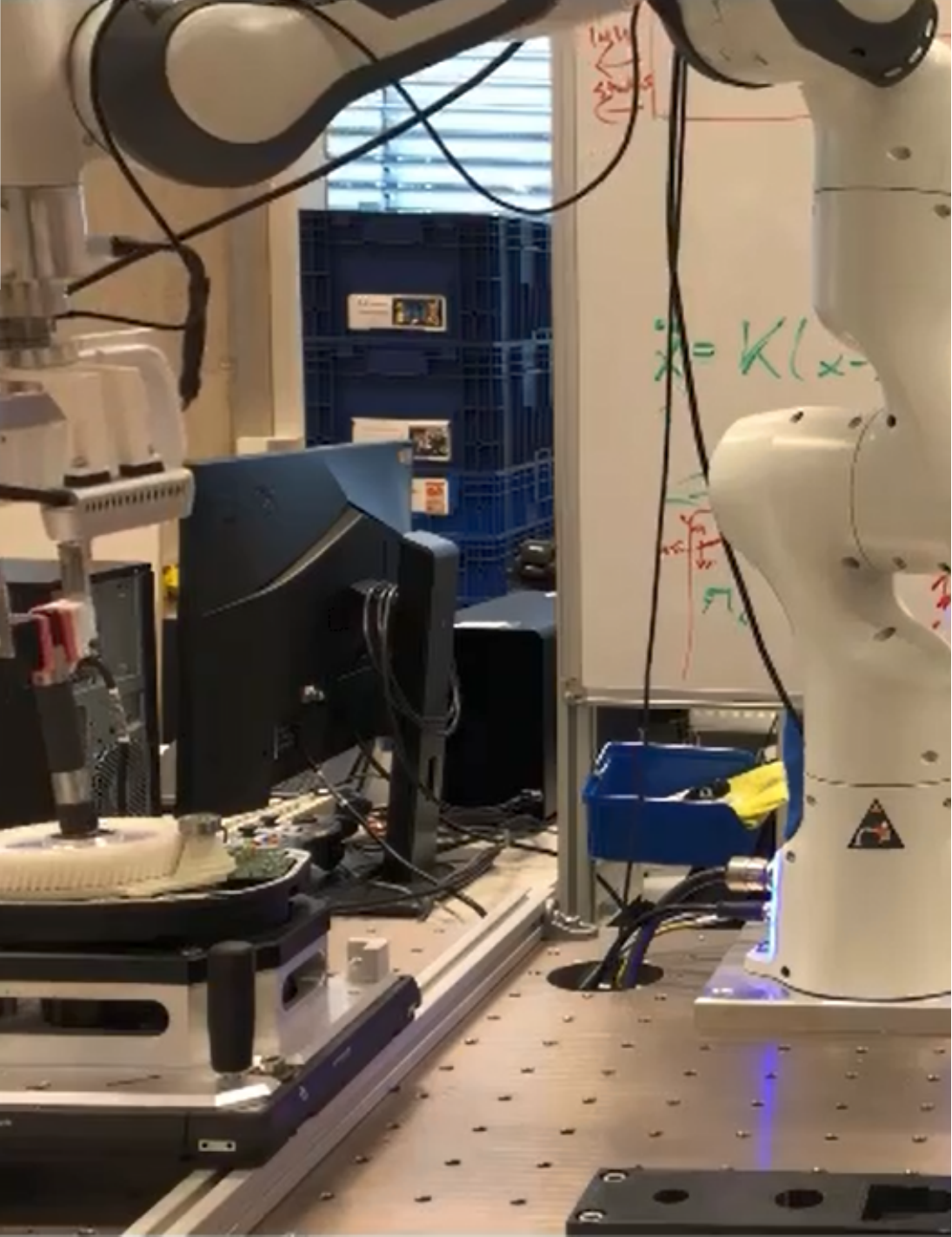}
	\includegraphics[width=0.314\linewidth]{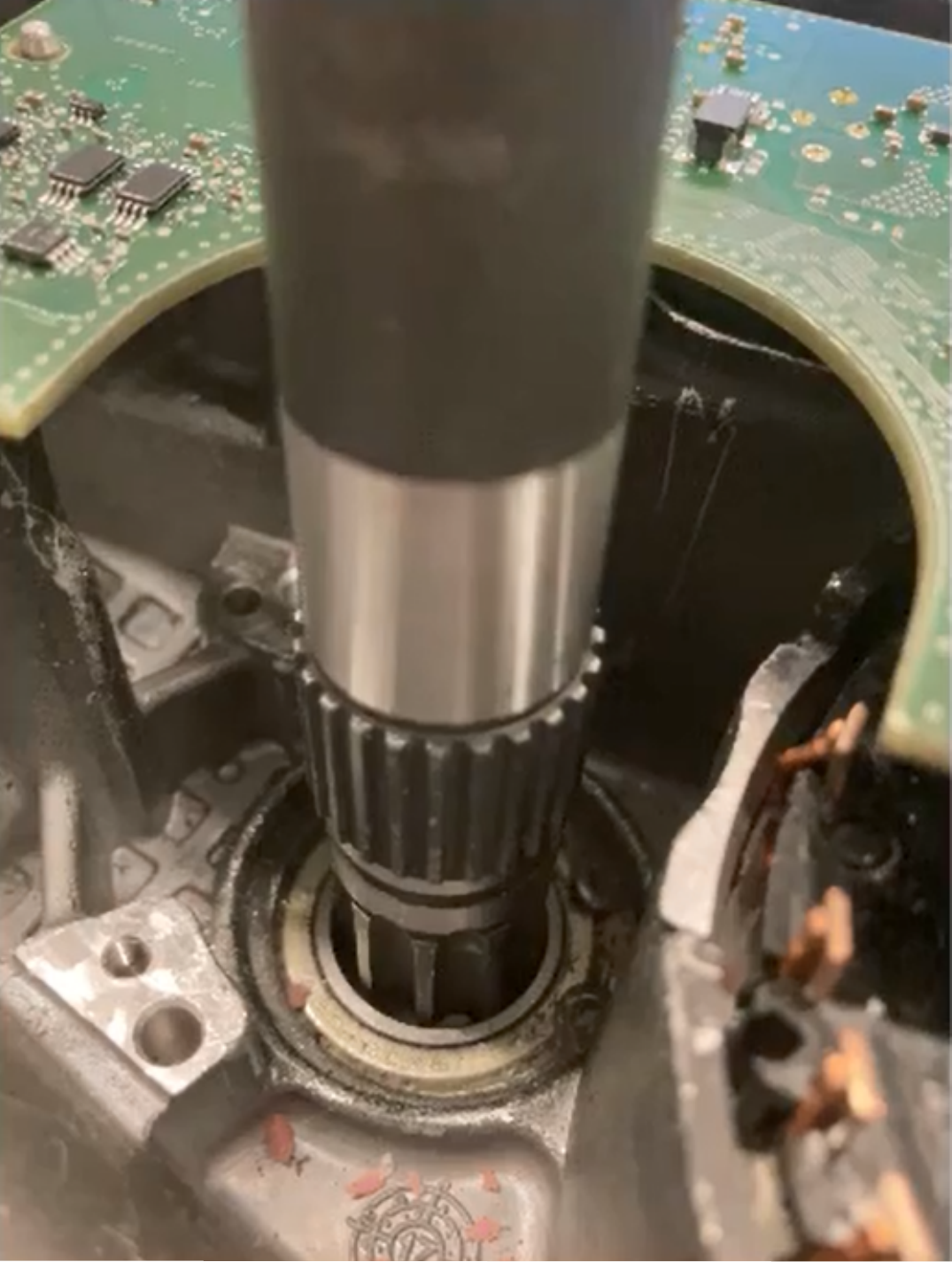}
	\includegraphics[width=0.315\linewidth]{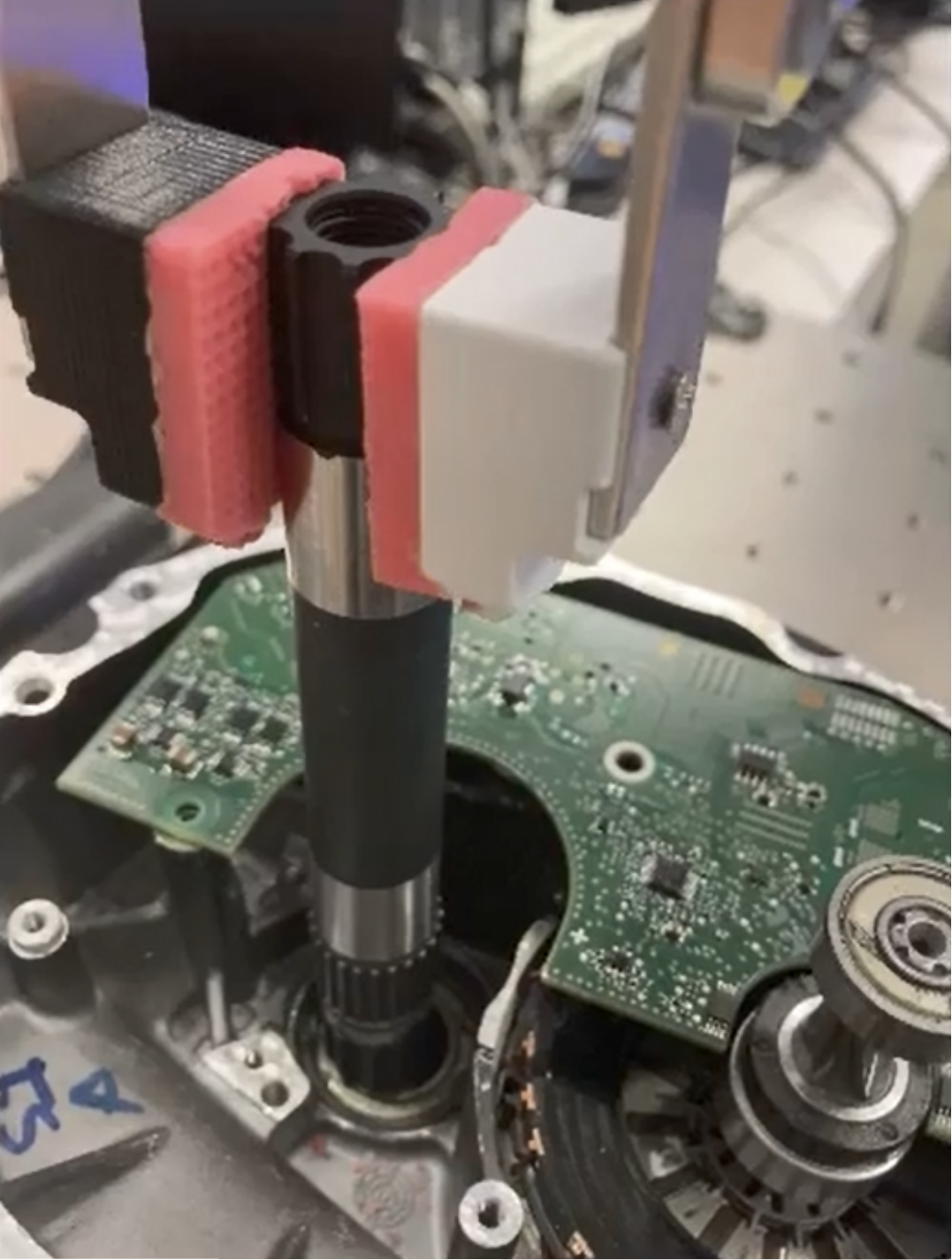}
	\caption{Snapshots of the \texttt{insert\_shaft} skill, including close-ups that show the complexity of the insertion process.}
	\label{fig:shaft_insertion}
\end{figure}

The Franka Emika Panda robot has $7$ DoF and is capable of direct kinesthetic teaching.
The arm is extended by a \emph{wrist-mounted} ATI FT45 sensor 
and a parallel gripper for the assembly task.
As shown in Fig.~\ref{fig:assembly-snapshots}, the workstation consists of the loading platform 
where components are picked and the assembly station where the components are assembled together. 
For learned skills, between $2$ and $4$ demonstrations were performed to specify the desired motions. In the following, we refer to the skills listed in Tab.~\ref{table:assembly-skills}.
The individual sub-tasks of the assembly sequence are prepared as follows:

\begin{itemize}
	\item \emph{[Press-PCB]}: 
		This task was solved with a single \texttt{press\_pcb} skill which consisted of pressing down on the PCB board in three different locations.
		
	\item \emph{[Mount-Gear]}: 
		Here, we used two skills - \texttt{grasp\_gear} and  \texttt{mount\_gear}. 
		
	\item \emph{[Insert-Shaft]}: 
	This is a complex sub-task which we addressed with two skills for picking and re-orienting the shaft
	(\texttt{pick\_shaft}, \texttt{orient\_shaft}), followed by re-grasping and insertion (\texttt{insert\_shaft}). The
	latter skill, shown in Fig.~\ref{fig:shaft_insertion}, encompasses a simultaneous pushing and \textquote{wiggling} motion. Furthermore, to successfully execute this skill,
	the self-supervised refinement strategy described in~\cref{subsec:skills_refinement} was used.
	
	\item \emph{[Slide-Peg]}:	
	Analougus to the previous task, this problem was solved with three skills for picking, re-orienting, re-grasping and
	insertion (\texttt{pick\_peg}, \texttt{orient\_peg}, \texttt{slide\_peg}).	
\end{itemize}

Note that the shaft and peg are placed freely on the loading platform. 
This requires to re-orient them from lying-flat to standing into a fixture, 
such that they can be subsequently grasped from the top, as shown in Fig.~\ref{fig:assembly-snapshots}.
In other words, such re-orientation skill allows the robot to reduce the sub-tasks uncertainty by taking the objects from arbitrary poses to a fixed fixture whose configuration is more precisely known. 
We also leveraged skill refinement for reducing undesired contact forces during re-orientation.

\subsubsection{Assembly Task Results}\label{subsubsec:assembly-results}

\begin{figure}[t]
	\centering
	\includegraphics[width=0.495\textwidth]{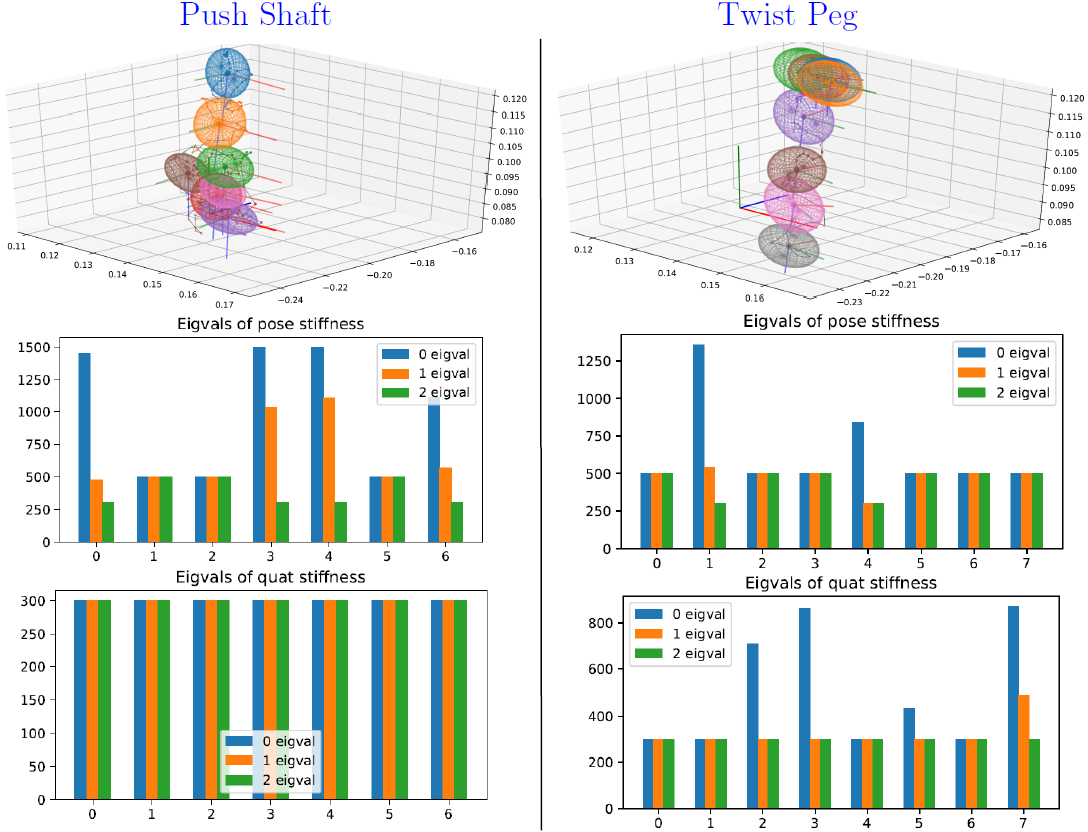}
	\caption{The learned attractor model (\textbf{top}), 
		the optimized translation stiffness (\textbf{middle}) 
		and the optimized angular stiffness (\textbf{bottom}), 
		for the last stage of push-shaft skill (\textbf{left}) and twist peg (\textbf{right}).}
	\label{fig:stiffness-exp}
	\vspace{-0.4cm}
\end{figure}

\begin{figure}[t]
	\centering
	\includegraphics[width=0.3\textwidth]{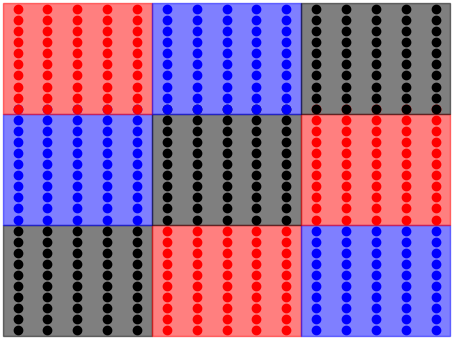}
	\caption{Choice of different branches for the picking skill (three branches in total, differentiated by color) in the assembly use case, given different positions of the object. The plot represents the area of the loading platform where three different skill instances (branches) can be activated.}
	\label{fig:branch_cap}
	\vspace{-0.4cm}
\end{figure}

\begin{table}[t]
	\centering
\begin{adjustbox}{height=0.2\linewidth}
\begin{tabular}{ccccccc}
\toprule
Skill Name & $M$ & $T[\si{\second}]$  & $N$ &  $\mathsf{TP}$ & $t\, (\boldsymbol{\Theta}_y\,\vert \, \boldsymbol{K}^\star)$ [\si{\second}]\\ \midrule
\texttt{grasp\_gear} & $3$ & $1.2$ & $8$ &  $\{\mathfrak{r},\mathfrak{o}\}$ & $3\, \vert\,  - $\\
\texttt{mount\_gear} & $3$ & $4.8$ & $18$ & $\{\mathfrak{r}, \mathfrak{g}\}$  & $14\, \vert\,  12 $\\
\texttt{pick\_shaft} & $3$ & $1.3$ & $7$ &  $\{\mathfrak{r}, \mathfrak{o}\}$ & $4\, \vert\,  - $\\
\texttt{orient\_shaft} & $3$ & $1.8$ & $12$ &  $\{\mathfrak{r}, \mathfrak{o}\}$ & $5\, \vert\,  - $\\
\texttt{insert\_shaft} & $2$ & $5.7$ & $23$ &  $\{\mathfrak{r},\mathfrak{g}\}$ & $20\, \vert\,  18 $ \\
\texttt{pick\_peg} & $3$ & $2$ & $10$ &  $\{\mathfrak{r}, \mathfrak{g}\}$ & $3\, \vert\,  - $ \\ 
\texttt{orient\_peg} & $3$ & $1.8$ & $12$ &  $\{\mathfrak{r}, \mathfrak{o}\}$ & $5\, \vert\,  - $\\
\texttt{slide\_peg} & $2$ & $5.3$ & $24$ &  $\{\mathfrak{r}, \mathfrak{o}\}$  & $18\, \vert\,  15 $ \\
\texttt{press\_pcb} & $4$ & $6$ & $22$ & $\{\mathfrak{r}, \mathfrak{g}\}$  & $15\, \vert\,  10 $ \\
\bottomrule
\end{tabular}
\end{adjustbox}
\caption{For each skill, the number of demonstrations $M$, the trajectory length $T$, 
number of components $N$, choice of task parameters $\mathsf{TP}$, 
and the training time for $\boldsymbol{\Theta}_y$ and $\boldsymbol{K}^\star$ (for forceful skills). 
Note that $\mathfrak{r},\mathfrak{g}, \mathfrak{o}$ are the robot, global, object frame,  respectively. }
\label{table:assembly-skills}
\end{table}
As shown in Table~\ref{table:assembly-skills}, some skills such as picking and re-orientation do not involve force tracking, while other skills such as insertion, pressing and sliding require specific force profiles during execution. 
Consequently, as described in~\cref{subsubsec:force-lfd}, an attractor model is learned for such skills along with the optimal stiffness. 
Figure~\ref{fig:stiffness-exp} shows the optimized translational and rotational stiffness for the \texttt{insert\_shaft} and \texttt{slide\_peg} skills. 
It can be seen that insertion requires relatively high translational stiffness upon contact while twisting requires high rotational stiffness.  
Furthermore, some skills such as picking and re-orientation contain multiple branches due to different orientations of the object.
Then, as described in~\cref{subsubsec:skill-models}, a condition model is learned to select the correct branch in new scenarios. 
An example of the picking skill is shown in Fig.~\ref{fig:branch_cap}.

\begin{figure}[t]
	\centering
	\includegraphics[width=0.49\textwidth]{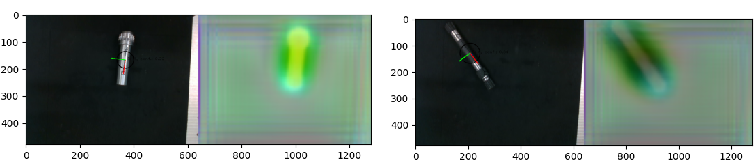}
	\caption{Visualization of the 6D pose estimation via DON for the shaft (\textbf{left}) and peg (\textbf{right}) in the assembly use case.}
	\label{fig:don-exp}
	\vspace{-0.4cm}
\end{figure}

During skill reproduction, for kinematic skills such as picking and re-orientation, the 6-D poses of the shaft and the peg are obtained via the trained DON, as shown in Fig.~\ref{fig:don-exp}.
For force-based skills, the learned attractor model is applied online provided with the observed robot positions and external forces. 
Table~\ref{table:results} shows the success rate of executing each skill with $10$ repetitions.
The \emph{Press-PCB} and \emph{Mount-Gear} sub-tasks are reliably reproduced without any manual tuning of the associated task parameters, while a small shift in the object pose is needed for the \emph{Insert-shaft} and \emph{Slide-Peg} sub-tasks to compensate for the tracking error of the underlying impedance controller.
This is achieved by leveraging the skill optimization processes described in~\cref{subsec:skills_refinement}. 
Note this such skill refinement is necessary to overcome the tracking errors caused by robot model mismatch during identification, and the internal joint force or torque control mechanism. 
Figure~\ref{fig:press-pcb-traj} highlights the differences between the executed trajectory, the reference attractor trajectory, and the demonstrations. 
In addition, it also shows that the exerted force profile during execution matches the demonstrations accurately.  

\subsubsection{Benchmarks}\label{subsubsec:assembly-benchmarks}
Our proposed framework is compared against three main baselines for the aforementioned $4$ sub-tasks:
The direct replay of the demonstration (demo-replay), the standard kinematic skill model (pose-based) as proposed in~\cite{RozoEtal:IROS20}, and manually-tuned skills (manual) by following the procedure proposed in~\cite{johannsmeier2019framework}. 
Pose-based methods simply ignore the force profile when learning the skill model, while 
the manual method demands tuning by an operator of \emph{all} fixed reference trajectory and the associated stiffness carried out. 

\begin{figure}[t]
	\centering
	\includegraphics[width=0.99\linewidth]{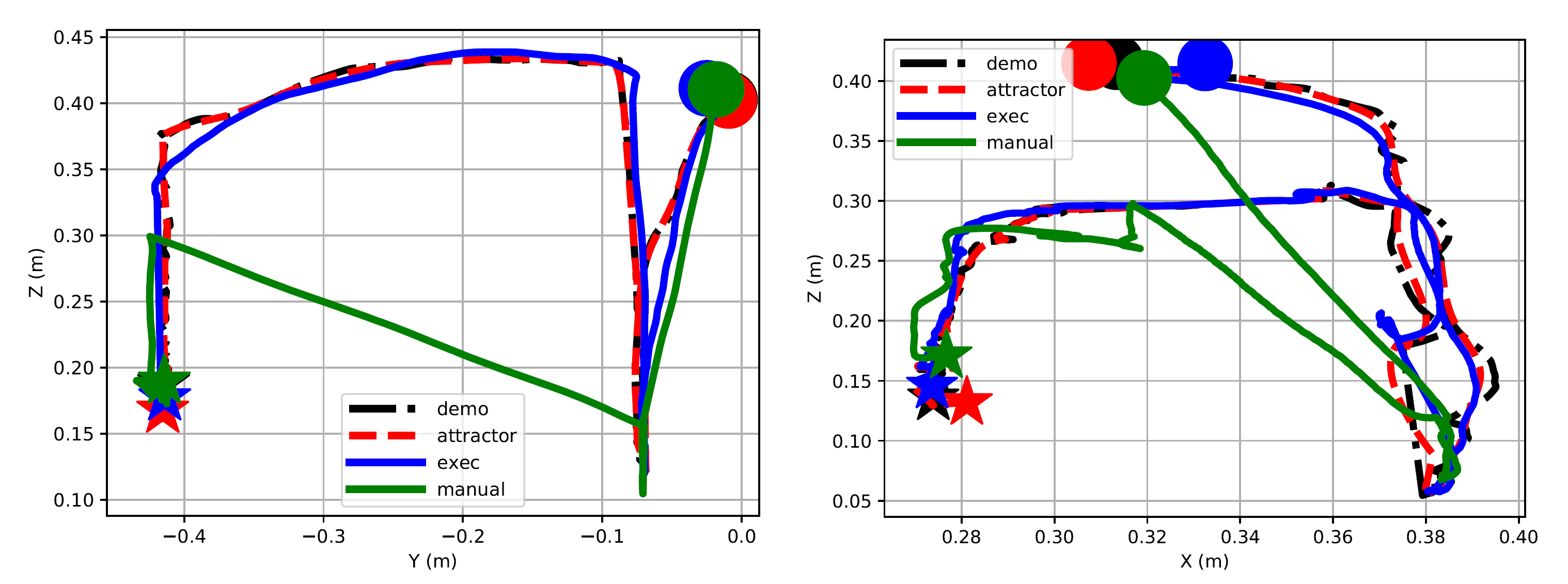}
	\caption{Comparison of executed trajectories via demo-replay, manual tunning and our method, for the Insert-Shaft skill (\textbf{left}) projected on $y-z$ plane and Slide-Peg skill (\textbf{right}) projected on $x-z$ plane.}
	\label{fig:compare}
	\vspace{-0.1cm}
\end{figure}

The resulting success rate is summarized in Table~\ref{table:results}. 
First, the success rate of demo-replay and pose-based methods are quite low for all four skills, especially when delicate force interaction is required.
Often, via both methods, the skill execution simply reaches the key reference points without exerting the desired forces, e.g., the robot only touches the pins without any pushing force during \emph{Press-PCB}. 
Moreover, raw human demonstrations are quite \emph{shaky} in general, as shown in Fig.~\ref{fig:compare}. 
Replaying such small unnecessary movements is often harmful for the execution, which makes demo-replay reproductions ineffective. 
Lastly, the manual tuning in the \texttt{insert\_shaft} and \texttt{slide\_peg} skills leads to rather reliable executions. 
However, the time taken to program such skills is \emph{significantly} longer than our self-refinement strategy (hours vs. minutes empirically). 
In addition, the resulting trajectory often follows a zig-zag pattern with harsh transitions due to linear interpolation between manually-chosen waypoints, as shown in Fig.~\ref{fig:compare}.

\subsection{Bin Sorting Task}\label{subsec:bin-sorting}

\begin{figure*}[t]
	\centering
	\includegraphics[height=0.195\linewidth]{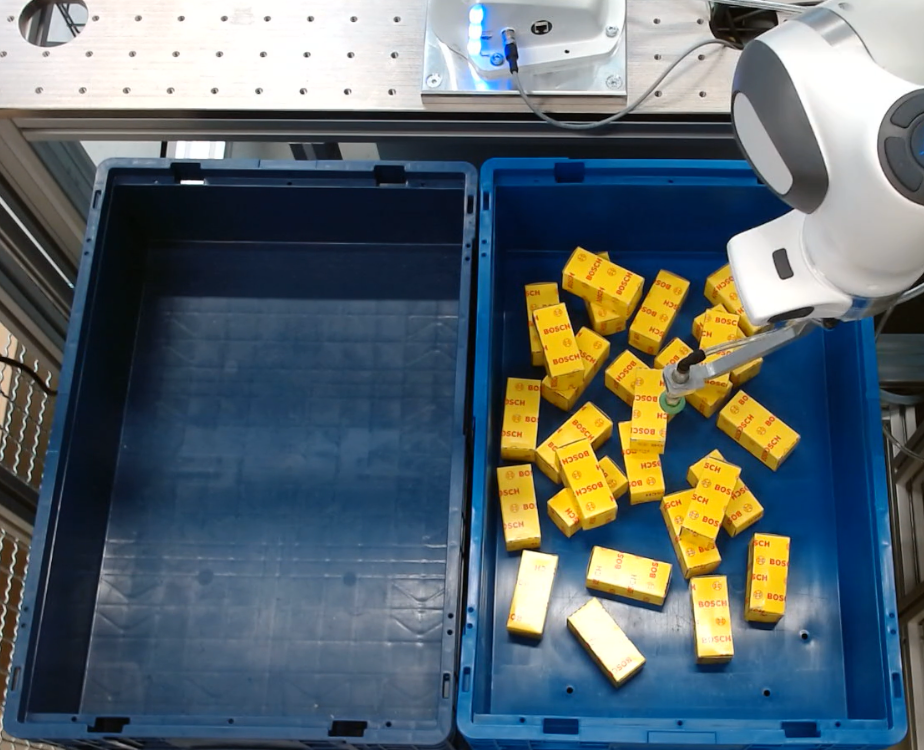}
	\includegraphics[height=0.195\linewidth]{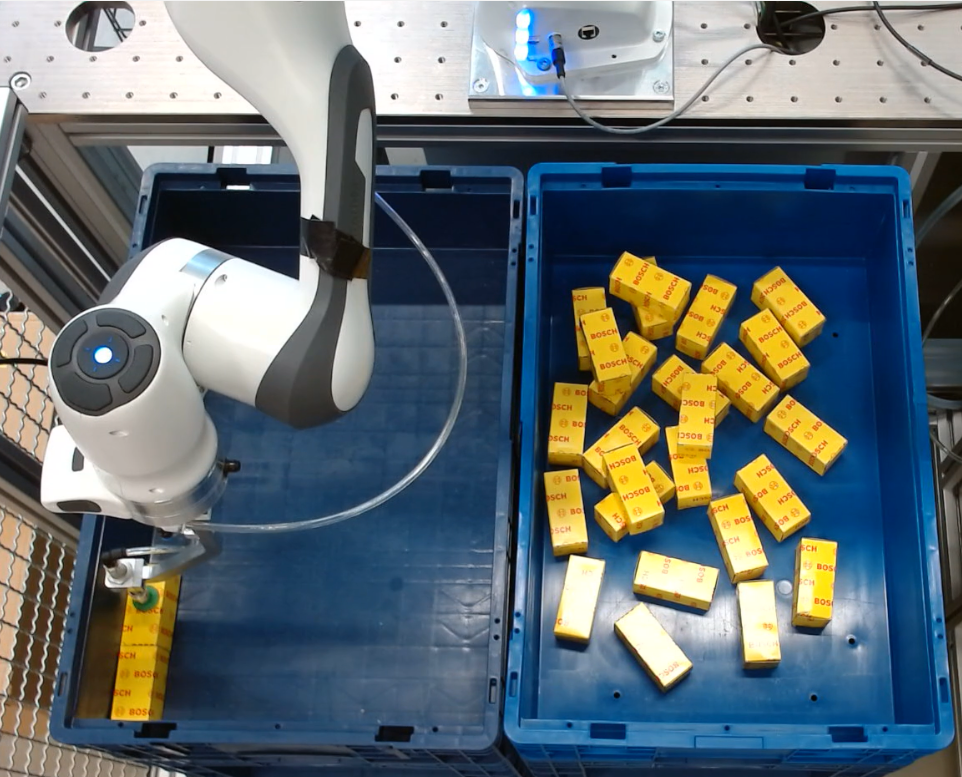}
	\includegraphics[height=0.195\linewidth]{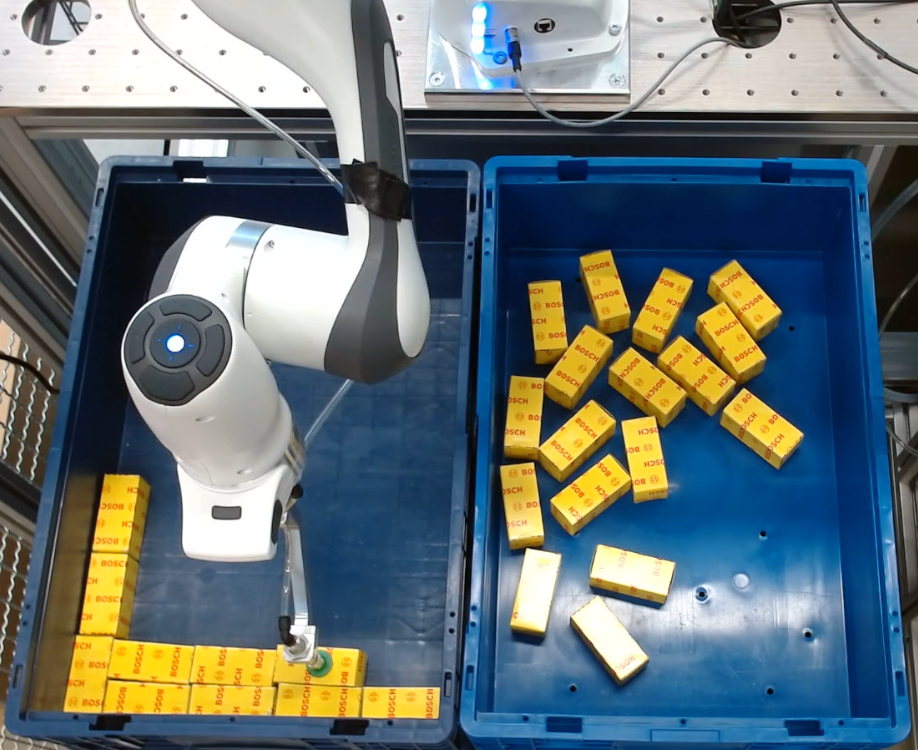}
	\includegraphics[height=0.195\linewidth]{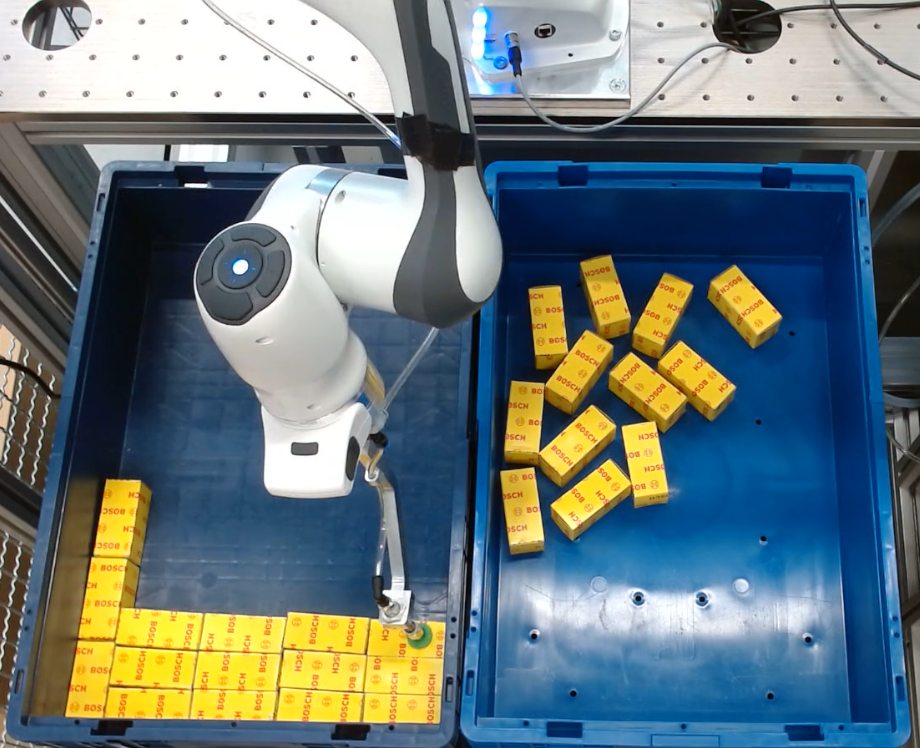}
	\caption{Snapshots for the bin sorting task: picking unsorted boxes (\textbf{left}), insertion and sorting via different maneuvers (\textbf{right} three).}
	\label{fig:bin-snapshots}
\end{figure*}

To demonstrate the transferability of our framework across different industrial use cases,
we consider in the following a second application that typically occurs in indoor logistic systems of warehouses.
In this use case, unsorted boxes arrive in a bin and should be picked and sorted 
into a packing formation as shown in Fig.~\ref{fig:bin-snapshots}. 
To achieve a dense packing, the boxes should be picked with different orientations when close to the bin boundaries
and placed with different maneuvers for different sorting formations. 
More specifically, the first row of the formation has to be placed by rotating the object $90^{\circ}$, while the last row should be inserted by slight tilting and wiggling due to tight space.
We used the same $7$-DoF Franka Emika Panda robot, this time extended by a suction gripper. 

\begin{table}[t]
	\begin{center}
		\begin{adjustbox}{width=0.95\linewidth}
			\begin{tabular}{ccccc}
				\toprule
				Methods & Press-PCB & Mount-Gear & Insert-Shaft & Slide-Peg \\
				\midrule
				Pose-based & 0 & 3 & 0 & 1 \\
				Demo-replay & 0 & 4 & 0 & 2 \\
				\textbf{Ours} & \textbf{9} & \textbf{10} & \textbf{8} & \textbf{9} \\
				Manual & N/A & N/A & 9 & 9 \\
				\bottomrule
			\end{tabular}
		\end{adjustbox}
		\caption{Success rate of four different methods out of $10$ repeated executions. 
			Manual skills are not programmed for the first two skills. }
		\label{table:results}
	\end{center}
	\vspace{-0.35cm}
\end{table}

\begin{figure}[t]
	\centering
	\includegraphics[width=0.95\linewidth]{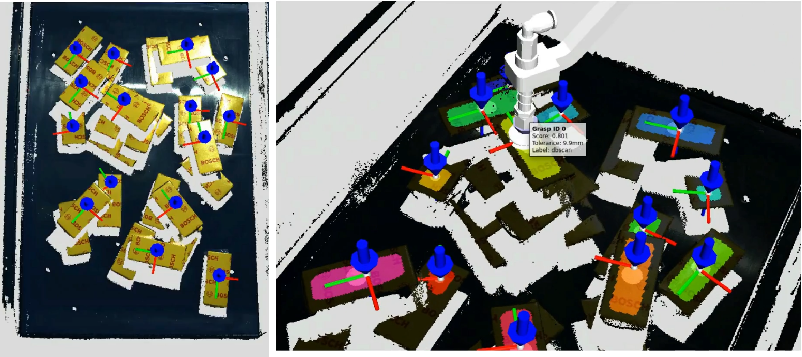}
	\caption{Grasp pose results in the bin picking use case.}
	\label{fig:dexnet-exp}
	\vspace{-0.1cm}
\end{figure}

First of all, to obtain reliable predication of the grasp poses, 
we rely in the shown experiment on a dense pixel-wise graspability estimation with RGB-D input,
based on UNet~\cite{ronneberger2015u}, which is trained on annotated pixel-wise labels of expected graspability.
Results from the perception system are shown in Fig.~\ref{fig:dexnet-exp}, 
where the candidate grasping pose with the highest score is chosen.

\begin{figure}[t]
	\centering
	\includegraphics[width=0.3\textwidth]{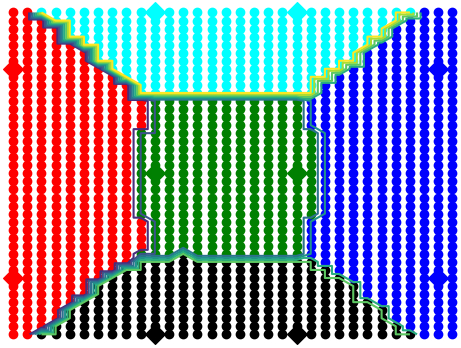}
	\caption{Choice of different branches for the picking skill (five branches in total) in the bin sorting use case, given different positions of the object. Each branch is characterized by a different color. 
		The plot represents the area within the bin where five different skill instances (branches) are activated. 
		The target grasping pose is used to compute the task parameter for the picking skill.}
	\label{fig:branch_box}
	\vspace{-0.4cm}
\end{figure}

In order to learn and execute this bin sorting task, we used the same system described in~\cref{sec:system}, except the perception method for object pose estimation, which was here replaced by the grasp-pose based approach.
For controlling the robot motions, two different skills were demonstrated, one for picking and one for placing.
Following the same learning strategy as described in~\cref{subsec:LfD}, the skill to pick boxes from the bin is taught such that different trajectories (i.e., skill instances) are taken for different boundaries of the bin. 
Figure~\ref{fig:branch_box} shows the area within the bin where five different branches, corresponding to the skill instances, are activated. 
The target grasping pose is used to compute the task parameter for the picking skill. 

For the placing skill, three branches (i.e., skill instances) in total are demonstrated for the first column, middle columns, and the last column respectively, as shown in Fig.~\ref{fig:bin-snapshots}. 
One particular challenge for this placing skill is that all trajectories need to be robust to uncertainties in the grasping position, originating from perception and from motion control during picking.
Note that the potential of LfD is again leveraged in this skill as we need to demonstrate a wiggling motion to fit the box in the tight space left for the last placing maneuver.
Hard coding such a motion is a clearly tedious process.
It is also worth mentioning that with three or less demonstrations for each branch, the placing skill can be reproduced reliably and thus, suffice to complete the whole bin sorting task.

	\section{Discussion}
	\label{sec:discussion}
	When developing an integrated learning system such as the manipulation framework presented in this paper, \emph{adaptability} and \emph{explainability} are important aspects to consider when facing unstructured and variable environments~\cite{stoica2017berkeley,amodei2016concrete,rajan2017towards}.
Consequently, we discuss both aspects in the sequel.
Still, despite the satisfactory advancements in integrating control, learning and planning into a single framework for advanced industrial robotic manipulation tasks, the complexity of flexible manufacturing settings gives rise to several issues that are discussed as well in the remainder of this section.

\subsection{Adaptability}
Adaptability considers the ability of the robot to adjust the task execution to the currently observed environment. 
We addressed this topic at multiple levels of our framework.
Firstly, each motion skill can individually adapt to object pose changes due to the task-parameterized formulation of the learning model presented in~\cref{subsec:LfD}.
For example, when demonstrating how to pick an object like the metal cap, the motion is learned and generalized with respect to the pose of the cap and thus, applicable for a wide range of different cap poses.
Secondly, when observing that some cases are executed less accurately, those can be improved by additional demonstrations or through data-efficient refinement based on BO.
In this regard, we consider that robot motion skills learning and refinement are technologies that can be readily used for a wide range of weakly-structured industrial settings, which seldome demand online adaptation. 

When sequencing multiple skills, our approaches described in~\cref{subsec:sequencing,subsec:reproduction} ensured that also previous or subsequent skills are adjusted whenever the robot needs to adapt e.g., by exploiting the learned condition models.
From a software perspective, such an adaptation is made possible by separating the selection of skills, as commanded by the \emph{skill interface}, from the management of perceived object data.
Our skills sequencing approach was built on the assumption that the task plan is provided by an operator. 
We believe this is a realistic assumption, as a plethora of industrial processes are still being performed by executing a sequence of sub-tasks predefined by human operators.
We argue that such task domain knowledge should be leveraged instead of discarded when it comes to integrate learning systems in industry.

Finally, at a task level, adaptation is supported by the FlexBE framework.
This enables us to formulate recovery behaviors for various sorts of failures like the two cases described in~\cref{sec:experiments}.
Facilitated by the layered structure of our software system presented in~\cref{subsec:software}, different motion skills and primitives can be easily composed to address different sorts of tasks.
During the setup phase of a new task, composition at the task level can be adjusted online during execution to best incorporate all variations of the environment.

\subsection{Explainability}
Explainability considers how the system facilitates that a user or expert understands the past and future actions taken by the robot.
A strong contribution to such an insight into our system was provided by the predictability resulting from the demonstrated motion skills (\cref{subsec:LfD}).
When a robot learns an end-to-end policy, it is often hard to predict which actions a certain input will lead to.
Even in the case of manually-defined motion planners, many sampling-based approaches like the RRT family might result in unexpected motions~\citep{Karaman11:SamplingMotion}.
In contrast, LfD approaches such as ours allow a user to ensure that motions are similar to what is expected, i.e., what was demonstrated.

In addition to expecting a certain trajectory profile, the specific trajectory obtained during execution (\cref{subsec:reproduction}) was visualized in advance for possible inspection. 
Moreover, each segment of the continuous trajectory can be associated with a discrete component of the skill, adding a possible semantic notion.
Similarly, the progress of the task execution was visualized in terms of the current skill, as well as the past and remaining parts of the task.
All the aforementioned features facilitate monitoring labor during human supervision. 

Note that the foregoing advantages naturally arise when leveraging two types of inductive bias in our robotic manipulation framework: Human domain knowledge and geometry. 
The former is mainly exploited when learning the robot motion skills and when providing the high-level plan of the task, which the skills sequence schema builds on. 
The latter is employed to formulate proper statistical models, trajectory optimizers, and controllers for full-pose robot end-effector motions, which make our framework not only technically sound, but also more reliable when learning and executing robot manipulation tasks. 

Last but not least, the modular approach taken in our system architecture (as illustrated in Figs.~\ref{fig:overview} and~\ref{fig:system-diagram}) allows us to debug and improve each core module independently when needed.
For example, we could evaluate if a reference trajectory retrieved by our skill model could be better tracked by the robot Cartesian impedance controller when using model parameters estimated by our model identification method.
We argue that this modular analysis and improvement may be harder to perform when using end-to-end frameworks. 

\subsection{Current limitations}
\subsubsection{Skills sequencing} 
When sequencing several skills for a specific task, the computation time of our approach grows combinatorially in the number of \emph{final} states for each skill along the sequence.
In other words, the more branches there are in one skill, the more time consuming to compose it with other skills.
This is because all branches should be compared in terms of likelihood for both the current skill and the subsequent one, as explained in~\cref{subsec:sequencing}.
Therefore our framework, at its present state, cannot be used for real-time replanning.
Nonetheless, several methods can be used to accelerate this process:
\emph{(i)} as done here, the skill sequence can often be decomposed in independent sub-sequences, which can be planned separately;
\emph{(ii)} prune early-on transitions whose probability is below a threshold both in the skill model and the transition probability matrix;
\emph{(iii)} construct $\widehat{\mathbf{\Theta}}_{\boldsymbol{\mathsf{a}}}$ \emph{specifically} for the given initial and final states, i.e., combine the computed transition probability with forward and backward observation probability.
Secondly, as discussed in our previous work~\cite{Schwenkel2019Optimizing}, certain skills may be formulated with \emph{free} task parameters, which could be optimized in a general way for various tasks.
This issue is not addressed here since all task parameters are assumed to be attached to physical objects in the scene.
Combination of these two methods will be addressed in future research.

\subsubsection{Open-loop skill execution} Our framework did not consider settings where the enviroment conditions may dynamically change once the robot workspace scanning and subsequent skill execution have been started.
This assumption worked successfully for the use cases reported in this paper. 
However, open-loop skill execution naturally comes at the cost of compromising adaptability in highly-varying settings.
In these cases, close-loop approaches are preferred, but these demand to endow the robot with rich and redundant sensory systems composed of, for example, multiple and in-hand cameras to track the state of the objects of interest.
Such sensory systems arise interesting research challenges not addressed in this paper such as fast multi-view object pose estimation, online robot skill adaptation, among others.

\subsection{Future directions}
As briefly discussed previously, a more flexible manipulation framework should deal with settings where objects move during the execution of a skill, e.g., while the robot approaches an object, as shown during one of our failure experiments.
In this case, the perception system should be able to track the objects of interest in (near) real-time. 
Based on the new object location, the robot needs to adapt its reference trajectory by first recomputing the sequence of the most-likely states and then generating an adapted motion using our Riemannian optimal controller.   
In this context, our current efforts are devoted to develop deep learning techniques for $6$D pose estimation and tracking of industrial objects using multiple cameras. 

However, a multi-view setting may not suffice when the grasped object slightly slides in the robot gripper, changing its relative pose w.r.t the robot end-effector. 
Detecting such changes may reduce failures as the robot may be able to quickly adapt the skill execution accordingly.
To do so, we plan to explore in-hand camera settings~\cite{Chen11:ActiveVisionSurvey} and tactile sensing~\cite{Lambeta20:DIGITsensor} to track the object pose while grasped more accurately.
This multi-modal tracking also demands to have skills that can be adapted in an online manner as needed.

	\section{Conclusions}
	\label{sec:conclusions}
	This paper introduced a robotic manipulation framework for flexible manufacturing systems. 
Its main components were built on the synergy of learning from demonstration, skills sequencing, optimal control and black-box optimization. 
Our system architecture was designed as a combination of software layers and modules that made possible the easy integration and interaction of different hardware components, algorithms, controllers, and user interfaces.
We successfully tested our framework on different subtasks of a real e-Bike assembly process and on a dense box-packing setup, where the robot properly sequenced and executed the necessary motion skills and action primitives to reach the given task goal.
We demonstrated that our FlexBE-based framework allowed us to monitor the robot execution to detect failures and to re-plan the task according to the updated environment state. 
We concluded this work by highlighting how adaptability and explainability are core features in our robotic manipulation framework, and discussed its current limitations and future extensions.

	\appendix
	\section{Notation}
\label{app:notation}
Table~\ref{tab:tech_notation} provides the notation employed to describe the most important variables and parameters used in the technical description of each of the components of our framework. 

\begin{table}[h]
	\footnotesize
	\centering
	\begin{tabularx}{\linewidth}{|l|X|}
		\rowcolor{lightgray}
		\hline
		\textbf{Symbol} & \textbf{Description} \\ [0.3ex] 
		\hline
		$\mathcal{M}$ & Riemannian manifold \\
		$\mathcal{T}_{\bm{\mathrm{p}}}\mathcal{M}$ & Tangent space of manifold $\mathcal{M}$ at $\bm{\mathrm{p}} \in \mathcal{M}$\\
		$\mathbb{R}^3$ & Three-dimensional Euclidean space \\
		$\mathcal{S}^3$ & $3$-sphere \\
		$\operatorname{SO}(3)$ & Special orthonormal rotation group\\
		$\operatorname{SE}(3)$ & Special Euclidean group of rigid motions\\
		$\vec \theta$ & Robot joint configuration in $\mathbb{R}^7$\\
		$\vec{x}_{\mathsf{e}}$ & End-effector pose in $\mathbb{R}^3 \times \mathcal{S}^3$\\
		$\vec{X}_{\mathsf{e}}$ & End-effector pose in $\operatorname{SE}(3)$\\
		$\vec{X}_{\mathsf{e},\mathsf{d}}$ & Desired end-effector pose in $\operatorname{SE}(3)$\\
		$\hat{\vec{c}}$ & End-effector twist in $\mathfrak{se}(3)$\\
		$\vec R$ & End-effector rotation matrix in $\operatorname{SO}(3)$\\
		$\vec p$ & End-effector position vector in $\mathbb{R}^3$\\
		$\vec \omega$ & End-effector angular velocity in $\mathbb{R}^3$\\
		$\vec v$ & End-effector linear velocity in $\mathbb{R}^3$\\
		$\bm{\Psi}$ & Barycentric parameters of the robot dynamics model\\
		$\vec{x}_{\mathsf{o}_j}$ & Pose of object $\mathsf{o}_j$ in $\mathbb{R}^3 \times \mathcal{S}^3$ \\
		$\vec{x}_{\mathsf{s}}$ & System state \\
		$\mathsf{A}$ & Set of demonstrated skills \\ 
		$\mathsf{O}_{\mathsf{a}}$ & Set of objects involved in skill $\mathsf{a}_h$ \\
		$\mathsf{D}_{\mathsf{a}}$ & Set of available demonstrations for skill $\mathsf{a}_h$ \\ 
		$\mathcal{N}(\bm{\mu}, \bm{\Sigma})$ & Gaussian distribution with mean $\bm{\mu}\in\mathbb{R}^n$ and covariance~$\bm{\Sigma} ·\in \mathbb{R}^{n\times n}$\\
		$\mathcal{N}_{\mathcal{M}}(\bm{\mu}, \bm{\Sigma})$ & Riemannian Gaussian distribution with mean $\bm{\mu} \in \mathcal{M}$ and covariance $\bm{\Sigma} \in \mathcal{T}_{\bm{\mu}}\mathcal{M}$\\
		$\bm{\Theta}$ & Parameters set of a TP-HSMM\\
		$\text{Exp}_{\bm{\mathrm{p}}}(\cdot)$ & Exponential map at $\bm{\mathrm{p}}\in \mathcal{M} $ \\
		$\text{Log}_{\bm{\mathrm{p}}}(\cdot)$ & Logarithmic map at $\bm{\mathrm{p}}\in \mathcal{M} $ \\
		$\transp{\bm{\mathrm{p}}}{\bm{\mathrm{q}}}$ & Parallel transport from $\mathcal{T}_{\bm{\mathrm{p}}}\mathcal{M}$ to $\mathcal{T}_{\bm{\mathrm{q}}}\mathcal{M}$ \\
		$\operatorname{Ad}_{\vec X}$ & Adjoint operator at $\vec X$\\
		\hline
	\end{tabularx}
	\caption{Notation}
	\label{tab:tech_notation}
\end{table}

\section{Impedance Control}
\label{app:robot_control}
\subsubsection{Lie groups}
We consider elements of the matrix Lie group $\operatorname{SE}(3)$ to represent frame configurations which can be written as homogeneous transformation matrices of the form
$$\mathbf{X}=\begin{bmatrix}
		\mathbf{R} & \mathbf{p} \\
		0          & 1
	\end{bmatrix},
$$
where $\mathbf{R} \in \operatorname{SO}(3)$ represents the orientation and $\mathbf{p} \in \mathbb{R}^3$ the translation. The tangent vector space at the identity of $\operatorname{SE}(3)$ is called the Lie algebra $\mathfrak{se}(3)$ which contains twists
$$\widehat{\mathbf{c}}=\begin{bmatrix}
		[\boldsymbol{\omega}]_{\times} & \mathbf{v} \\
		0                              & 0          \\
	\end{bmatrix},$$
where $[\boldsymbol{\omega}]_{\times} \in \mathfrak{so}(3)$ denotes an angular velocity in skew-symmetric matrix form and $\mathbf{v}\in\mathbb{R}^3$ a
translational velocity. 
As $\mathfrak{se}(3)$ is isomorphic to $\mathbb{R}^6$, the twist $\widehat{\mathbf{x}}$ can also be specified via
its coordinate vector $\mathbf{x}=[\mathbf{v}^{\trsp}, \boldsymbol{\omega}^{\trsp}]^{\trsp}$. 
Furthermore, we respectively denote the adjoint operator and the matrix commutator
(i.e., the Lie bracket of $\operatorname{SE}(3)$) as
$$\operatorname{Ad}_{\mathbf{X}}=\begin{bmatrix}
		\mathbf{R} & [\mathbf{p}]_{\times}\mathbf{R} \\
		0          & \mathbf{R}                      \\
	\end{bmatrix} \;\; \mbox{and} \;\; \operatorname{ad}_{\widehat{\mathbf{x}}}=
	\begin{bmatrix}
		[\boldsymbol{\omega}]_{\times} & [\mathbf{v}]_{\times}          \\
		0                              & [\boldsymbol{\omega}]_{\times} \\
	\end{bmatrix}.$$

Via $\operatorname{Ad}_{\mathbf{X}}$, Cartesian vectors from the tangent space at $\mathbf{X}$ can be mapped to the identity. A physical interpretation
of the adjoint mapping is the transformation of generalized velocities under a change of reference frame given by $\mathbf{X}$.

The Lie logarithmic map allows to exactly transfer an element from the $\operatorname{SE}(3)$ manifold to $\mathfrak{se}(3)$, its tangent
space at the identity. 
In this work, we use the capitalized logarithmic map which maps elements $\mathbf{X}$ directly to
vectors $\mathbf{x}$~\cite{Sola18}.

$$\mbox{Log}: \quad \operatorname{SE}(3) \rightarrow \mathbb{R}^6; \quad \mathbf{X} \rightarrow \mathbf{x}$$.

The vector $\mathbf{x}=\mbox{Log}(\mathbf{X})$ is referred to as the exponential coordinates of $\mathbf{X}$ and can be interpreted as parametrizing a constant twist which moves a frame from the identity to configuration $\mathbf{X}$ in one unit of time.

\subsection{Cartesian Impedance Control on the Lie group $\operatorname{SE}(3)$}
\label{app:CIC}
In order to formulate the tracking control law, we denote the current end-effector frame and the desired end-effector
frame configurations with respect to any given inertial frame as~$\mathbf{X}_e \mathbf{X}_{e,d} \in \operatorname{SE}(3)$ respectively. 
Correspondingly, the associated generalized body velocities
are defined by the twist coordinate vectors $\mathbf{c, c_d} \in \mathbb{R}^6$ with the respective accelerations being $\dot{\mathbf{c}},
\dot{\mathbf{c}}_d\in\mathbb{R}^6$. Note that we use left-trivialized velocities in this work, which represent body
twists in the corresponding body coordinate frame, i.e., $\widehat{\mathbf{c}}={\mathbf{X}_e}^{-1}\dot{\mathbf{X}}_e$ (see~\cite{Trav16} for details). 
Following~\cite{Bull95}, we define the natural tracking error in terms of the $\operatorname{SE}(3)$ group
operation as
$\mathbf{E}\triangleq {\mathbf{X}_e}^{-1}\mathbf{X}_{e,d}$, i.e., as the desired end-effector frame expressed in the current end-effector frame.
Accordingly, the adjoint operator which maps elements from the tangent space at $\mathbf{X}_{e,d}$ to the Lie algebra at $\mathbf{X}_e$ is
given by $\operatorname{Ad}_{\mathbf{E}}=\operatorname{Ad}^{-1}_{\mathbf{C}}\operatorname{Ad}_{\mathbf{C}_d}$. 
Considering that $\frac{d}{dt}(\operatorname{Ad}_{\mathbf{E}})=\operatorname{Ad}_{\mathbf{E}}\operatorname{ad}_{\mathbf{E}^{-1}\dot{\mathbf{E}}}$ as in~\cite{Trav16}, allows us to write the error velocity and its derivative as
\begin{align}
	\mathbf{e}&=\mathbf{c}-\operatorname{Ad}_{\mathbf{E}} \mathbf{c}_d\nonumber ,\\
	\dot{\mathbf{e}}&=\dot{\mathbf{c}}-\operatorname{Ad}_{\mathbf{E}}\dot{\mathbf{c}}_d-\operatorname{Ad}_{\mathbf{E}}\operatorname{ad}_{\widehat{\mathbf{e}}} . \mathbf{c}_d \label{eq:dot_e}
\end{align}
We employ a classical acceleration-resolved Cartesian impedance control scheme, where the goal is to impose error dynamics of the form
\begin{align}
	Z(\mathbf{E,e},\mathbf{f}_{ext})&=\mathbf{M}_d^{-1}\left(\mathbf{f}_{ext} +\mathbf{K} \; \text{Log}(\mathbf{E}) - \mathbf{B e} - \operatorname{Ad}_{\mathbf{E}^{-1}}^{\trsp} \mathbf{f}_{ff} \right), \label{eq:error_dynamics}
\end{align}
where $\mathbf{M}_d, \mathbf{K, B} \in \mathbb{R}^{6 \times 6}$ are positive definite matrices allowing to shape the desired inertia, stiffness
and damping properties in Cartesian space. Note the use of exponential coordinates in~\eqref{eq:error_dynamics} to express
the error in the Lie algebra~\cite{Bull95}. The optional feedforward wrench $\mathbf{f}_{ff}$ can be designed, e.g., to
maintain downward pressure in insertion tasks. 

Using~\eqref{eq:error_dynamics}, it is easy to see that a control input function $\mathbf{u}\triangleq \dot{\mathbf{c}}$ of the form 
\begin{align}
	\mathbf{u}&=Z(\mathbf{E,e},\mathbf{f}_{ext})+\operatorname{Ad}_{\mathbf{E}}\left(\dot{\mathbf{c}}_d+\operatorname{ad}_{\widehat{\mathbf{e}}} \mathbf{c}_d \right) , \label{eq:control_input}
\end{align}
cancels out the additional terms in~\eqref{eq:dot_e} and ensures the desired dynamics $\dot{\mathbf{e}}=Z(\mathbf{E,e},\mathbf{f}_{ext})$. To derive the
final computed torque law, we consider the dynamics of a manipulator with joint configuration $\mathbf{q}\in\mathbb{R}^n$
\begin{align}
	\mathbf{H(\bm{\theta})}\ddot{\bm{\theta}}+\mathbf{N}(\bm{\theta},\dot{\bm{\theta}}) & =\boldsymbol{\tau}_u + \mathbf{J}(\bm{\theta})^{\trsp} \mathbf{f}_{ext} ,
	\label{eq:manipulator_dynamics}
\end{align}
where $\mathbf{H}(\bm{\theta})\in \mathbb{R}^{n \times n}$ denotes the positive-definite inertia matrix, the vector $\mathbf{N}(\bm{\theta}, \dot{\bm{\theta}})\in\mathbb{R}^n$ collects the nonlinear terms stemming from Coriolis effects and gravity, and $\boldsymbol{\tau}_u\in \mathbb{R}^n$ are the actuator torques. 
Furthermore, the geometric Jacobian $\mathbf{J}(\bm{\theta})\in \mathbb{R}^{6 \times n}$ is used to
map the generalized force $\mathbf{f}_{ext}\in \mathbb{R}^6$, expressed w.r.t. the end-effector frame $\mathbf{C}$, to its equivalent
joint torque. In the following, we drop dependencies on $\bm{\theta}$ for notational simplicity.    

Via the Jacobian, the kinematic quantities can be mapped to task space by $\mathbf{c}=\mathbf{J}\dot{\mathbf{q}}$ and $\dot{\mathbf{c}}=\mathbf{J}\ddot{\mathbf{q}}+\dot{\mathbf{J}}\dot{\mathbf{q}}$. 
Therefore,
considering the control input in~\eqref{eq:control_input}, the joint-level reference acceleration can be computed as
\begin{align}
	\ddot{\mathbf{q}}_u & =\mathbf{J}^{\dagger}\left(\mathbf{u} - \dot{\mathbf{J}}\dot{\mathbf{q}}\right)+\ddot{\mathbf{q}}_{ns} . \label{eq:q_u}
\end{align}
The additional term $\ddot{\mathbf{q}}_{ns}$ in~\eqref{eq:q_u} denotes an added acceleration in the the nullspace of the Jacobian $\mathbf{J}$ in order to resolve
redundancy, as we are working with a $7$-DoF platform~\cite{Sici09}. This is done by formulating an attractor to a desired reference
joint configuration.
Inserting~\eqref{eq:q_u} in the manipulator dynamics~\eqref{eq:manipulator_dynamics} after inverting them yields the final computed-torque law
\begin{align}
	\boldsymbol{\tau}_u&=\mathbf{H}\left[\mathbf{J}^{\dagger}\left(\mathbf{u} - \dot{\mathbf{J}}\dot{\mathbf{q}}\right)+\ddot{\mathbf{q}}_{ns}\right]+\mathbf{N-J}^{\trsp} \mathbf{f}_{ext}.\label{eq:control_law}
\end{align}

We alleviate the need to measure the external forces by choosing $\mathbf{M}_d=(\mathbf{JH}^{-1}\mathbf{J}^{\trsp})^{-1}$ to be the manipulator inertia matrix $\mathbf{H}$ projected to operational space~\cite{Sici09}. 
For this specific choice of $\mathbf{M}_d$, the external wrench $\mathbf{f}_{ext}$ cancels in~\eqref{eq:control_law} and the input control~\eqref{eq:control_input}. 
This comes at the cost of a configuration-dependent impedance behavior. 
If certain directions of free motion require increased accuracy, we selectively add an integral action on the error term $\mbox{Log}(\mathbf{E})$ in~\eqref{eq:control_input}.

In task-space directions which are absent of interaction forces, the postulated impedance control law is equivalent to
an inverse dynamics position control scheme~\cite{Sici09}. Therefore, the closed-loop behavior is the more degraded by
disturbances the more compliant the controller is made. Consequently, we place high importance on accurate modeling of
the manipulator dynamics which is described below.

\subsection{Model Identification}
\label{app:modelID}
In order to obtain more accurate estimates of the robot dynamics model parameters, we proceed as follows.
An inverse model of the robot is identified to map the desired joint accelerations to actuation torques in the computed-torque control law~\eqref{eq:control_law}.
Provided that there are no external forces acting on the robot during the identification experiments, the robot dynamics~\eqref{eq:manipulator_dynamics} can be reorganized as $\bm{\tau}_u = \bm{\Phi}(\bm{\theta}, \dot{\bm{\theta}}, \ddot{\bm{\theta}}) \bm{\Psi}$, such that they are linear in the so-called barycentric parameters~\cite{lewis2003robot}, $\bm{\Psi} = \left[ \bm{\Psi}_1^\trsp, \ldots, \bm{\Psi}_7^\trsp \right]^\trsp$.
Since $\bm{\Phi}$ is not unique in general, we assume an equivalent relation $\bm{\tau}_u~=~\hat{\bm{\Phi}}(\bm{\theta}, \dot{\bm{\theta}}, \ddot{\bm{\theta}}) \hat{\bm{\Psi}}$ where the columns of $(\bm{\theta}, \dot{\bm{\theta}}, \ddot{\bm{\theta}})~\mapsto~\hat{\bm{\Phi}}(\bm{\theta}, \dot{\bm{\theta}}, \ddot{\bm{\theta}})$ are linearly independent.
Then, using the experimental data, we can estimate the barycentric parameters by solving the following regression problem:
\begin{equation}\label{eq:regression}
	\min_{\bm{\theta}(\cdot)} \, \sum_{k} \left\| \bm{\Phi}\left(\bm{\theta}(t_k), \dot{\bm{\theta}}(t_k), \ddot{\bm{\theta}}(t_k)\right) \bm{\Psi} - \bm{\tau}_u\left(t_k\right) \right\|^2.
\end{equation}

To this end, we designed an experiment that ``maximizes'' the information contained in the recorded data.
Following~\cite{Swevers1997}, a parametric harmonic trajectory is selected:
$\bar{\bm{\theta}}(t) := \bar{\bm{\theta}}_0 + \sum_{l=1}^{n_\text{h}} \bm{a}_l \sin(k \omega_\text{b} t) + \bm{b}_l \cos(k \omega_\text{b} t)$,
where $\bar{\bm{\theta}}_0,~\bm{a}_{1\ldots n_\text{h}}, \bm{b}_{1\ldots n_\text{h}}~\in~\mathbb{R}^n$, $\omega_\text{b}$ denotes some base frequency, and $n_\text{h}$ is the number of harmonics.
The base frequency is chosen sufficiently low such that a large part of the configuration space can be covered without violating physical constraints.
The number of harmonics is limited such that $n_\text{h} \omega_\text{b}$ is sufficiently low to not excite undesired high-frequency phenomena.
We then formulate the following optimization problem
\begin{equation}\label{eq:d-optimality-nlp}
	\begin{aligned}
		\max_{\bar{\bm{\theta}}_0, \bm{a}_k, \bm{b}_k\in \mathbb{R}^n} &\quad \operatorname{logdet} \left( \sum_{k = 1}^{N} \hat{\bm{\Phi}}_k^\trsp \hat{\bm{\Phi}}_k  \right)\\
		\st & \quad {\bm{\theta}}_l \leq \bar{\bm{\theta}}(t_k) \leq {\bm{\theta}}_u, \quad k=1,\ldots,N ,\\
		~   & \quad \dot{{\bm{\theta}}}_l \leq \dot{\bar{\bm{\theta}}}(t_k) \leq \dot{{\bm{\theta}}}_u, \quad k=1,\ldots,N ,\\
		~   & \quad \ddot{{\bm{\theta}}}_l \leq \ddot{\bar{\bm{\theta}}}(t_k) \leq \ddot{{\bm{\theta}}}_u, \quad k=1,\ldots,N ,\\
		~   & \quad \dddot{{\bm{\theta}}}_l \leq \dddot{\bar{\bm{\theta}}}(t_k) \leq \dddot{{\bm{\theta}}}_u, \quad k=1,\ldots,N ,
	\end{aligned}
\end{equation}
where $\hat{\bm{\Phi}}_k$ is short for $\hat{\bm{\Phi}}\big(\bar{\bm{\theta}}(t_k), \dot{\bar{\bm{\theta}}}(t_k), \ddot{\bar{\bm{\theta}}}(t_k)\big)$ and $t_k~=~0,~\ldots,~\frac{2\pi}{\omega_b}\sec$.

The optimization problem~\eqref{eq:d-optimality-nlp} is non-convex and can be solved to local optimality using a suitable Non-Linear Programming (NLP) solver.
We used CasADi~\cite{Andersson2019} to formulate the optimization problem and perform the automatic differentation, the NLP was solved by the interior-point solver IPOPT\cite{Waechter2006}.
Once measurements were collected, the convex regression problem~\eqref{eq:regression} was solved while enforcing physical feasibility of the barycentric parameters by a semi-definite programming approach~\cite{Sousa2014,Sousa2019}.
\section{6D Pose Estimation}
\label{app:6d_don_pose}

In our manipulation framework we wish to predict 6D object poses from their descriptor space representations.
Specifically, using the descriptor space image $\vec I_{\mathsf{DON}}$ of an object we select $N$ object-specific keypoints, or descriptors $\{\vec d_i\}_{i=1}^N,~\vec d_i\in\mathbb{R}^D$.
The keypoints can be selected manually, or by an automated process using object masks.
Assuming rigid objects, the set of selected keypoints form an object specific, annotated pointcloud with 3D coordinates $\vec{A} = [\vec x, \vec y, \vec z]\in \mathbb{R}^{N\times 3}$.
The coordinates $(x_i, y_i, z_i)$ can be computed by unprojecting the selected keypoint locations in camera frame.
We store a centralized version of the annotated pointcloud $\vec{A}^*,~\mbox{s.t.}~\vec{1}^\trsp \vec{A}^* = \vec{0}^\trsp$.

During object pose estimation, we evaluate the descriptor image $\vec I_{\mathsf{DON}}$ using the RGB camera image $\vec I$ and locate the most likely image coordinates of the target keypoints $\vec d_i$.
The image coordinates $(u_i, v_i)$ can be computed by 
$$ (u_i, v_i) = \arg\min_{(u, v)} \operatorname{dist}\left(\vec I_{\mathsf{DON}}(u, v), \vec d_i\right), $$
with $\operatorname{dist}(\cdot, \cdot)$ as distance metric.
To measure prediction confidence we compute $c(u,v) = \exp(-\operatorname{dist}(\vec I_{\mathsf{DON}}(u, v), \vec d_i) / \eta)$ with a temperature parameter $\eta \in \mathbb{R}^+$. 
We discard keypoints with confidence $c(u,v) < c_{low},~c_{low} \in [0, 1)$, typically the ones that are occluded in the current image.
We are now left with $M \leq N$ detected keypoints in image coordiantes.
We unproject these keypoint coordinates $(u_i, v_i)$ in camera frame using the depth image and camera intrinsics.
The unprojected keypoints form an annotated pointcloud with coordinates $\vec B~=~[\vec{x}, \vec{y}, \vec{z}] \in \mathbb{R}^{M\times 3}$.
Finally, we compute the object pose $\vec T_{\mathsf{O}} \in \operatorname{SE}(3)$ in camera coordinates by solving $\vec T_{\mathsf{O}} [\vec A^*, \vec 1_M]^\trsp = [\vec B, \vec 1_M]^\trsp$.
The resulting object pose $\vec T_{\mathsf{O}}$ is transformed to ${\vec{x}_{\mathsf{o}}\in \mathbb{R}^3 \times \mathcal{S}^3}$ through a simple homeomorphic transformation.  

Note that the above 6D pose estimation procedure might fail in several circumstances. 
As keypoints are assumed to be unique, our method is prone to fail in case of symmetric objects. 
Additionally, there might not exist a unique solution for $\vec T_{\mathsf{O}}$ in case of insufficient or colinear keypoints. 
Nevertheless, we use heuristics to define unique poses to such objects.
For example, for rotation-symmetric objects like the shaft in Fig.~\ref{fig:don_pose}, we can align one axis of the coordinate system with a keypoint pair and another with the $z$-axis of the camera.
The resulting pose will be fully determined, albeit camera-view dependent, as done in~\cite{kupcsik2021supervised}.

\section{Skills learning}
\label{app:skill_learn}
In order to understand the formulation of TP-HSMMs, we first provide a brief introduction to task-parametrized Gaussian mixture models (TP-GMMs) which are the backbone of TP-HSMMs. 
We then provide a general background on the main Riemannian tools used in this paper to extend TP-HSMMs to the Riemannian data regime.

\subsection{Task-parametrized GMM (TP-GMM)}
\label{app:tp-gmm}
We assume we are given $M$ demonstrations, each of which contains $T_m$ data points for a dataset of $N=\sum_m T_m$ total observations $\vec\xi=\left\{\vec \xi_t\right\}_{t=1}^{N}$, where~$\vec\xi_t\in\R^n$ for the sake of clarity.
Also, we assume the same demonstrations are recorded from the perspective of $P$ different coordinate systems.
In our setting, these describe the object poses in the robot workspace, hereinafter also called \emph{task parameters}.
Such task-parameters make it possible to design an object-centric learning model. 
One common way to obtain object-centric data is to transform the demonstrations from a static global coordinate system to coordinate system~$p$ by $\vec \xi_t^{(p)} = \vec A^{(p)^{-1}}(\vec \xi_t - \vec b^{(p)})$, where $\{(\vec b^{(p)},\vec A^{(p)})\}_{p=1}^P$ is the translation and rotation of the coordinate system $p$ w.r.t. the world frame, which we assume is static across the demonstration.
Then, we fit a Gaussian Mixture Model (GMM) with $K$ components in every coordinate system $p$ to the transformed demonstrations $\{\vec \xi_t^{(p)}\}_{t=1}^N$.
The resulting TP-GMM is described by the model parameters $\{\pi_k,\{\vec\mu_k^{(p)},\vec\Sigma_k^{(p)}\}_{p=1}^P\}_{k=1}^K$, where $\pi_k$ is the prior probability of each component, and $\{\vec\mu_k^{(p)},\vec\Sigma_k^{(p)}\}_{p=1}^P$ are the parameters of the $k$-th Gaussian component within coordinate system~$p$.
Note that the mixing coefficients $\pi_k$ are shared by all frames and the \mbox{$k$-th} component in coordinate system $p$ must map to the corresponding \mbox{$k$-th} component in the global frame.
Expectation-Maximization (EM) is a well-established method to learn such models.

Once learned, the TP-GMM is used to reproduce a trajectory for the learned skill.
Namely, given the observed coordinate systems $\{\vec b^{(p)},\vec A^{(p)}\}_{p=1}^P$, which represent a new configuration of the objects of interest, the learned TP-GMM is transformed into a single GMM with parameters $\{\pi_k, (\vec{\hat{\mu}}_{k}, \vec{\hat{\Sigma}}_{k})\}_{k=1}^K$.
The parmeters of this GMM is a product of affine-transformed Gaussian components across different frames, as follows 
\begin{equation}\label{eq:gaussian-product}
\vec{\hat{\Sigma}}_{k}=\left[ \sum_{p=1}^P\left(\vec{\hat{\Sigma}}_{k}^{(p)}\right)^{-1}\right]^{-1}, \vec{\hat{\mu}}_{k}=\vec{\hat{\Sigma}}_{k}\left[\sum_{p=1}^P\left(\vec{\hat{\Sigma}}_{k}^{(p)}\right)^{-1}\vec{\hat{\mu}}_{k}^{(p)}\right],
\end{equation}
where the parameters of the updated Gaussians at each coordinate system $p$ are computed as $\vec{\hat{\mu}}_{k}^{(p)} = \vec A^{(p)} \vec\mu_k^{(p)} + \vec b^{(p)}$ and ${\vec{\hat{\Sigma}}_{k}^{(p)} = \vec A^{(p)} \vec\Sigma_k^{(p)} \vec A^{(p)^\trsp}}$. 
While the task parameters may vary over time, we dropped the time index for sake of clarity.
Figure~\ref{fig:tphsmm} illustrates the object-centric structure of TP-GMM for a simple reaching task.
More details about TP-GMM can be found in~\cite{calinon2016tutorial}.

\subsection{TP-HSMMs}
\label{app:hsmm}
Hidden Semi Markov Models (HSMMs) extend the standard hidden Markov Model (HMM) by embedding temporal information of the underlying stochastic process, which means that the state process is semi-Markovian (see Fig.~\ref{subfig:hsmm}).
Specifically, a transition to the next state depends on the current state as well as on the elapsed time since the state was entered.
A task-parameterized HSMM (TP-HSMM) model is defined as
${\tphsmm = \left\{ \{a_{hk}\}_{h=1}^K, (\mu_k^D, \sigma_k^D), \pi_k, \{(\vec \mu_k^{(p)}, \vec\Sigma_k^{(p)})\}_{p=1}^P \right\}_{k=1}^K}$,
where $a_{hk}$ is the transition probability from state~$h$ to $k$, $(\mu_k^D, \sigma_k^D)$ describe the Gaussian distributions for the temporal duration of state $k$, 
$\{\pi_k, \{\vec \mu_k^{(p)}, \vec \Sigma_k^{(p)}\}_{p=1}^P\}_{k=1}^K$ equal the TP-GMM introduced earlier, representing the observation probability corresponding to state $k$.
In our HSMM, $K$ corresponds to the number of Gaussian components in the \textquote{attached} TP-GMM.

We then learn a robot skill from demonstrations by training a TP-HSMM, which captures the complete spatio-temporal nature of the demonstrations. 
This allows us to encode different skill instances and timing patterns due to additional structure gained by exploiting the HSMM transition and duration probabilities.
For motion synthesis, we exploit the HSMM forward algorithm to compute the belief state.
This allows us to extract a stepwise reference trajectory featuring the spatio-temporal patterns encoded by our TP-HSMM.
Specifically, given a partial sequence of observed data points $\{\vec\xi_\ell\}_{\ell=1}^t$, assume that the associated sequence of states in~$\tphsmm$ is given by $\vec s_t = s_1s_2\cdots s_t$.
As shown by Tanwani \etal\cite{Tanwani2016}, the probability of data point $\vec\xi_t$ belonging to state $k$, i.e., $s_t=k$, is given by the \emph{forward} variable ${\alpha_t(k)=p(s_t = k, \{\vec\xi_\ell\}_{\ell=1}^t)}$:
\begin{equation}\label{eq:forward_variable}
	\alpha_t(k) = \sum_{\tau=1}^{t-1} \sum_{h=1}^K \alpha_{t-\tau}(h) a_{hk}\, \normal(\tau|\mu_k^D, \sigma_k^D)\, o_{\tau}^t\, ,
\end{equation}
where $o_{\tau}^t = \prod_{\ell=t-\tau+1}^t \normal(\vec \xi_{\ell}|\hat{\vec \mu}_{k},\hat{\vec \Sigma}_{k})$ is the emission probability and $(\hat{\vec \mu}_{k},\hat{\vec \Sigma}_{k})$ are the parameters of the global affine-transformed Gaussian components given the learned model and the current task parameters.

\begin{figure}[t]
	\centering
	\includegraphics[width=.49\linewidth]{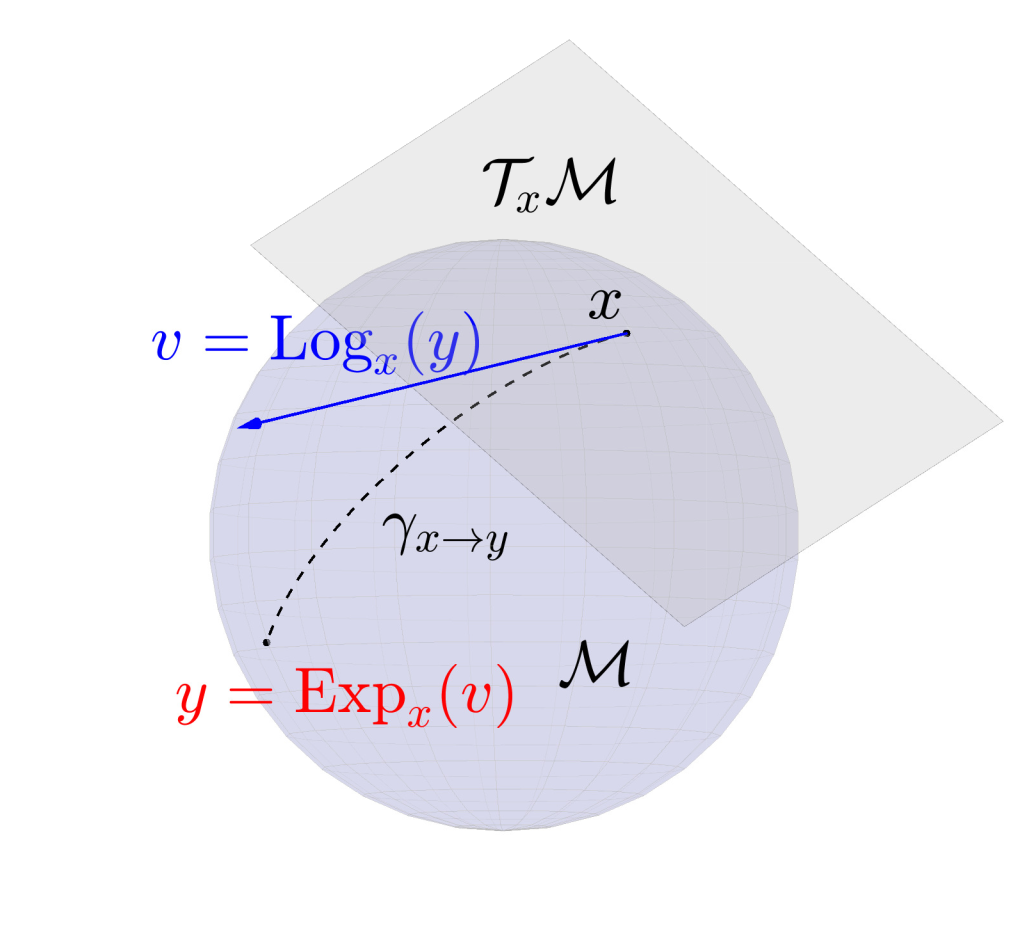}
	\includegraphics[width=.49\linewidth]{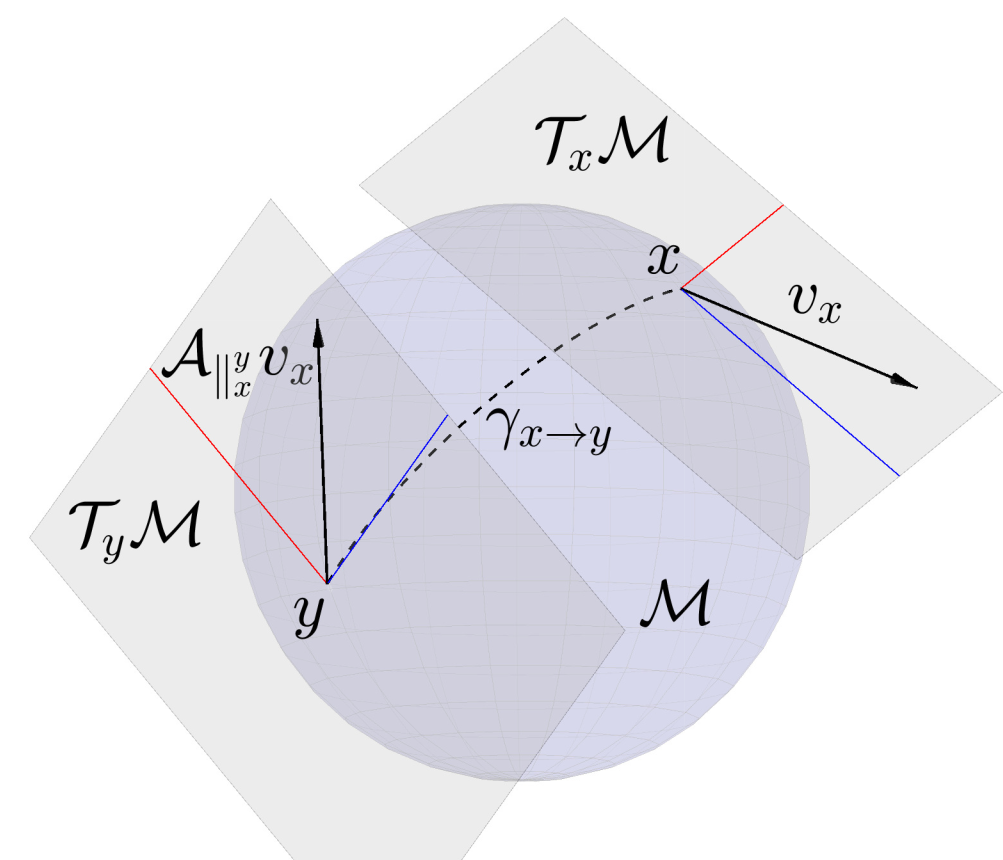}
	\caption{\textbf{Left:} Illustration of the Log and Exp maps with geodesic $\gamma_{x \rightarrow y}$ (\blackdashed). Note that $\left\| v \right\|_2 = \left\| \gamma_{x \rightarrow y} \right\|_2$. \textbf{Right:} Illustration of the parallel transport operation. $v_x$ is a vector (\blackarrow) defined in the tangent space of $x$, while the parallel transported vector $\mathcal{A}_{\parallel_x^y}v_x$ will lie in the tangent space of $y$ and is considered parallel to $v_x$.}
	\label{fig:riemannian}
	\vspace{-0.4cm}
\end{figure}

Furthermore, the same forward variable can also be used during reproduction to predict future steps until $T_m$.
In this case however, since future observations are not available, only transition and duration information are used, i.e., by setting $\normal(\vec \xi_\ell|\hat{\vec \mu}_{k},\hat{\vec \Sigma}_{k}) = 1$ for all $k$ and $\ell>t$ in~\eqref{eq:forward_variable}, as explained by Yu \etal\cite{yu2003missingHSMM}.
At last, the sequence of the most-likely states $\vec s^\star_{T_m}=s_1^\star s_2^\star\cdots s_{T_m}^\star$ is determined by choosing ${s^\star_t=\arg \max_k \alpha_t(k)}$, $\forall 1\leq t\leq T_m$.
We can then use $\vec s^\star_{T_m}$ to define a probabilistic stepwise reference computed from the mean and covariance of the corresponding Gaussian component of state $s^\star_t$.

\subsection{Riemannian Manifolds}
\label{app:riemannian}
As pointed out earlier, the demonstrations consist of time-varying end-effector poses, which correspond to position and orientation data in the form of Cartesian and quaternion variables.
To train our learning model, we need to compute statistics over non-Euclidean data (due to the orientation), thus classical Euclidean methods are inadequate.
This is due to classical techniques overlooking the geometry of the data space, and instead relying on rough approximations that lead to e.g., inaccurate estimations of Gaussian distribution parameters and unstable controllers.
We instead consider the robot task space as a Riemannian manifold $\mathcal{M}$~\cite{Zeestraten17riemannian}.
For each point $\vec x$ in the manifold $\mathcal{M}$, there exists an Euclidean tangent space $\mathcal{T}_{\vec x} \mathcal{M}$.
This allows us to carry out Euclidean operations locally, while being geometrically consistent with manifold constraints.

In order to switch between $\mathcal{M}$ and $\mathcal{T}_{\vec x} \mathcal{M}$, we resort to well-known local Riemannian operations, namely the exponential and logarithmic maps.
The exponential map ${\text{Exp}_{\vec x}: \mathcal{T}_{\vec x} \mathcal{M}\to \mathcal{M}}$ maps a point in the tangent space of point $\vec{x}$ to a point on the manifold, while maintaining the geodesic distance.
The inverse operation is called the logarithmic map $\text{Log}_{\vec x}: \mathcal{M}\to \mathcal{T}_{\vec x}\mathcal{M}$, for an illustration see Fig.~\ref{fig:riemannian}-left.
These mappings allow us to locally project manifold data to the Euclidean space, where local statistics are computed, and then project these back to the manifold.
Note that we also need to compute statistics over data that may not lie on the same tangent space.
In such situations, we exploit the parallel transport operation ${\transp{\vec x}{\vec y}: \mathcal{T}_{\vec x}\mathcal{M}\to\mathcal{T}_{\vec{y}}\mathcal{M}}$, which moves elements between tangent spaces without introducing distortion (see right side of Fig.~\ref{fig:riemannian}).
The exact form of the aforementioned operations depends on the Riemannian metric associated to the manifold, which in our case corresponds to the formulations in~\cite{Zeestraten17riemannian}.
Notice that we also exploit Riemannian manifolds to retrieve the control actions for each skill within the task plan using a Riemannian optimal controller, as explained in~\cref{subsec:reproduction}.

\subsection{Motion Trajectory Generation}
\label{app:lqt}
For motion trajectory generation we apply the Riemannian extension to the finite-horizon Linear Quadratic Tracking problem.
We assume the dynamics are defined in the tangent space of the robot end-effector state $\vec \xi \in \mathcal{M}_R$, namely $\mathcal{T}_{\vec \xi}\mathcal{M}_R$.
Also, we define the tangent-space robot state as $\vec{\mathfrak{x}}_t~=~ [\text{Log}_{\vec \xi_t}(\vec \xi_t)^\trsp,{\vec v}_t^\trsp]^\trsp~\in~\mathcal{T}_{\vec \xi_t}\mathcal{M}_R$, with velocity ${\vec v}_t$ and ${\vec v}_1 = \vec 0$.
Note that due to the differential formulation the tangent space position is $\vec 0$ and the full state becomes $\vec{\mathfrak{x}}_t = [\vec 0^\trsp, {\vec v}_t^\trsp]^\trsp$.
Given the global GMM components of the sub-sequence $\widehat{\boldsymbol{s}}_h^\star$ parametrized by $\{\vec \mu_i, \vec \Sigma_i\}$ with $i \in \widehat{\boldsymbol{s}}_h^\star$, our control problem becomes:
\begin{align}
& \min_{\vec u}  \sum_{t=1}^T \text{Log}_{\vec \xi_t} ({\vec \mu}_{\widehat{\boldsymbol{s}}_h^\star(t)})^\trsp \tilde{
	\vec \Sigma}_{\widehat{\boldsymbol{s}}_h^\star(t)} \text{Log}_{\vec \xi_t} ({\vec \mu}_{\widehat{\boldsymbol{s}}_h^\star(t)}) + \vec u_t^\trsp \vec R \vec u_t , \label{eq:lqt}                                      \\
& \qquad \mbox{s.t.}~ \vec{\mathfrak{x}}_{t+1} = \vec A \vec{\mathfrak{x}}_t + \vec B \vec u_t, ~~\vec{\mathfrak{x}}_{t+1}, \vec{\mathfrak{x}}_t, \vec u_t \in  \mathcal{T}_{\vec \xi_t}\mathcal{M}_R , \nonumber \\
& \qquad \vec A = \left[ \begin{array}{cc}
\vec I & \vec I \Delta t \\
\vec 0 & \vec I          \\
\end{array} \right], ~\vec B = \left[ \begin{array}{c} \vec 0 \\ \vec I \Delta t \end{array} \right] , \nonumber                                                                                      \\
& \qquad	\vec \xi_{t+1} = \text{Exp}_{\vec \xi_t}(\vec{\mathfrak{x}}_{t+1}) \in \mathcal{M}_R , \nonumber
\end{align}
with time horizon $T = |\widehat{\boldsymbol{s}}_h^\star|$.
After finding the optimal control sequence (in our case, acceleration) we perform forward integration to obtain the reference trajectory.
Notice that the cost function, control signal and dynamics are defined in the tangent space of the current state $\mathcal{T}_{\vec \xi_t}\mathcal{M}_R$.
To avoid the distortion of the objective we need to parallel-transport the covariance to this tangent space with $\tilde{\vec \Sigma}_i = \mathcal{A}_{\parallel_{\vec \mu_i}^{\vec \xi_t}} \vec{\Sigma}_i \mathcal{A}_{\parallel_{\vec \mu_i}^{\vec \xi_t}}^\trsp$.
Additionally, during forward integration, we have to propagate the velocity between consecutive time-steps with $\vec v_{t+1}~=~\mathcal{A}_{\parallel_{\vec \xi_t}^{\vec \xi_{t+1}}} \vec v_t~\in~\mathcal{T}_{\vec \xi_{t+1}}\mathcal{M}_R $.
We assume $\vec 0$ target velocity, therefore we omit this term in the cost function.
Solving this optimal control problem follows the same ideas as the Euclidean formulation, that is, minimizing the Bellman error in a forward-backward computation pass.
After obtaining the reference trajectory $\{\vec \xi_t\}_{t=1}^T$ in $\mathbb{R}^3 \times \mathcal{S}^3$ we compute its $\operatorname{SE}(3)$ equivalent $\{\mathbf{C}_{d,t}\}_{t=1}^T$. Finally, we rely on the Cartesian impedance controller introduced in Section~\ref{subsec:control} to track it with the robot end-effector.

\subsection{Force-based skills}
\label{app:force-lfd}
Kinesthetic demonstrations for a force-based skill contain multi-modal data:
\begin{equation}\label{eq:demos}
	\mathsf{D}_m = \left\{\boldsymbol{\xi}_t\right\}_{t=1}^{T_m} = \left\{\big{(}(\vec{x}_{\mathsf{e},t}, \dot{\boldsymbol{x}}_t, \ddot{\boldsymbol{x}}_t, \boldsymbol{f}_t), \vec{x}_{\mathsf{o},t} \big{)}\right\}_{t=1}^{T_m}, 
\end{equation}
where at each time $t$ the observation $\boldsymbol{\xi}_t$ consists of the robot pose $\boldsymbol{x}_t$, its velocity $\dot{\boldsymbol{x}}_t$, its acceleration $\ddot{\boldsymbol{x}}_t$, the external force and torque $\boldsymbol{f}_t$, and finally the object pose $\vec{x}_{\mathsf{o},t}$. 
Such observations are often obtained from a state estimation module, a perception module or dedicated sensors.

Given these demonstrations, we employ the attractor interaction model~\cite{le2021learning} to
transform the pose and force demonstrations to attractor trajectories by assuming that the attractor is driven by a virtual mass-spring-damper system. 
Consider any demonstration~$\mathsf{D}_m=\{\boldsymbol{\xi}_t\}$ from~\eqref{eq:demos}, the associated attractor trajectory $\{\boldsymbol{y}_t\}$ can be computed by:
\begin{equation} \label{eq:attractor}
	\text{Log}_{\vec{x}_{\mathsf{e},t}} (\boldsymbol{y}_t) = 
	\boldsymbol{K}^{-\rho}_t\left(\boldsymbol{K}^\nu_t\dot{\boldsymbol x}_t + \ddot{\boldsymbol x}_t - \boldsymbol f_t\right), 
\end{equation}
where $(\vec{x}_{\mathsf{e},t}, \dot{\boldsymbol{x}}_t, \ddot{\boldsymbol{x}}_t, \boldsymbol{f}_t)\in \boldsymbol{\xi}_t$ is part of the demonstration data, 
$\boldsymbol{K}^\rho_t$, $\boldsymbol{K}^\nu_t$ are the stiffness and the damping terms, the design of which is described in the sequel,
$\boldsymbol{K}^{-\rho}_t=(\boldsymbol{K}^\rho_t)^{-1}$ for brevity, and
$\text{Log}()$ is the logarithmic mapping defined in Sec.~\ref{app:riemannian}.
Intuitively, the position, velocity, acceleration and force demonstrations are transformed into a \emph{single} entity: the pose of a virtual attractor. 
Examples of the demonstrated pose trajectory and the computed attractor trajectory for the press-pcb skill are shown in Fig.~\ref{fig:press-pcb-traj}. 
It can be seen that the resulting attractor pose can differ greatly from the demonstrated pose when large velocities and sensed forces are present.  
In other words,~\eqref{eq:attractor} allows us to transform each demo $\mathsf{D}_m \in \mathsf{D}$ into an attractor demo $\boldsymbol{\Psi}_m=\{(\boldsymbol{y}_t,\, \vec{x}_{\mathsf{o},t})\}$, i.e., the attractor trajectory and the associated object pose. 
As a result, the standard procedure as described in Sec.~\ref{app:tp-gmm} can be followed to learned a TP-HSMM model from the set of attractor demonstrations~$\boldsymbol{\Psi} = \{\boldsymbol{\Psi}_m\}$. 

Furthermore, the stiffness and damping terms $\boldsymbol{K}^\rho_t$, $\boldsymbol{K}^\nu_t$ are equally important to track the demonstrated force profiles. 
Instead of solving them for each time instant, we propose to optimize these terms locally for each \emph{component} within~$\mathbf{\Theta}_y$. 
Particularly, consider a component $k$ within $\mathbf{\Theta}_y$. 
For each attractor trajectory $\boldsymbol{\Psi}_m$, the accumulative residual of the computed attractor trajectory with respect to this component is given by:
\begin{equation}\label{eq:error}
	\varepsilon_m = \sum_{\boldsymbol{\xi}_t \in \mathtt{D}_m} p_{t,k}\left(
	\text{Log}_{\boldsymbol{\mu}_k} (\vec{x}_{\mathsf{e},_t}) - \boldsymbol{K}_k^{-\rho} \left(\boldsymbol{K}_t^{\nu}\dot{\boldsymbol{x}}_t + \ddot{\boldsymbol{x}}_t - \boldsymbol{f}_t \right)\right),
\end{equation}
where $p_{t,k}$ is the probability of state $\vec{x}_{\mathsf{e},t}$ belonging to component $k$, which is a by-product of the EM algorithm when deriving $\mathbf{\Theta}_y$, $\boldsymbol{\mu}_k$ is the mean of component $k$ from $\mathbf{\Theta}_y$, $(\vec{x}_{\mathsf{e},t}, \dot{\boldsymbol{x}}_t, \ddot{\boldsymbol{x}}_t, \boldsymbol{f}_t) \in \boldsymbol{\xi}_t$ is the demonstration point at time $t$ of $\mathsf{D}_m$, $\boldsymbol{K}_k^{-\rho}$ is the inverse of the stiffness term to be optimized, while the damping term~$\boldsymbol{K}_t^{\nu}$ remains unchanged. 
Consequently, the optimal local stiffness for the component $k$ can be computed by minimizing the complete residual over all demonstrations, namely:
\begin{equation}\label{eq:whole-error}
	\boldsymbol{K}_k^{\rho,\star} = \underset{\boldsymbol{K}^\rho_{k}}{\min}\; \left\|\sum_{\mathtt{D}_m}\, \varepsilon_m \right\|\, , \quad \text{s.t.}\;\; \boldsymbol{K}^\rho_{k} \succeq \boldsymbol{0},
\end{equation}
which requires the stiffness matrix to be positive semidefinite.
The above optimization problem belongs to the semidefinite program (SDP), 
which can be solved efficiently using techniques such as interior-point methods~\cite{ boyd_vandenberghe_2004}.

To summarize, an initial choice of $\boldsymbol{K}_t^\rho$ and $\boldsymbol{K}_t^\nu$ is set to compute the attractor model as described in~\eqref{eq:attractor}.
A common choice is the default stiffness of the underlying impedance controller in~\eqref{subsubsec:CIC} and its critical damping term.
Afterwards, the local stiffness of each component can be optimized by~\eqref{eq:whole-error} as described above, denoted by~$\{\boldsymbol{K}^{\rho, \star}_k\}$. 
Note that the learned stiffness varies along the attractor trajectory in order to match the robot stiffness during kinesthetic teaching.

\subsection{Skill Condition Models} 
\label{app:condition-model}
We first recall the precondition and effect model proposed in our earlier work~\cite{Schwenkel2019Optimizing}.
In particular, the learned precondition model, denoted by~$\boldsymbol{\gamma}_{1, \mathsf{a}}$, contains TP-GMMs for the initial robot state, namely, 
$\boldsymbol{\gamma}_{1, \mathsf{a}} = \{(\vec{\hat{\mu}}^{(p)}_{1}, \vec{\hat{\Sigma}}^{(p)}_{1}),\, \forall p\in P_{1,\mathsf{a}}\}$,
where $P_{1,\mathsf{a}}$ is the chosen set of task parameters obtained from the initial system state (e.g., initial pose of objects).
In addition, we introduce here the final condition model~$\boldsymbol{\gamma}_{T, \mathsf{a}}$, which is learned in a similar way as~$\boldsymbol{\gamma}_{1, \mathsf{a}}$, but for the final robot state, i.e., 
$\boldsymbol{\gamma}_{T,\mathsf{a}} = \{(\vec{\hat{\mu}}^{(p)}_{T}, \vec{\hat{\Sigma}}^{(p)}_{T}),\, \forall p\in P_{T,\mathsf{a}}\}$,
where $P_{T,\mathsf{a}}$ is the chosen set of frames derived from the final system state.
In other words, $\boldsymbol{\gamma}_{1,\mathsf{a}}$ models the initial configuration before executing skill~$\mathsf{a}$, while $\boldsymbol{\gamma}_{T,\mathsf{a}}$ models the final configuration afterwards.
Furthermore, the learned effect model $\boldsymbol{\gamma}_{1T,\mathsf{a}}$, encapsulates TP-GMMs for the \emph{predicted} final system state, i.e.,
$\boldsymbol{\gamma}_{1T,\mathsf{a}} = \big{\{}\{(\vec{\hat{\mu}}^{(p)}_{1, \mathsf{o}}, \vec{\hat{\Sigma}}^{(p)}_{1, \mathsf{o}}),\, \forall p\in P_{1,\mathsf{a}}\},\, \forall \mathsf{o} \in \mathsf{O}_{\mathsf{a}}\cup \mathsf{e}\big{\}}$,
where $P_{1,\mathsf{a}}$ is defined in $\boldsymbol{\gamma}_{1,\mathsf{a}}$.
The differences among the foregoing models are: The task parameters for $\boldsymbol{\gamma}_{T,\mathsf{a}}$ are computed from the final system state, while those for $\boldsymbol{\gamma}_{1,\mathsf{a}}$ and $\boldsymbol{\gamma}_{1T,\mathsf{a}}$ are extracted from the initial system state.
Their full derivation is given in~\cite{Schwenkel2019Optimizing}.
For the sake of notation, we define $\boldsymbol{\Gamma}_{\mathsf{a}}\triangleq\{\boldsymbol{\gamma}_{1,\mathsf{a}}, \boldsymbol{\gamma}_{T,\mathsf{a}}, \boldsymbol{\gamma}_{1T,\mathsf{a}}\}$.

The aforementioned transition models are then used to compute the transition probabilities of the task model $\widehat{\mathbf{\Theta}}_{\boldsymbol{\mathsf{a}}^\star}$ for a given desired skill sequence, as summarized in Section~\ref{subsec:sequencing}. 
We here provide further details on this process, which works as follows: Firstly, the transition probability from one final component~$k_f$ of $\tphsmm_{\mathsf{a}_1}$ to one initial component $k_i$ of $\tphsmm_{\mathsf{a}_2}$ is:
\begin{equation}\label{eq:transition-prob}
a_{k_f,k_i} \,\propto\, \exp\left(- \sum_{p\in P_c} \operatorname{KL}\left(\boldsymbol{\gamma}^{(p)}_{T,\mathsf{a}_1}(k_f)||\boldsymbol{\gamma}_{1,\mathsf{a}_2}^{(p)}(k_i)\right)\right) ,
\end{equation}
where~$\operatorname{KL}(\cdot||\cdot)$ is the KL-divergence from~\cite{Hershey2007:ApproxKLgmm}, $\boldsymbol{\gamma}^{(p)}_{T,\mathsf{a}_1}(k_f)$ is the GMM associated with component $k_f$ for frame $p$, $\boldsymbol{\gamma}^{(p)}_{1,\mathsf{a}_2}(k_i)$ is the GMM associated with component $k_i$ for frame $p$; $P_c=P_{T,\mathsf{a}_1}\cap P_{1,\mathsf{a}_2}$ is the set of \emph{common} frames shared by these two models, which can be forced to be nonempty by always adding the global frame.
This process is repeated for all pairs of final components in $\tphsmm_{\mathsf{a}_1}$ and initial components in $\tphsmm_{\mathsf{a}_2}$.
Note that the out-going probability of any final component in $\tphsmm_{\mathsf{a}_1}$ should be normalized.

Secondly, given one final component $k_f$ of $\tphsmm_{\mathsf{a}_1}$, each component $k$ of $\tphsmm_{\mathsf{a}_2}$ should be affine-transformed as follows:
\begin{equation}\label{eq:update-gmm}
\left(\vec{\hat{\mu}}_{k}^{(\hat{p})},\, \vec{\hat{\Sigma}}_{k}^{(\hat{p})}\right) \triangleq \left(\vec\mu_k^{(p)},\,\vec\Sigma_k^{(p)}\right) \otimes \left(\vec b_{k_f}^{(\hat{p})},\,\vec A_{k_f}^{(\hat{p})}\right) ,
\vspace{-.1cm}
\end{equation}
where the operation $\otimes$ is defined as the same operation of~\eqref{eq:gaussian-product}, and $(\vec b_{k_f}^{(\hat{p})},\vec A_{k_f}^{(\hat{p})})$ is the task parameter computed from the \emph{mean} of $\boldsymbol{\Gamma}_{1T,\mathsf{a}_1}^{(\hat{p}),\mathsf{o}}(k_f)$, where $\mathsf{o}$ is the object associated with the old frame $p$ in $\tphsmm_{\mathsf{a}_1}$ and $\hat{p}$ is the new frame in $\boldsymbol{\gamma}_{1T,\mathsf{a}_1}^{\mathsf{o}}(k_f)$.
Note that the change of frames is essential to compute directly all components of $\tphsmm_{\mathsf{a}_2}$ given an initial system state of $\tphsmm_{\mathsf{a}_1}$.
The same process is also applied to each component of $\boldsymbol{\gamma}_{1T,\mathsf{a}_2}$ by changing its frames based on $\boldsymbol{\gamma}_{1T,\mathsf{a}_1}^{\mathsf{o}}(k_f)$.

Finally, other model parameters of $\widehat{\tphsmm}$ such as duration probabilities, initial and final distributions are set with minor changes from $\tphsmm_{\mathsf{a}_1}$ and $\tphsmm_{\mathsf{a}_2}$.
For instance, the duration probability of $\tphsmm_{\mathsf{a}_2}$ is duplicated to the $k_f$ multiple copies, the initial distributions $\tphsmm_{\mathsf{a}_2}$ are set to zero as the initial states of $\widehat{\tphsmm}$ correspond to those of the first model $\tphsmm_{\mathsf{a}_1}$, the final components of $\tphsmm_{\mathsf{a}_1}$ are removed since the final states of $\widehat{\tphsmm}$ are now those of $\tphsmm_{\mathsf{a}_2}$ updated to its multiple instances.
	\label{sec:appendix}
	
	\bibliographystyle{elsarticle-num}
	\bibliography{References}

\begin{thebibliography}{10}
\expandafter\ifx\csname url\endcsname\relax
  \def\url#1{\texttt{#1}}\fi
\expandafter\ifx\csname urlprefix\endcsname\relax\def\urlprefix{URL }\fi
\expandafter\ifx\csname href\endcsname\relax
  \def\href#1#2{#2} \def\path#1{#1}\fi

\bibitem{ZhihaoLiu22:RobotLearningSmartManuf}
Z.~Liu, Q.~Liu, W.~Xu, L.~Wang, Z.~Zhou,
  \href{https://doi.org/10.1016/j.rcim.2022.102360}{Robot learning towards
  smart robotic manufacturing: A review}, Robotics and Computer-Integrated
  Manufacturing 77 (2022) 102360.
\newline\urlprefix\url{https://doi.org/10.1016/j.rcim.2022.102360}

\bibitem{JingzhouSong21:AssemblyLfDGMM}
J.~Song, Q.~Chen, Z.~Li, \href{https://doi.org/10.1016/j.rcim.2020.101996}{A
  peg-in-hole robot assembly system based on gauss mixture model}, Robotics and
  Computer-Integrated Manufacturing 67 (2021) 101996.
\newline\urlprefix\url{https://doi.org/10.1016/j.rcim.2020.101996}

\bibitem{stoica2017berkeley}
I.~Stoica, D.~Song, R.~A. Popa, D.~Patterson, M.~W. Mahoney, R.~Katz, A.~D.
  Joseph, M.~Jordan, J.~M. Hellerstein, J.~E. Gonzalez, K.~Goldberg, A.~Ghodsi,
  D.~Culler, P.~Abbeel, \href{https://doi.org/10.48550/ARXIV.1712.05855}{A
  {B}erkeley view of systems challenges for {AI}} (2017).
\newline\urlprefix\url{https://doi.org/10.48550/ARXIV.1712.05855}

\bibitem{yang2018grand}
G.-Z. Yang, J.~Bellingham, P.~E. Dupont, P.~Fischer, L.~Floridi, R.~Full,
  N.~Jacobstein, V.~Kumar, M.~McNutt, R.~Merrifield, et~al.,
  \href{https://doi.org/10.1126/scirobotics.aar7650}{The grand challenges of
  science robotics}, Science Robotics 3~(14) (2018).
\newline\urlprefix\url{https://doi.org/10.1126/scirobotics.aar7650}

\bibitem{johannsmeier2019framework}
L.~Johannsmeier, M.~Gerchow, S.~Haddadin,
  \href{https://doi.org/10.1109/ICRA.2019.8793542}{A framework for robot
  manipulation: Skill formalism, meta learning and adaptive control}, in:
  {IEEE} Intl. Conf. on Robotics and Automation ({ICRA}), 2019, pp. 5844--5850.
\newline\urlprefix\url{https://doi.org/10.1109/ICRA.2019.8793542}

\bibitem{amodei2016concrete}
D.~Amodei, C.~Olah, J.~Steinhardt, P.~Christiano, J.~Schulman, D.~Mané,
  \href{https://doi.org/10.48550/ARXIV.1606.06565}{Concrete problems in {AI}
  safety} (2016).
\newline\urlprefix\url{https://doi.org/10.48550/ARXIV.1606.06565}

\bibitem{henderson2018deep}
P.~Henderson, R.~Islam, P.~Bachman, J.~Pineau, D.~Precup, D.~Meger,
  \href{https://ojs.aaai.org/index.php/AAAI/article/view/11694}{Deep
  reinforcement learning that matters}, in: {AAAI} Conf. on Artificial
  Intelligence, 2018.
\newline\urlprefix\url{https://ojs.aaai.org/index.php/AAAI/article/view/11694}

\bibitem{Billard16}
A.~Billard, S.~Calinon, R.~Dillmann,
  \href{https://doi.org/10.1007/978-3-319-32552-1_74}{Learning from humans},
  in: B.~Siciliano, O.~Khatib (Eds.), Handbook of Robotics, Springer, Secaucus,
  NJ, USA, 2016, Ch.~74, pp. 1995--2014, 2nd Edition.
\newline\urlprefix\url{https://doi.org/10.1007/978-3-319-32552-1_74}

\bibitem{Hollerbach16:ModelID}
J.~Hollerbach, W.~Khalil, M.~Gautier,
  \href{https://doi.org/10.1007/978-3-319-32552-1_6}{Model Identification},
  Springer International Publishing, 2016, pp. 113--138.
\newline\urlprefix\url{https://doi.org/10.1007/978-3-319-32552-1_6}

\bibitem{Reuss22:End2EndInvDyn}
M.~Reuss, N.~v. Duijkeren, R.~Krug, P.~Becker, V.~Shaj, G.~Neumann,
  \href{https://www.roboticsproceedings.org/rss18/p066.pdf}{{End-to-End
  Learning of Hybrid Inverse Dynamics Models for Precise and Compliant
  Impedance Control}}, in: Robotics: Science and Systems ({R:SS}), 2022.
\newline\urlprefix\url{https://www.roboticsproceedings.org/rss18/p066.pdf}

\bibitem{Chatzilygeroudis20:SurveyPS}
K.~Chatzilygeroudis, V.~Vassiliades, F.~Stulp, S.~Calinon, J.-B. Mouret,
  \href{https://doi.org/10.1109/TRO.2019.2958211}{A survey on policy search
  algorithms for learning robot controllers in a handful of trials}, {IEEE}
  Trans. on Robotics 36~(2) (2020) 328--347.
\newline\urlprefix\url{https://doi.org/10.1109/TRO.2019.2958211}

\bibitem{Pozzi21:Editorial}
M.~Pozzi, V.~R. Garate, A.~Ajoudani, K.~Harada, M.~Malvezzi, L.~Rozo, M.~A.
  Roa, \href{https://doi.org/10.1109/MRA.2021.3067685}{Emerging paradigms for
  robotic manipulation: From the lab to the productive world}, {IEEE} Robotics
  and Automation Magazine 28~(2) (2021) 10--12.
\newline\urlprefix\url{https://doi.org/10.1109/MRA.2021.3067685}

\bibitem{Gold20:MPCframework}
T.~Gold, A.~Lomakin, T.~Goller, A.~Volz, K.~Graichen,
  \href{https://doi.org/10.1109/ICMA49215.2020.9233628}{Towards a generic
  manipulation framework for robots based on model predictive interaction
  control}, in: IEEE International Conference on Mechatronics and Automation
  (ICMA), 2020, pp. 401--406.
\newline\urlprefix\url{https://doi.org/10.1109/ICMA49215.2020.9233628}

\bibitem{Jung21:ILframework}
E.~Jung, I.~Kim, \href{https://doi.org/10.3390/s21103409}{Hybrid imitation
  learning framework for robotic manipulation tasks}, Sensors 21~(10) (2021).
\newline\urlprefix\url{https://doi.org/10.3390/s21103409}

\bibitem{zhan2021a:ILRLframework}
A.~Zhan, R.~Zhao, L.~Pinto, P.~Abbeel, M.~Laskin,
  \href{https://openreview.net/forum?id=RBfB8MOxjE-}{A framework for efficient
  robotic manipulation}, in: Deep RL Workshop NeurIPS 2021, 2021.
\newline\urlprefix\url{https://openreview.net/forum?id=RBfB8MOxjE-}

\bibitem{Singh19:RLframework}
A.~Singh, L.~Yang, C.~Finn, S.~Levine,
  \href{https://doi.org/10.15607/RSS.2019.XV.073}{End-to-end robotic
  reinforcement learning without reward engineering}, in: Robotics: Science and
  Systems ({R:SS}), 2019.
\newline\urlprefix\url{https://doi.org/10.15607/RSS.2019.XV.073}

\bibitem{calinon2016tutorial}
S.~Calinon, \href{https://doi.org/10.1007/s11370-015-0187-9}{A tutorial on
  task-parameterized movement learning and retrieval}, Intelligent Service
  Robotics 9~(1) (2016) 1--29.
\newline\urlprefix\url{https://doi.org/10.1007/s11370-015-0187-9}

\bibitem{lioutikov2016learning}
R.~Lioutikov, O.~Kroemer, G.~Maeda, J.~Peters,
  \href{https://doi.org/10.1007/978-3-319-08338-4_115}{Learning manipulation by
  sequencing motor primitives with a two-armed robot}, in: Intelligent
  Autonomous Systems 13, Springer, 2016, pp. 1601--1611.
\newline\urlprefix\url{https://doi.org/10.1007/978-3-319-08338-4_115}

\bibitem{manschitz2014learning}
S.~Manschitz, J.~Kober, M.~Gienger, J.~Peters,
  \href{https://doi.org/10.1109/IROS.2014.6943187}{Learning to sequence
  movement primitives from demonstrations}, in: {IEEE/RSJ} Intl. Conf. on
  Intelligent Robots and Systems ({IROS}), 2014, pp. 4414--4421.
\newline\urlprefix\url{https://doi.org/10.1109/IROS.2014.6943187}

\bibitem{konidaris2012robot}
G.~Konidaris, S.~Kuindersma, R.~Grupen, A.~Barto,
  \href{https://doi.org/10.1177/0278364911428653}{Robot learning from
  demonstration by constructing skill trees}, Intl. Journal of Robotics
  Research 31~(3) (2012) 360--375.
\newline\urlprefix\url{https://doi.org/10.1177/0278364911428653}

\bibitem{rajan2017towards}
K.~Rajan, A.~Saffiotti,
  \href{https://doi.org/10.1016/j.artint.2017.03.003}{Towards a science of
  integrated {AI} and robotics}, Artificial Intelligence 247 (2017) 1--9,
  special Issue on AI and Robotics.
\newline\urlprefix\url{https://doi.org/10.1016/j.artint.2017.03.003}

\bibitem{de2020learning}
E.~De~Coninck, T.~Verbelen, P.~Van~Molle, P.~Simoens, B.~Dhoedt,
  \href{https://doi.org/10.1016/j.robot.2020.103474}{Learning robots to grasp
  by demonstration}, Robotics and Autonomous Systems 127 (2020) 103474.
\newline\urlprefix\url{https://doi.org/10.1016/j.robot.2020.103474}

\bibitem{ros}
M.~Quigley, K.~Conley, B.~Gerkey, J.~Faust, T.~Foote, J.~Leibs, R.~Wheeler,
  A.~Y. Ng, {ROS}: an open-source robot operating system, in: ICRA workshop on
  open source software, Vol.~3, Kobe, Japan, 2009, p.~5.

\bibitem{flexbe}
P.~Schillinger, S.~Kohlbrecher, O.~von Stryk,
  \href{https://doi.org/10.1109/ICRA.2016.7487442}{Human-robot collaborative
  high-level control with application to rescue robotics}, in: {IEEE} Intl.
  Conf. on Robotics and Automation ({ICRA}), 2016, pp. 2796--2802.
\newline\urlprefix\url{https://doi.org/10.1109/ICRA.2016.7487442}

\bibitem{malavolta2020you}
I.~Malavolta, G.~A. Lewis, B.~Schmerl, P.~Lago, D.~Garlan,
  \href{https://ieeexplore.ieee.org/document/9276566}{How do you architect your
  robots? state of the practice and guidelines for ros-based system}, in: Intl.
  Conf. on Software Engineering: Software Engineering in Practice (ICSE-SEIP),
  2020, pp. 31--40.
\newline\urlprefix\url{https://ieeexplore.ieee.org/document/9276566}

\bibitem{Florence2018}
P.~R. Florence, L.~Manuelli, R.~Tedrake,
  \href{https://proceedings.mlr.press/v87/florence18a.html}{{Dense Object Nets:
  Learning Dense Visual Object Descriptors By and For Robotic Manipulation}},
  in: Conference on Robot Learning ({CoRL}), 2018, pp. 373--385.
\newline\urlprefix\url{https://proceedings.mlr.press/v87/florence18a.html}

\bibitem{kupcsik2021supervised}
A.~G. Kupcsik, M.~Spies, A.~Klein, M.~Todescato, N.~Waniek, P.~Schillinger,
  M.~B{\"u}rger,
  \href{https://ojs.aaai.org/index.php/AAAI/article/view/16759}{Supervised
  training of dense object nets using optimal descriptors for industrial
  robotic applications}, in: {AAAI} Conf. on Artificial Intelligence, 2021, pp.
  6093--6100.
\newline\urlprefix\url{https://ojs.aaai.org/index.php/AAAI/article/view/16759}

\bibitem{Manuelli2020}
L.~Manuelli, Y.~Li, P.~Florence, R.~Tedrake,
  \href{https://proceedings.mlr.press/v155/manuelli21a.html}{{Keypoints into
  the Future: Self-Supervised Correspondence in Model-Based Reinforcement
  Learning}}, in: Conference on Robot Learning ({CoRL}), 2020, pp. 693--710.
\newline\urlprefix\url{https://proceedings.mlr.press/v155/manuelli21a.html}

\bibitem{Sundaresan2020}
P.~Sundaresan, J.~Grannen, B.~Thananjeyan, A.~Balakrishna, M.~Laskey, K.~Stone,
  J.~Gonzalez, K.~Goldberg,
  \href{https://doi.org/10.1109/ICRA40945.2020.9197121}{Learning rope
  manipulation policies using dense object descriptors trained on synthetic
  depth data}, in: {IEEE} Intl. Conf. on Robotics and Automation ({ICRA}),
  2020, pp. 9411--9418.
\newline\urlprefix\url{https://doi.org/10.1109/ICRA40945.2020.9197121}

\bibitem{Sici09}
B.~Siciliano, L.~Sciavicco, L.~Villani, G.~Oriolo, Modelling, planning and
  control, Springer, 2009.

\bibitem{Villani2016}
L.~Villani, J.~De~Schutter,
  \href{https://doi.org/10.1007/978-3-319-32552-1_9}{Force Control}, Springer
  International Publishing, 2016, pp. 195--220.
\newline\urlprefix\url{https://doi.org/10.1007/978-3-319-32552-1_9}

\bibitem{Ott2008}
C.~Ott, \href{https://doi.org/10.1007/978-3-540-69255-3_3}{Cartesian Impedance
  Control: The Rigid Body Case}, Springer Berlin Heidelberg, 2008, pp. 29--44.
\newline\urlprefix\url{https://doi.org/10.1007/978-3-540-69255-3_3}

\bibitem{lewis2003robot}
F.~L. Lewis, D.~M. Dawson, C.~T. Abdallah,
  \href{https://doi.org/10.1201/9780203026953}{Robot Manipulator Control:
  Theory and Practice}, CRC Press, 2003.
\newline\urlprefix\url{https://doi.org/10.1201/9780203026953}

\bibitem{Swevers1997}
J.~Swevers, C.~Ganseman, D.~B. Tukel, J.~De~Schutter, H.~Van~Brussel,
  \href{https://doi.org/10.1109/70.631234}{Optimal robot excitation and
  identification}, {IEEE} Transactions on Robotics and Automation 13~(5) (1997)
  730--740.
\newline\urlprefix\url{https://doi.org/10.1109/70.631234}

\bibitem{Andersson2019}
J.~A. Andersson, J.~Gillis, G.~Horn, J.~B. Rawlings, M.~Diehl,
  \href{https://doi.org/10.1007/s12532-018-0139-4}{Casadi: a software framework
  for nonlinear optimization and optimal control}, Mathematical Programming
  Computation 11~(1) (2019) 1--36.
\newline\urlprefix\url{https://doi.org/10.1007/s12532-018-0139-4}

\bibitem{Waechter2006}
A.~W{\"a}chter, L.~T. Biegler,
  \href{https://doi.org/10.1007/s10107-004-0559-y}{On the implementation of an
  interior-point filter line-search algorithm for large-scale nonlinear
  programming}, Mathematical programming 106~(1) (2006) 25--57.
\newline\urlprefix\url{https://doi.org/10.1007/s10107-004-0559-y}

\bibitem{Sousa2014}
C.~D. Sousa, R.~Cortes{\~a}o,
  \href{https://doi.org/10.1177/0278364913514870}{Physical feasibility of robot
  base inertial parameter identification: A linear matrix inequality approach},
  Intl. Journal of Robotics Research 33~(6) (2014) 931--944.
\newline\urlprefix\url{https://doi.org/10.1177/0278364913514870}

\bibitem{Wensing17:InertialParamID}
P.~M. Wensing, S.~Kim, J.-J.~E. Slotine,
  \href{https://doi.org/10.1109/LRA.2017.2729659}{Linear matrix inequalities
  for physically consistent inertial parameter identification: A statistical
  perspective on the mass distribution}, {IEEE} Robotics and Automation Letters
  3~(1) (2017) 60--67.
\newline\urlprefix\url{https://doi.org/10.1109/LRA.2017.2729659}

\bibitem{Sousa2019}
C.~D. Sousa, R.~Cortes{\~a}o,
  \href{https://doi.org/10.1109/TMECH.2019.2891177}{Inertia tensor properties
  in robot dynamics identification: A linear matrix inequality approach},
  IEEE/ASME Transactions on Mechatronics 24~(1) (2019) 406--411.
\newline\urlprefix\url{https://doi.org/10.1109/TMECH.2019.2891177}

\bibitem{Traversaro2016}
S.~Traversaro, S.~Brossette, A.~Escande, F.~Nori,
  \href{https://doi.org/10.1109/IROS.2016.7759801}{Identification of fully
  physical consistent inertial parameters using optimization on manifolds}, in:
  {IEEE/RSJ} Intl. Conf. on Intelligent Robots and Systems ({IROS}), 2016, pp.
  5446--5451.
\newline\urlprefix\url{https://doi.org/10.1109/IROS.2016.7759801}

\bibitem{Osa18:ImitationLearning}
T.~Osa, J.~Pajarinen, G.~Neumann, J.~A. Bagnell, P.~Abbeel, J.~Peters,
  \href{https://doi.org/10.1561/2300000053}{An algorithmic perspective on
  imitation learning}, Foundations and Trends® in Robotics 7~(1-2) (2018)
  1--179.
\newline\urlprefix\url{https://doi.org/10.1561/2300000053}

\bibitem{RozoEtal:IROS20}
L.~Rozo, M.~Guo, A.~G. Kupcsik, M.~Todescato, P.~Schillinger, M.~Giftthaler,
  M.~Ochs, M.~Spies, N.~Waniek, P.~Kesper, M.~Burger,
  \href{https://doi.org/10.1109/IROS45743.2020.9341570}{Learning and sequencing
  of object-centric manipulation skills for industrial tasks}, in: {IEEE/RSJ}
  Intl. Conf. on Intelligent Robots and Systems ({IROS}), 2020, pp. 9072--9079.
\newline\urlprefix\url{https://doi.org/10.1109/IROS45743.2020.9341570}

\bibitem{Rozo16:pHRC_learning}
L.~Rozo, S.~Calinon, D.~G. Caldwell, P.~Jim\'{e}nez, C.~Torras,
  \href{https://doi.org/10.1109/TRO.2016.2540623}{Learning physical
  collaborative robot behaviors from human demonstrations}, {IEEE} Trans. on
  Robotics 32~(3) (2016) 513--527.
\newline\urlprefix\url{https://doi.org/10.1109/TRO.2016.2540623}

\bibitem{le2021learning}
A.~T. Le, M.~Guo, N.~van Duijkeren, L.~Rozo, R.~Krug, A.~G. Kupcsik,
  M.~B\"urger, \href{https://doi.org/10.1109/IROS51168.2021.9636828}{Learning
  forceful manipulation skills from multi-modal human demonstrations}, in:
  {IEEE/RSJ} Intl. Conf. on Intelligent Robots and Systems ({IROS}), 2021, pp.
  7770--7777.
\newline\urlprefix\url{https://doi.org/10.1109/IROS51168.2021.9636828}

\bibitem{Schwenkel2019Optimizing}
L.~Schwenkel, M.~Guo, M.~B\"{u}rger,
  \href{https://proceedings.mlr.press/v100/schwenkel20a.html}{Optimizing
  sequences of probabilistic manipulation skills learned from demonstration},
  in: Conference on Robot Learning ({CoRL}), 2019, pp. 273--282.
\newline\urlprefix\url{https://proceedings.mlr.press/v100/schwenkel20a.html}

\bibitem{Shahriari16:BOreview}
B.~Shahriari, K.~Swersky, Z.~Wang, R.~P. Adams, N.~de~Freitas,
  \href{https://doi.org/10.1109/JPROC.2015.2494218}{Taking the human out of the
  loop: A review of {B}ayesian optimization}, Proceedings of the {IEEE} 104~(1)
  (2016) 148--175.
\newline\urlprefix\url{https://doi.org/10.1109/JPROC.2015.2494218}

\bibitem{Rasmussen06}
C.~E. Rasmussen, C.~K.~I. Williams,
  \href{https://gaussianprocess.org/gpml/}{{Gaussian processes for machine
  learning}}, MIT Press, 2006.
\newline\urlprefix\url{https://gaussianprocess.org/gpml/}

\bibitem{jaquier2019bayesian}
N.~Jaquier, L.~Rozo, S.~Calinon, M.~B{\"u}rger,
  \href{https://proceedings.mlr.press/v100/jaquier20a.html}{Bayesian
  optimization meets {R}iemannian manifolds in robot learning}, in: Conference
  on Robot Learning ({CoRL}), 2019, pp. 233--246.
\newline\urlprefix\url{https://proceedings.mlr.press/v100/jaquier20a.html}

\bibitem{jaquier2021geometryaware}
N.~Jaquier, V.~Borovitskiy, A.~Smolensky, A.~Terenin, T.~Asfour, L.~Rozo,
  \href{https://openreview.net/forum?id=ovRdr3FOIIm}{Geometry-aware bayesian
  optimization in robotics using riemannian mat\'ern kernels}, in: Conference
  on Robot Learning ({CoRL}), 2021.
\newline\urlprefix\url{https://openreview.net/forum?id=ovRdr3FOIIm}

\bibitem{Englert16:BOandRL}
P.~Englert, M.~Toussaint,
  \href{http://www.roboticsproceedings.org/rss12/p33.pdf}{Combined optimization
  and reinforcement learning for manipulation skills}, in: Robotics: Science
  and Systems ({R:SS}), 2016, pp. 1--9.
\newline\urlprefix\url{http://www.roboticsproceedings.org/rss12/p33.pdf}

\bibitem{Driess17:GPclassBO}
D.~Driess, P.~Englert, M.~Toussaint,
  \href{https://doi.org/10.1109/ICRA.2017.7989111}{Constrained bayesian
  optimization of combined interaction force/task space controllers for
  manipulations}, in: {IEEE} Intl. Conf. on Robotics and Automation ({ICRA}),
  2017, pp. 902--907.
\newline\urlprefix\url{https://doi.org/10.1109/ICRA.2017.7989111}

\bibitem{Dhonthi22:STL_BO}
A.~Dhonthi, P.~Schillinger, L.~Rozo, D.~Nardi,
  \href{https://doi.org/10.1109/IROS47612.2022.9981384}{Optimizing demonstrated
  robot manipulation skills for temporal logic constraints}, in: {IEEE/RSJ}
  Intl. Conf. on Intelligent Robots and Systems ({IROS}), 2022, pp. 1255--1262.
\newline\urlprefix\url{https://doi.org/10.1109/IROS47612.2022.9981384}

\bibitem{ros_control}
S.~Chitta, E.~Marder-Eppstein, W.~Meeussen, V.~Pradeep,
  A.~Rodr{\'i}guez~Tsouroukdissian, J.~Bohren, D.~Coleman, B.~Magyar,
  G.~Raiola, M.~L{\"u}dtke, E.~Fern{\'a}ndez~Perdomo,
  \href{https://doi.org/10.21105/joss.00456}{ros\_control: A generic and simple
  control framework for {ROS}}, The Journal of Open Source Software (2017).
\newline\urlprefix\url{https://doi.org/10.21105/joss.00456}

\bibitem{tf}
T.~Foote, \href{https://doi.org/10.1109/TePRA.2013.6556373}{tf: The transform
  library}, in: IEEE International Conference on Technologies for Practical
  Robot Applications (TePRA), Open-Source Software workshop, 2013, pp. 1--6.
\newline\urlprefix\url{https://doi.org/10.1109/TePRA.2013.6556373}

\bibitem{ronneberger2015u}
O.~Ronneberger, P.~Fischer, T.~Brox,
  \href{https://doi.org/10.1007/978-3-319-24574-4_28}{U-net: Convolutional
  networks for biomedical image segmentation}, in: International Conference on
  Medical image computing and computer-assisted intervention, Springer, 2015,
  pp. 234--241.
\newline\urlprefix\url{https://doi.org/10.1007/978-3-319-24574-4_28}

\bibitem{Karaman11:SamplingMotion}
S.~Karaman, E.~Frazzoli,
  \href{https://doi.org/10.1177/0278364911406761}{Sampling-based algorithms for
  optimal motion planning}, Intl. Journal of Robotics Research 30~(7) (2011)
  846--894.
\newline\urlprefix\url{https://doi.org/10.1177/0278364911406761}

\bibitem{Chen11:ActiveVisionSurvey}
S.~Chen, Y.~Li, N.~M. Kwok,
  \href{https://doi.org/10.1177/0278364911410755}{Active vision in robotic
  systems: A survey of recent developments}, Intl. Journal of Robotics Research
  30~(11) (2011) 1343--1377.
\newline\urlprefix\url{https://doi.org/10.1177/0278364911410755}

\bibitem{Lambeta20:DIGITsensor}
M.~Lambeta, P.-W. Chou, S.~Tian, B.~Yang, B.~Maloon, V.~R. Most, D.~Stroud,
  R.~Santos, A.~Byagowi, G.~Kammerer, D.~Jayaraman, R.~Calandra,
  \href{https://doi.org/10.1109/LRA.2020.2977257}{Digit: A novel design for a
  low-cost compact high-resolution tactile sensor with application to in-hand
  manipulation}, {IEEE} Robotics and Automation Letters 5~(3) (2020)
  3838--3845.
\newline\urlprefix\url{https://doi.org/10.1109/LRA.2020.2977257}

\bibitem{Sola18}
J.~Sola, J.~Deray, D.~Atchuthan, \href{https://arxiv.org/abs/1812.01537}{A
  micro {L}ie theory for state estimation in robotics}, arXiv preprint
  arXiv:1812.01537 (2018).
\newline\urlprefix\url{https://arxiv.org/abs/1812.01537}

\bibitem{Trav16}
S.~Traversaro, A.~Saccon, Multibody dynamics notation, Tech. rep., Technische
  Universiteit Eindhoven (2016).

\bibitem{Bull95}
F.~Bullo, R.~M. Murray,
  \href{https://resolver.caltech.edu/CaltechCDSTR:1995.CIT-CDS-95-010}{Proportional
  derivative (pd) control on the {E}uclidean group}, Tech. rep., California
  Institute of Technology (1995).
\newline\urlprefix\url{https://resolver.caltech.edu/CaltechCDSTR:1995.CIT-CDS-95-010}

\bibitem{Tanwani2016}
A.~K. Tanwani, S.~Calinon,
  \href{https://doi.org/10.1109/LRA.2016.2517825}{Learning robot manipulation
  tasks with task-parameterized hidden semi-{M}arkov model}, {IEEE} Robotics
  and Automation Letters 1~(1) (2016) 235--242.
\newline\urlprefix\url{https://doi.org/10.1109/LRA.2016.2517825}

\bibitem{yu2003missingHSMM}
S.-Z. Yu, H.~Kobayashi, \href{https://doi.org/10.1016/S0165-1684(02)00378-X}{A
  hidden semi-{M}arkov model with missing data and multiple observation
  sequences for mobility tracking}, Signal Processing 83~(2) (2003) 235--250.
\newline\urlprefix\url{https://doi.org/10.1016/S0165-1684(02)00378-X}

\bibitem{Zeestraten17riemannian}
M.~Zeestraten, \href{https://hdl.handle.net/11567/930621}{Programming by
  demonstration on {R}iemannian manifolds}, {P}hD thesis (2018).
\newline\urlprefix\url{https://hdl.handle.net/11567/930621}

\bibitem{boyd_vandenberghe_2004}
S.~Boyd, L.~Vandenberghe,
  \href{https://doi.org/10.1017/CBO9780511804441}{Convex Optimization},
  Cambridge University Press, 2004.
\newline\urlprefix\url{https://doi.org/10.1017/CBO9780511804441}

\bibitem{Hershey2007:ApproxKLgmm}
J.~R. Hershey, P.~A. Olsen,
  \href{https://doi.org/10.1109/ICASSP.2007.366913}{Approximating the
  {K}ullback {L}eibler divergence between {G}aussian mixture models}, in:
  {IEEE} Intl. Conf. on Acoustics, Speech, and Signal Processing ({ICASSP}),
  Vol.~4, 2007, pp. 317--320.
\newline\urlprefix\url{https://doi.org/10.1109/ICASSP.2007.366913}

\end{thebibliography}

\end{document}